\documentclass[letterpaper]{article}
\usepackage{uai2019}
\usepackage[margin=1in]{geometry}

\usepackage{microtype}
\usepackage{graphicx}
\usepackage{booktabs} 

\usepackage{natbib}
\usepackage{amssymb, amsmath,amsthm}
\usepackage{amsfonts}
\usepackage{algorithm}
\usepackage{algorithmic}
\usepackage{paralist}
\usepackage{multirow}
\usepackage{times}
\usepackage{wrapfig}

\usepackage{url,enumerate}
\usepackage{color,xcolor}
\usepackage{makeidx}  
\usepackage{amsmath,amssymb}
\usepackage[small, compact]{titlesec}
\usepackage{xspace}
\usepackage{epstopdf}
\usepackage{mathrsfs}
\usepackage{times}
\usepackage{enumerate}
\usepackage{color}
\usepackage{graphicx,epsfig}
\usepackage{amsmath,amssymb,xspace}
\usepackage{url}
\usepackage{hyperref}

\usepackage{bm}
\usepackage{bbm}
\usepackage{dsfont}
\usepackage{upgreek}
\usepackage{cleveref}
\usepackage{multirow}
\usepackage{ulem}
\usepackage{subcaption}
\newtheorem{theorem}{Theorem}[section]
\newtheorem{corollary}{Corollary}[theorem]
\newtheorem{lem}[theorem]{Lemma}
\usepackage{hyperref}

\newcommand{\tr}{{\rm tr}}
\renewcommand{\vec}{{\rm vec}}

\renewcommand{\b}{{\bf b}}

\renewcommand{\d}{{\rm d}}  

\newcommand{\f}{{\bf f}}

\renewcommand{\k}{{\bf k}}
\newcommand{\m}{{\bf m}}

\renewcommand{\u}{{\bf u}}

\newcommand{\w}{{\bf w}}
\newcommand{\x}{{\bf x}}
\newcommand{\y}{{\bf y}}

\newcommand{\A}{{\bf A}}
\newcommand{\B}{{\bf B}}
\newcommand{\C}{{\bf C}}

\newcommand{\G}{{\bf G}}
\renewcommand{\H}{{\bf H}}
\newcommand{\I}{{\bf I}}

\newcommand{\K}{{\bf K}}
\renewcommand{\L}{{\bf L}}
\newcommand{\M}{{\bf M}}
\newcommand{\N}{\mathcal{N}}  
\newcommand{\MN}{\mathcal{MN}} 

\newcommand{\Ocal}{\mathcal{O}}
\newcommand{\Dcal}{\mathcal{D}}

\newcommand{\Fcal}{\mathcal{F}}

\newcommand{\Lcal}{\mathcal{L}}

\newcommand{\R}{{\bf R}}

\newcommand{\W}{{\bf W}}
\newcommand{\X}{{\bf X}}
\newcommand{\Y}{{\bf Y}}
\newcommand{\Z}{{\bf Z}}

\newcommand{\Wcal}{{\mathcal{W}}}
\newcommand{\Ucal}{{\mathcal{U}}}

\newcommand{\bepsilon}{\boldsymbol{\epsilon}}

\newcommand{\bOmega}{\mathbf{\Omega}}

\newcommand{\bSigma}{\boldsymbol{\Sigma}}
\newcommand{\bPsi}{\boldsymbol{\Psi}}

\newcommand{\0}{{\bf 0}}

\newcommand{\ben}{\begin{enumerate}}
\newcommand{\een}{\end{enumerate}}

\newcommand{\ie}{{\textit{i.e.,}}\xspace}
\newcommand{\eg}{{\textit{e.g.,}}\xspace}
\newcommand{\etc}{{\textit{etc.}}\xspace}
\newcommand{\EE}{\mathbb{E}}

\newcommand{\cmt}[1]{}

\title{Multi-Fidelity Nonlinear Coregionalization for Physical Simulations}

\author{
	Zheng Wang, Wei Xing, Robert Kirby, Shandian Zhe\\
	University of Utah\\
	Salt Lake City, UT 84112\\	
	\texttt{wzhut@cs.utah.edu}, \texttt{wxing@sci.utah.edu}, \texttt{kirby@cs.utah.edu}, \texttt{zhe@cs.utah.edu}\\
}

\newcommand{\ours}{{MFHoGP}\xspace}

\begin{document}

\maketitle

\begin{abstract}  
The key task of physical simulation is to solve partial differential equations (PDEs) on discretized domains, which is known to be costly.  In particular, high-fidelity solutions are much more expensive than low-fidelity ones. To reduce the cost, we consider novel Gaussian process (GP) models that leverage simulation examples of different fidelities to predict high-dimensional PDE solution outputs. Existing GP methods are either not scalable to high-dimensional outputs or lack effective strategies to integrate multi-fidelity examples. To address these issues, we propose \underline{M}ulti-\underline{F}idelity \underline{H}igh-\underline{O}rder \underline{G}aussian \underline{P}rocess (\ours) that can capture complex correlations both between the outputs and between the fidelities to enhance solution estimation, and scale to large numbers of outputs.  Based on a novel nonlinear coregionalization model,  \ours propagates bases throughout fidelities to fuse information, and places a deep matrix GP prior over the basis weights to capture the (nonlinear) relationships across the fidelities. To improve inference efficiency and quality, we use bases decomposition to largely reduce the model parameters, and layer-wise matrix Gaussian posteriors to capture the posterior dependency and to simplify the computation. Our stochastic variational learning algorithm successfully handles millions of outputs without extra sparse approximations. We show the advantages of our method in several typical applications.
\end{abstract}

\section{Introduction}
Physical simulation~\citep{keane2005computational} is critical for many science and engineering problems such as climate prediction and aircraft design. The core task of physical simulation is to solve partial differential equations (PDEs) for various physical models. Given the PDE parameters and initial/boundary conditions, traditional numerical solvers~\citep{peiro2005finite} place a grid over the problem domain to discretize the PDEs and convert solving them into iteratively solving a linear system of equations. The solution field is represented by the solved function values at the grid points and hence are high-dimensional. Despite the success of traditional methods, they are known to be computationally costly~\citep{santner2003design}. Even worse, any change of the PDE parameters or initial/boundary conditions will require re-computation from scratch~\citep{oakley2002bayesian}. To reduce the cost, it is natural to consider using examples generated by the numerical solvers to train a machine learning model~\citep{kennedy2000predicting}, with which, we can directly predict the solution field (output) for new parameters and (parameterized) conditions (\ie input).

However, due to computational restrictions, the number of simulation examples is usually limited, and can be much smaller than the dimension of the solution output. Furthermore, collecting high-fidelity examples (with very accurate solution fields) is even more expensive, because we have to run the numerical solvers with very dense grids, which leads to an explosion in computation cost~\citep{keane2005computational}. In contrast, generating low-fidelity samples with coarser grids is cheaper, but low-fidelity samples can be quite inaccurate and biased. Hence, in practice we often can only obtain mixed examples where most are low-fidelity and only a few high-fidelity~\citep{peherstorfer2018survey}. Training with many low-fidelity examples can result in small variances but large biases, while training with very few high-fidelity samples can have small biases but much larger variances. To improve the prediction accuracy, it is crucial to effectively synergize and  exploit the examples of all the fidelities. 

To address this problem, we consider developing a novel Gaussian process (GP) model. While many excellent multi-output GPs can capture complex output correlations~\citep{alvarez2012kernels}, they are often not scalable to high dimensional outputs and lack strategies to exploit multiple-fidelity samples to further improve training. Although \citet{perdikaris2017nonlinear} and \citet{cutajar2019deep} fulfilled multi-fidelity GP learning, they only estimate single output functions.  
While we can extend their work outright to a deep GP~\citep{damianou2013deep} for multiple outputs, the outputs are fed into the next layer as the input of another GP and  hence cannot be high-dimensional, say, hundreds of thousands or millions. In addition, the outputs in each layer are assumed to be independent given the inputs, so their strong dependencies might not be fully captured.

We propose \ours, a multi-fidelity high-order GP  model, which can capture the complex, strong correlations both between the fidelities and between the outputs to enhance function estimation, and efficiently scale up to large numbers of outputs. Our major contributions are listed as follows.  
\begin{compactitem}
	\item We first propose a nonlinear coregionalization model for single-fidelity data. By introducing a matrix GP prior over the basis weights in the linear model of coregionalization (LMC) framework, our model is flexible enough to capture various nonlinear output correlations, while maintaining the scalability to high-dimensional outputs and a compact structure (\ie bases and weights) to enable efficient information propagation and fusion across different fidelities.
	\item Based on the nonlinear coregionalization, we propose a deep model to integrate multi-fidelity data. The model propagates bases throughout the  fidelities, and uses a deep matrix GP prior to recursively sample the basis weights in each layer, so as to absorb the information from and capture the nonlinear relationship with the previous fidelities to further enhance function learning.
		\item We develop two simple yet effective tricks to improve inference efficiency and quality. First, we impose a decomposition structure upon the bases to greatly reduce the model parameters  to save  the computational cost and to avoid overfitting. 
		Second, we propose a matrix Gaussian distribution as the variational posterior of the basis weights in each fidelity to capture their posterior dependency. The intrinsic Kronecker product structure further  simplifies computation.  We use the reparameterization trick to develop a stochastic variational learning algorithm that can handle millions of outputs without extra sparse approximations. 
\end{compactitem}

For evaluation, we first examined \ours on small datasets to predict tens of thousands of outputs which correspond to solving classical Burgers', Poisson's and heat equations in small spatial/temporal domains. We trained \ours with examples having one, two and three fidelities. In most cases, \ours outperforms the state-of-the-art multi-output GP regression methods. The visualization of individual output prediction errors shows \ours also better restores the outputs locally.  Finally, we used  \ours to predict one million dimensional pressure fields of the lid-driven cavity flows, with only a few hundreds of training examples. By leveraging samples of two fidelities, our approach often achieves significant error reduction as compared with the single-fidelity competitors.

\vspace{-0.1in}
\section{Background}
\vspace{-0.05in}
The standard GP learns a single-output function $f: \mathbb{R}^s\rightarrow \mathbb{R}$ from the training data  $\Dcal = \{(\x_1, y_1), \ldots, (\x_N, y_N)\}$ where each $\x_n$ is an \cmt{$s$-dimensional }input vector. The function values $\f = [f(\x_1), \ldots, f(\x_N)]$ are assumed to follow a multivariate Gaussian distribution, $p(\f|\X) = \N(\f|\m, \K)$, where $\m$ are the mean function values of every input and usually set to $\0$, $[\K]_{ij} = k(\x_i, \x_j)$ is a kernel function of the input vectors. The observed outputs $\y$ are assumed to be sampled from a noisy model, \eg $p(\y|\f) = \N(\y|\f, \tau\I)$. Integrating out $\f$, we obtain the marginal likelihood $p(\y|\X) = \N(\y|\0, \K_{nn} + \tau\I)$. We can maximize the likelihood to estimate the kernel parameters and noise variance $\tau$. 

Many tasks  require learning a function with multiple outputs. 
A classical multi-output regression framework is the Linear Model of Coregionalization (LMC)~\citep{journel1978mining}, which assumes the outputs are a linear combination of a set of basis vectors weighted by independent random functions. We introduce $K$ bases $\B = [\b_1, \ldots, \b_K]^\top$ and model a $d$-dimensional vector function by 
\begin{align}
\f(\x) = \sum\nolimits_{k=1}^K w_k(\x) \b_k = \B^\top \w(\x) \label{eq:LMC}
\end{align}
where $K$ is often chosen to be much smaller than $d$, and the random weight functions $\w(\x) = [w_1(\x), \ldots, w_K(\x)]^\top$ are sampled from independent GPs. In spite of a linear structure, the outputs $\f(\x)$ are still nonlinear to the input $\x$ due to the nonlinearity of the weight functions. LMC can easily scale up to a large number of outputs: once the bases $\B$ are identified, we only need to estimate a small number of GP models ($K\ll d$). For example, we can perform PCA on the training outputs to find the bases, and use the singular values as the outputs to train the weight functions. This is also referred to as PCA-GP~\citep{higdon2008computer}. 

LMC is particularly useful for our physical simulation tasks because it is very efficient and scalable to high-dimensional solution outputs. Also, the compact structure --- a small set of bases and weight functions --- can be used to efficiently propagate and fuse information across multiple fidelities. Therefore, we will ground our work on LMC (other excellent models will be discussed in Sec. 5). 

\vspace{-0.1in} 
\section{Model}
\vspace{-0.1in}
A critical bottleneck of LMC is that it can only model linear correlations among the outputs (see the illustration mentioned below), which is oversimplified for physical simulation, where the high-dimensional solution outputs are governed by complex PDEs, implying strong nonlinear correlations.
To fix this problem, one can place a GP prior over each element of the bases $\B$ (see \eqref{eq:LMC}). This method is referred to as GP regression network (GPRN)~\citep{wilson2012gaussian}, and can greatly promote the flexibility to capture nonlinear output correlations. However, it will meanwhile largely increase the computational cost--- an extra $Kd$ GP models need to be jointly estimated, which is  very expensive for large $d$. Therefore, we propose a nonlinear generalization of LMC, which not only is flexible enough to capture nonlinear output correlations, but also maintains the efficiency and scalability  to high-dimensional outputs. Based on the nonlinear generalization, we further develop a deep model to effectively integrate multi-fidelity data. 

\vspace{-0.1in}
\subsection{Nonlinear Coregionalization}\label{sec:nc}
\vspace{-0.1in}
The original LMC assumes independent random weight functions, which leads to oversimplified, linear output correlations. To see this,  given an arbitrary input $\x$,  we can derive the covariance of the outputs $\f(\x)$ according to \eqref{eq:LMC}: $\mathrm{cov}(\f) = \B^\top \mathrm{cov}\big(\w(\x)\big) \B$. Since the weight functions $\w(\x)$ are sampled independently,  $\mathrm{cov}\big(\w(\x)\big)$ must be diagonal,  and therefore $\mathrm{cov}(\f)$\cmt{the covariance of the outputs $\f(\x)$} is essentially a $d\times d$ linear kernel matrix on $\B^\top$ (which is  $d\times k$), implying linear correlations. 


To grasp the nonlinear output correlations, we break the independent assumption of the weight functions. Instead, we consider the weights also as a nonlinear function of the bases, and model their correlations with a nonlinear kernel of the bases (\eg RBF and Matern). To this end, we jointly sample the $K$ weight functions from a matrix-variate GP. Given $N$ training inputs $\X=[\x_1, \ldots, \x_N]^\top$ and $K$ bases $\B = [\b_1, \ldots, \b_K]^\top$, the weight functions' projection $\W$ (which is an $N\times K$ matrix and each element $[\W]_{ij} = w_j(\x_i)$) then follows a matrix Gaussian distribution, 
\begin{align}
p(\W|\X, \B) = \MN(\W|\0, \K, \K_{BB}), \label{eq:mgp}
\end{align}
where the row-covariance $\K$ is the kernel matrix on the inputs $\X$, $[\K]_{ij} = k(\x_i, \x_j)$,  and the column-covariance  $\K_{BB}$ the kernel matrix on the bases $\B$, $[\K_{BB}]_{mt} = k_b(\b_m, \b_t)$.
Given the weights and bases, we sample the observed $N \times d$ output matrix $\Y$ from a Gaussian noise model, $p(\Y|\W, \B) = \N\big(\vec(\Y)|\vec(\W\B), \eta^{-1}\I \big)$,
where $\vec(\cdot)$ is the vectorization and $\eta$ the inverse noise variance. 
This new model, referred to as nonlinear coregionalization,  turns out to be a GP model. 
\begin{lem}
	The marginal distribution of \cmt{the output matrix} the output $\Y$ is 
	\[
	p(\Y|\X,\B) = \N\big(\vec(\Y)|\0,  (\B^\top \K_{BB}\B)\otimes \K + \eta^{-1}\I\big).
	\]
	Given two arbitrary outputs $y_m(\x_i)$ and $y_t(\x_j)$, \ie the $m$-th output for input $\x_i$ and $t$-th output for input $\x_j$, \cmt{their covariance is} we have $\mathrm{cov}\big(y_m(\x_i), y_t(\x_j)\big) = k(\x_i, \x_j) \tilde{b}_m^\top  \K_{BB}  \tilde{b}_t + \eta^{-1}\cdot \delta(\x_i = \x_j, m=t)$, where $\tilde{b}_m$ and $\tilde{b}_t$ are the $m$-th and $t$-th column of $\B$, respectively, and $\delta(\cdot)$ is the indicator function. 
\end{lem}\label{lem:1}
The proof is given in the supplementary material. Now, we can see that given any input $\x$, $\mathrm{cov}\big(\y(\x)\big) =k(\x,\x) \B^\top \K_{BB} \B +\eta^{-1}\I$. As long as $\K_{BB}$ is constructed from a nonlinear kernel, the covariance matrix is nonlinear to the bases and so are the output correlations. The LMC can be viewed as an instance of our model with a particular choice of the bases' kernel. 
\begin{corollary}
	When we set the bases' kernel $k_b(\b_m, \b_t) = \delta(\b_m = \b_t)$ , the model is reduced to LMC with the same kernel for all the weight functions.  
\end{corollary}
Note that by placing a matrix GP prior over $\W$, we enable LMC to capture nonlinear output dependencies, without the need for any additional latent functions (like GPRN). By exploiting the inherent Kronecker product (see Lemma 3.1), we can further simplify the computation to avoid calculating the full covariance matrix~\citep{stegle2011efficient}. The extra calculation only involves  one small $K \times K$ kernel matrix on the bases, namely $\K_{BB}$ (in practice, $K$ is usually chosen to be less than $N$ ~\citep{higdon2008computer}). By contrast, GPRN places a GP prior over every element of $\B$ and hence needs to compute/estimate $Kd$ prior/posterior covariance matrices of all the latent functions in $\B$, which will be very expensive for large $d$, \eg millions ($\Ocal(N^3Kd)$ time complexity).

\vspace{-0.1in}
\subsection{Multi-Fidelity Nonlinear Coregionalization}\label{sect:model}
\vspace{-0.1in}
Next, to exploit training samples with multiple fidelities, we use the nonlinear coregionalization as the basic component and propose a deep model that propagates bases and places a deep matrix GP prior over the weight functions in all the fidelities. In each level, we use one component to sample the observed outputs in a particular fidelity. Each component inherits the bases from \cmt{the previous level }and samples the weight functions conditioned on the weights of the previous level. In this way, we capture the (nonlinear) relationships with and reuse the valuable bases from previous fidelities to enhance the predictions for the current fidelity. 

Specifically, suppose we have training examples of $F$ fidelities,  $\{(\X^{(i)}, \Y^{(i)})\}_{i=1}^F$ where $\X^{(i)}$ and $\Y^{(i)}$ are the $N_i \times s$ input and $N_i \times d$ output matrices at fidelity $i$. Note that although the solutions of different fidelities are calculated from distinct grids, we align them to the same dimension with a fixed grid via interpolation (note that it does not influence the fidelity)~\citep{zienkiewicz1977finite}.  Fidelity $i$ is lower than its successive fidelity $i+1$ and hence $N_1 > \ldots > N_F$. 
Following the standard multi-fidelity simulation setting ~\citep{perdikaris2017nonlinear,peherstorfer2018survey}, we assume the inputs of higher fidelity samples are a subset of the lower fidelity ones, \ie $\X^{(F)} \subset \ldots \subset \X^{(1)}$. 
However, our method can be trivially adjusted for non-overlapping inputs (see the discussion in Sec. 3 of the supplementary material).
Denote by $\W^{(i)}$ and $\B^{(i)}$ the bases and weights in each fidelity $i$. We sample the output matrix $\Y^{(i)}$ from $p(\Y^{(i)} |\W^{(i)}, \B^{(i)}, \{\eta_j\}_{j=1}^i) = \N(\vec(\Y^{(i)}) | \vec(\W^{(i)} \cdot \B^{(i)}), \prod\nolimits_{j=1}^i \eta_j^{-1}\cdot \I)$,
 where each $\eta_j$ is independently sampled from a Gamma prior, $p(\eta_j) = \mathrm{Gamma}(\eta_j|\alpha, 1)$ where $\alpha >1$. Note that we use a product of Gamma random variables as the inverse variance to gradually diminish the noise level with the increase of the fidelity. This is consistent with the fact that samples of higher fidelities should be more accurate and less noisy. 
 
 In the first (lowest) fidelity ($i=1$), we sample the bases $\B^{(1)}$ from a continuous prior, say, Gaussian, and the weights $\W^{(1)}$ from the matrix GP prior in \eqref{eq:mgp}. In each higher fidelity ($i>1$), we inherit the bases from the previous level, and sample additional $K$ bases $\C^{(i)}$  from the continuous prior again. We combine $\B^{(i-1)}$ and $\C^{(i)}$ to construct the bases for the current fidelity, $\B^{(i)} = [\B^{(i-1)}; \C^{(i)}]$.  In this way, we take advantage of not only the valuable bases from the previous fidelities --- an effective summary of lower fidelities' information, but also the ones specific to the current fidelity.  Between fidelities can be  complex yet strong relationships.  To capture and exploit these relationships, we involve the weights of the previous fidelity in generating the weights of the current fidelity. Specifically, we append to the current inputs $\X^{(i)}$ the corresponding basis weights of the previous fidelity, 
 $\widehat{\X}^{(i)} = [\X^{(i)}, \W^{(i-1)}(\X^{(i)}, \B^{(i-1)})]$. 
  We then sample $\W^{(i)}$ from a  matrix GP prior similar to \eqref{eq:mgp},  
 \begin{align}
 &p(\W^{(i)}|\B^{(i)}, \X^{(i)}, \W^{(i-1)}) = p(\W^{(i)}|\B^{(i)}, \widehat{\X}^{(i)}) \nonumber \\
 &=  \MN(\W^{(i)}| \0, \K^{(i)}, \K_{BB}^{(i)}) \label{eq:mgp2}
 \end{align}
 where $\K^{(i)}$ is the kernel matrix on the augmented inputs $\widehat{\X}^{(i)}$ and $\K_{BB}^{(i)}$ the kernel matrix on $\B^{(i)}$. The chain of the matrix GPs hence forms a deep matrix GP prior over all the weight functions $\{\W^{(i)}\}_{i=1}^F$to capture the (nonlinear) relationships across the fidelities. Finally, the graphical representation of our model is given in Fig. \ref{fig:graphical}.
 
\vspace{-0.1in}
\section{Algorithm}
\vspace{-0.05in}
For efficient model estimation, we develop a stochastic variational learning algorithm that jointly updates the bases $\B$, the kernel and noise parameters $\{\eta_i\}$, and the variational posterior of the weight functions $\{\W^{(i)}\}$. 

\vspace{-0.1in}
\subsection{Decomposition Structure for Bases}\label{sect:bases}
\vspace{-0.1in}
First, we introduce bases decomposition to further reduce the model parameters, the computation cost and also to avoid overfitting. In practice,  the output dimension $d$ can be very large, say, millions. Since each basis in $\{\B^{(i)}\}_{i=1}^F$ is a $d$ dimensional vector, it will introduce too many  parameters. The estimation of these parameters will be costly and the model can easily overfit the (small) data. To overcome these problems, we impose a decomposition structure on the bases to greatly reduce the parameters. Specifically, for each basis $\b^{i}_j$ in fidelity $i$ (note that $\B^{(i)} = [\b^{i}_1, \ldots, \b^{i}_K]^\top$), we introduce $R$ compositional vectors,  $U_j^i = \{\u_{j1}^i, \ldots, \u_{jR}^i\}$, each with length $\sqrt[R]{d}$, and parameterize $\b_j^{i} = \u_{j1}^i \otimes \ldots \otimes \u_{jR}^i $ where $\otimes$ is the Kronecker product. The kernel function of two bases $\b_{j_1}^i$ and $\b_{j_2}^i$ is  then defined on their compositional vectors, $k_b(\b_{j_1}^i, \b_{j_2}^i) = k_b(U_{j_1}^i, U_{j_2}^i)$. Take $d=10^6$ as an example. If we choose $R=3$,  we only need to use three $100$ dimensional compositional vectors to calculate each basis, and the parameters are reduced by $99.97\%$. The proposed structure is essentially a rank-1 CP ~\citep{Harshman70parafac} decomposition on the tensorized basis with $R$ modes. We can also use higher ranks or other decomposition structures, but this simple  structure has already shown the advantages of our model in the experiments (see Sec. \ref{sect:expr}). 

We assign a standard Gaussian prior over each compositional vector,  $p(\u^i_j)= \N(\u^i_j|\0, \I)$. We then parameterize each row of $\B^{(i)}$ and $\C^{(i)}$ by the Kronecker product of their corresponding compositional vectors. Note that the bases $\B^{(i)}$  are still constructed as $[\B^{(i-1)}; \C^{(i)}]$ when $i>1$. Denote the compositional vectors in each fidelity $i$ by $\Ucal^{(i)}=\{U_1^i, \ldots, U_K^i\}$. The joint probability now is
\begin{align}
&p(\{\Ucal^{(i)}, \W^{(i)}, \Y^{(i)}, \eta_i\}_{i=1}^F|\{\X^{(i)}\}_{i=1}^F) =  \prod_{i=1}^F \mathrm{Gam}(\eta_j|\alpha,1)\notag \\
&\cdot \prod_{i=1}^F\prod_{j=1}^K\prod_{r=1}^R \N(\u^i_{jr}|\0, \I) \prod_{i=1}^F\mathcal{MN}(\W^{(i)}|\0, \K^{(i)}, \K_{BB}^{(i)}) \notag \\
&\cdot \prod_{i=1}^F  \N\big(\vec(\Y^{(i)})|\vec(\W^{(i)}\B^{(i)}), \prod_{j=1}^i \eta_j^{-1}\I \big). \label{eq:joint2}
\end{align}
The  model inference amounts to estimating the  compositional vectors $\{\Ucal^{(i)}\}$ for the bases,  the posteriors of the weight functions in each fidelity and other parameters.  
\vspace{-0.1in}
\subsection{Layer-Wise Matrix Gaussian Posterior}
\vspace{-0.1in}
Next, we introduce a variational posterior for the weight functions in all the fidelities $\Wcal = \{\W^{(i)}\}_{i=1}^F$and construct a variational model evidence lower bound, 
$\Lcal = \EE_{q(\Wcal)} \big[\log\big(p(\{\Ucal^{(i)}, \W^{(i)}, \Y^{(i)}, \eta_i\}_{i=1}^F|\{\X^{(i)}\}_{i=1}^F)\big) - \log(q(\Wcal))\big]$.
While we can follow the standard mean-field framework to use fully independent posteriors, they will break the strong posterior dependency among the weights, and may result in inferior inference quality. 
Note that the matrix GP prior of each $\W^{(i)}$ (see \eqref{eq:mgp}  \eqref{eq:mgp2}) has incorporated (nonlinear) correlations between the weight functions. To capture the posterior dependency, we introduce a matrix Gaussian distribution as the variational posterior of each $\W^{(i)}$, consistent with the prior. The variational posterior of all the weights $\Wcal$  is then given by 
\begin{align}
q(\Wcal) = \prod_{i=1}^F q(\W^{(i)}) = \prod_{i=1}^F \MN(\W^{(i)}|\M^{(i)}, \bSigma^{(i)}, \bOmega^{(i)}), \notag 
\end{align}
where $\M^{(i)}$, $\bSigma^{(i)}$ and $\bOmega^{(i)}$ are the posterior mean, row and column covariances of each $\W^{(i)}$. Another advantage is the computational efficiency. Due to the intrinsic Kronecker product, we never need to compute the full covariance matrix of the  density~\citep{stegle2011efficient}. Instead, it can be  calculated  by the row and column covariance matrices and hence the cost is largely reduced, $q(\W^{(i)}) = \MN(\W^{(i)}|\M^{(i)}, \bSigma^{(i)}, \bOmega^{(i)})=\N\big(\vec(\W^{(i)})|\vec(\M^{(i)}), \bOmega^{(i)} \otimes  \bSigma^{(i)}\big) =\exp\big(-\frac{1}{2}\tr[{\bOmega^{(i)}}^{-1}{(\W^{(i)} - \M^{(i)})}^\top {\bSigma^{(i)}}^{-1}(\W^{(i)}-\M^{(i)})]\big)/{\big({(2\pi)}^{iNK/2}|{\bOmega^{(i)}}|^{N_i/2}|\bSigma^{(i)}|^{iK/2}\big)}$. 
\cmt{
\begin{align}
&q(\W^{(i)}) = \MN(\W^{(i)}|\M^{(i)}, \bSigma^{(i)}, \bOmega^{(i)})\notag \\
&=\N(\vec(\W^{(i)})|\0, \bOmega^{(i)} \otimes  \bSigma^{(i)}) \notag \\
& =\frac{\exp\big(\frac{1}{2}\tr({\bOmega^{(i)}}^{-1}{\W^{(i)}}^\top {\bSigma^{(i)}}^{-1}\W^{(i)})\big)}{{(2\pi)}^{iNK/2}|{\bOmega^{(i)}}|^{N_i/2}|\bSigma^{(i)}|^{iK/2}}. \notag 
\end{align}
}
Note that the same computation applies to the prior of $\{\W^{(i)}\}$. We derive the variational evidence lower bound (ELBO) finally,
\begin{align}
&\Lcal = \sum_{i=1}^F\sum_{j=1}^K\sum_{r=1}^R -\frac{1}{2}\|\u_{jr}^i\|^2 + \sum_{j=1}^i \frac{N_i}{2}\big(d\log\eta_i - \log |\K_{BB}^{(i)}|\big) \notag \\
&-\frac{1}{2}\sum_{i=1}^F  iK\log|\bSigma^{(i)}| +  (\prod_{j=1}^i\eta_j) \tr(\bSigma^{(i)})\tr(\bOmega^{(i)}\B^{(i)}\B^{{(i)}^\top}) \notag \\
&+\frac{1}{2}\sum_{i=1}^F N_i\log|\bOmega^{(i)}| -(\prod_{j=1}^i\eta_j) \big(\|\Y_i - \M^{(i)}\B^{(i)}\|^2_\Fcal \big)   \notag \\ 
&+\sum_{i=1}^F (\alpha-1)\log\eta_i - \eta_i -\frac{1}{2}\EE_{q(\Wcal)}\big[iK\log(\K^{(i)})\big]  \notag \\
&-\frac{1}{2}\sum\nolimits_{i=1}^F   \EE_{q(\Wcal)}\big[\tr\big(\K_{BB}^{{(i)}^{-1}} {\W^{(i)}}^\top \K^{{(i)}^{-1}}\W^{(i)} \big)\big] + \mathrm{const}. \label{eq:elbo} \raisetag{1.15in}
\end{align}
where $\|\cdot\|_\Fcal$ is the Frobenius norm.
\vspace{-0.1in}
\subsection{Stochastic Optimization}
\vspace{-0.1in}
We aim to maximize the variational ELBO in \eqref{eq:elbo}. However, the expectation terms involving each $\K^{(i)}$ are intractable, because they are kernel matrices on the augmented inputs $\widehat{\X}^{(i)} = [\X^{(i)}, \W^{(i-1)}(\X^{(i)}, \B^{(i-1)})]$, where the weights from the previous fidelity are (partly) coupled in the nonlinear kernels. To address this issue, we use the reparameterization trick~\citep{kingma2013auto} to calculate an unbiased stochastic gradient for optimization. In each fidelity $i$, we  sample a standard matrix Gaussian random variable, $\Z^{(i)} \sim \MN(\Z^{(i)}|\0, \I, \I)$. Then we construct a parameterized  sample, $\widetilde{\W}^{(i)} = \M^{(i)} + \L^{(i)} \Z^{(i)} \R^{{(i)}^\top}$, where $\L^{(i)}$ and $\R^{(i)}$ are the Cholesky decompositions of the row covariance $\bSigma^{(i)}$ and column covariance $\bOmega^{(i)}$ in $q(\W^{(i)})$, respectively. 
According to the following theorem, $\widetilde{\W}^{(i)}$ is guaranteed to be a sample of $q(\W^{(i)})$. 
\begin{theorem}~\citep{gupta1999matrix}
Given $n \times p$ matrix $\Z$, $m \times n$ matrix $\G$ and $p \times l$ matrix $\H$, If  $\Z \sim \MN(\cdot |\A, \bSigma, \bPsi)$, $\mathrm{rank}(\G)\le n$ and $\mathrm{rank}(\H)\le p$, then $\L\Z\R \sim \MN(\cdot | \G\bSigma\G^\top, \H^\top\bPsi\H )$. 
\end{theorem}
\begin{corollary}
	The constructed sample $\widetilde{\W}^{(i)} \sim \MN(\cdot | \M^{(i)}, \bSigma^{(i)}, \bOmega^{(i)})$, namely $q(\W^{(i)})$.
\end{corollary}
Next, we sequentially append each $\widetilde{\W}^{(i-1)}(\X^{(i)}, \B^{(i-1)})$ to $\X^{(i)}$ to obtain the augmented inputs, based on which we compute the random kernel matrix $\widetilde{\K}^{(i)}$ ($i>1$). We then replace each  $\EE_{q(\Wcal)}\big[\tr\big(\K_{BB}^{{(i)}^{-1}} \W^{{(i)}^\top} \K^{{(i)}^{-1}}\W^{(i)} \big)\big]$ and $\EE_{q(\Wcal)}\big[iK\log(\K^{(i)})\big] $ in \eqref{eq:elbo} with their unbiased estimates $\tr\big(\K_{BB}^{{(i)}^{-1}} \widetilde{\W}^{(i)^\top}\widetilde{\K}^{{(i)}^{-1}}\widetilde{\W}^{(i)} \big)$ and $iK\log(\widetilde{\K}^{(i)})$, respectively, so as to obtain an unbiased stochastic bound $\widetilde{\Lcal}$. We compute $\nabla \widetilde{\Lcal}$ as an unbiased stochastic gradient of $\Lcal$ for optimization. We can use any stochastic optimization algorithm to jointly update the basis compositional vectors $\{\Ucal^{(i)}\}$,  the variational posterior $q(\Wcal)$ (determined by $\{\M^{(i)}, \L^{(i)}, \R^{(i)} \}$) and all the other parameters. 

\vspace{-0.1in}
\subsection{Prediction}
\vspace{-0.1in}
Given a new input, the predictive distribution of the outputs is not analytical. Hence, we recursively sample the weights in each fidelity to generate posterior samples, with which we compute an empirical distribution.  Due to the space limit, we leave the details in the supplementary material. 

\vspace{-0.1in}
\subsection{Algorithm Complexity}
\vspace{-0.1in}
The time complexity of our inference algorithm is $\Ocal(\sum_{i=1}^F iN_iKd + (iK)^2iRd^{\frac{1}{R}} + (iK)^3 + N_i^3)$. Since we can always choose $R$ such that $Rd^{\frac{1}{R}} \le d$ (the simplest choice is $R=1$), the time complexity is linear to $Nd$,  where $N$ is the total number of samples. The space complexity is $\Ocal(FKd + \sum_{i=1}^F  iN_iK + (iK)^2 + (N_i)^2)$, including the storage  of the bases, the weights, and the row and column covariance matrices of the weights in each fidelity.

\vspace{-0.12in}
\section{Related Work}
\vspace{-0.05in}
Many multi-output GP regression approaches have been proposed. An excellent review is given in ~\citep{alvarez2012kernels}. A classical framework is the linear model of coregionalization (LMC)~\citep{matheron1982pour, goulard1992linear}, which introduces a set of basis vectors, and use their linear combination weighted by independent random functions to predict the output vector. A popular instance is PCA-GP~\citep{higdon2008computer} that finds a set of bases from Singular Value Decomposition (SVD) on the training outputs. The variants of PCA-GP include  KPCA-GP~\citep{xing2016manifold}, IsoMap-GP~\citep{xing2015reduced}, \etc Despite its efficiency and scalability, the standard LMC only models linear output correlations. GP regression networks (GPRNs)~\citep{wilson2012gaussian} overcome this problem by placing independent GP priors over the basis elements. While being much more expressive, GPRNs bring in much more computation cost --- the number of GPs need to be estimated is quite a few times (\eg tens) of the output dimension, and hence it will be very expensive for high dimensions.\cmt{ By contrast, our nonlinear coregionalization model places a matrix GP prior over the weight functions to enable capturing the nonlinear output correlations, while still maintaining the computational efficiency for large numbers of outputs.}
Important multi-output GP models also include convolved GPs~\citep{higdon2002space,boyle2005dependent,alvarez2019non} and multi-task GPs~\citep{bonilla2007kernel,bonilla2008multi,rakitsch2013all}. 
Both types of models are very elegant and flexible, however, they might be computationally too costly ($\Ocal((Nd)^3)$ or $\Ocal(N^3 + d^3)$ time complexity) for massive outputs. To mitigate this issue, several sparse approximations have been developed~\citep{alvarez2009sparse,alvarez2010efficient}. Recently, \citet{zhe2019scalable} tensorized the high dimensional output, and introduced latent coordinate features in the tensor space to model complex output correlations. Overall, all these methods are developed for single-fidelity data. 

To enable GP training on multi-fidelity data,  \citet{perdikaris2017nonlinear} sequentially learned a chain of GPs, where each GP estimates the output of one fidelity as a function of the current input and the output of the previous fidelity.  \citet{cutajar2019deep} jointly learned these GP models 
to propagate the uncertainty throughout different fidelities. These excellent works focus on single output functions. While we can extend them to a standard deep GP~\citep{damianou2013deep,hebbalmulti}  that samples multiple functions in each layer,  all the outputs in one layer are poured as the input to the GP in the next layer, and hence cannot be many, say, millions. Moreover, standard deep GPs consider the outputs in each layer as independent given the inputs, and might not fully capture the strong output dependencies, which is crucial for learning from a small set of training examples (in physical simulation). To address these problems, we inherit the compact structure of LMC, \ie a small number of bases and weight functions  to handle massive outputs. We first generalize LMC to flexibly capture  nonlinear output correlations. We then propagate the (decomposed) bases and place a deep matrix GP prior over the weight functions to fuse information throughout the fidelities (rather than use the entire outputs), and hence it is much more efficient. Very recently, \citet{hamelijnck2019multi} proposed a multi-task multi-resolution GP model based on GPRN, deep GP and mixture of experts~\citep{rasmussen2002infinite}. This excellent work aims to integrate sensor data with different resolutions. Distinct from our model, it needs to integrate over the sampling periods of the sensors to sample the observations, and emphasizes one particular task (output). 

Recently, a few excellent works were proposed to learn (deep) neural networks to solve PDEs~\citep{raissi2018deep,raissi2019physics}. These works differ from ours in that (1) the input is the spatial/temporal location and the output is a scalar to predict the solution function value at that location, and (2) their training and test focus on solving one particular PDE, rather than mapping parameters of different PDEs to their corresponding solution fields at a specific grid.

\vspace{-0.15in}
\section{Experiments}\label{sect:expr}
\vspace{-0.1in}
\subsection{Predicting Small Solution Fields}
\vspace{-0.05in}
\begin{figure*}
	\centering
	\setlength\tabcolsep{0.01pt}
	\begin{tabular}[c]{cccc}
		\raisebox{-0.1in}
		{
			\begin{subfigure}[t]{0.23\textwidth}
				\centering
				\includegraphics[width=\textwidth]{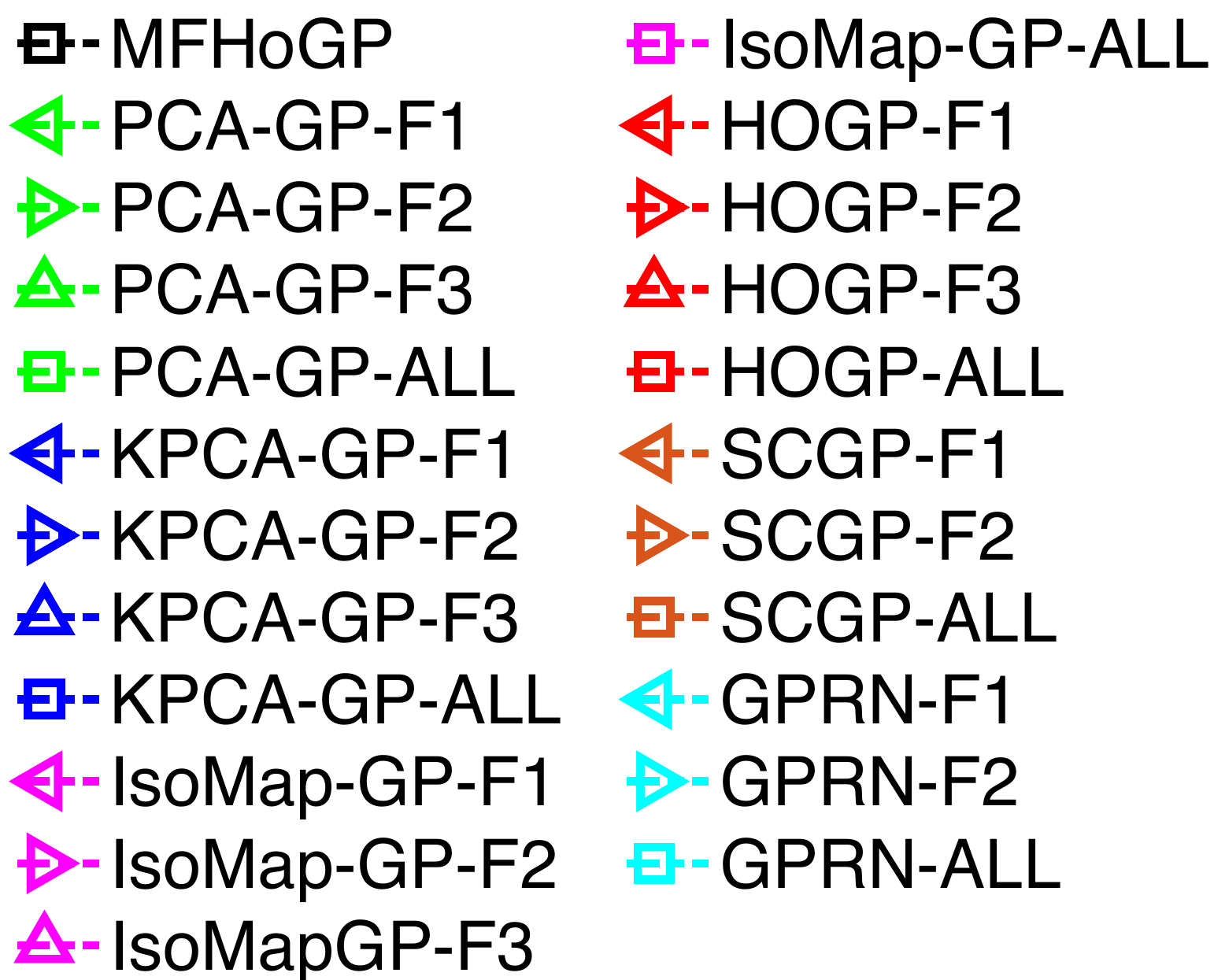}
			\end{subfigure}
		}
		&
		\begin{subfigure}[t]{0.23\textwidth}
			\centering
			\includegraphics[width=\textwidth]{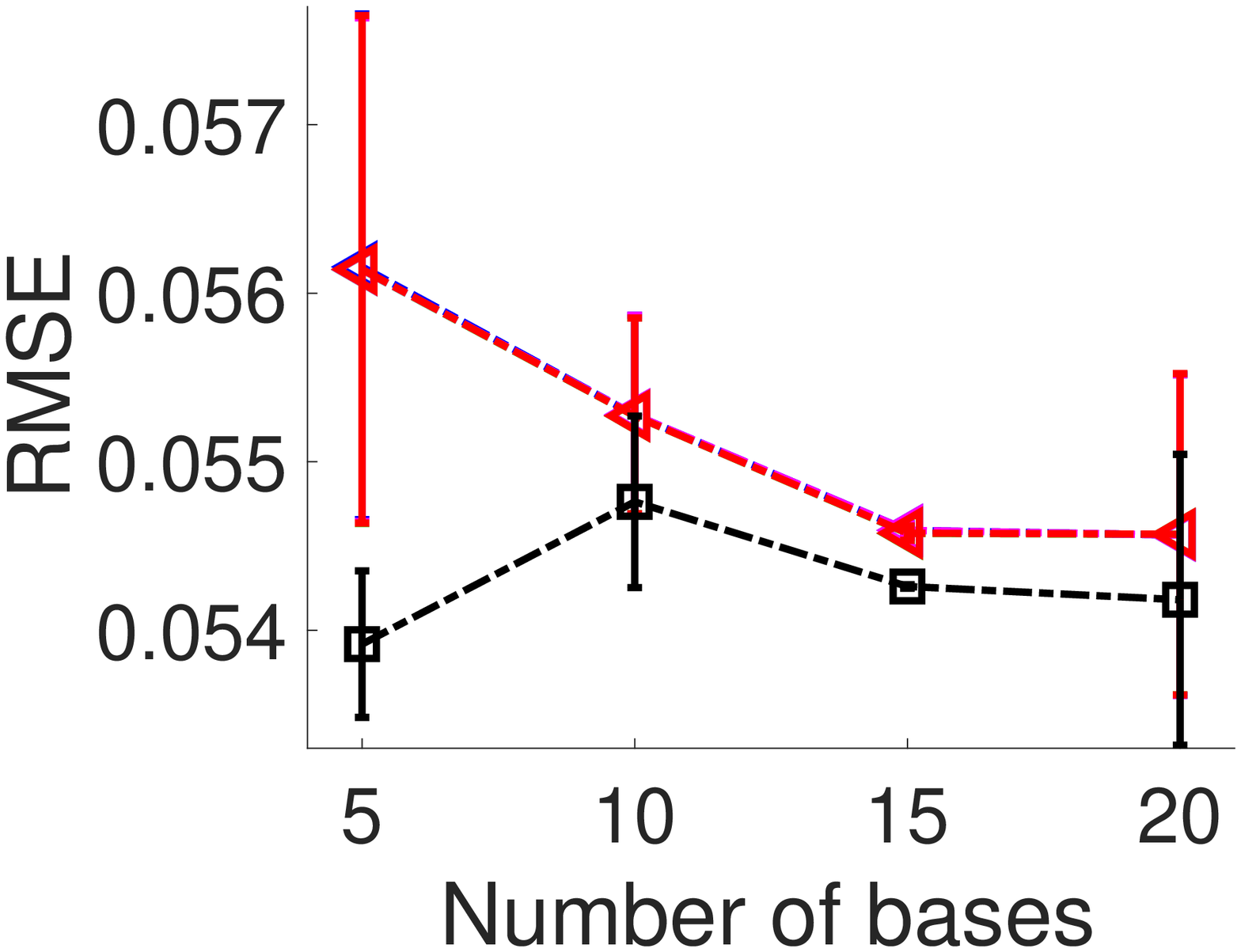}
			\caption{\textit{Burgers-I}}
		\end{subfigure} 
		&
		\begin{subfigure}[t]{0.23\textwidth}
			\centering
			\includegraphics[width=\textwidth]{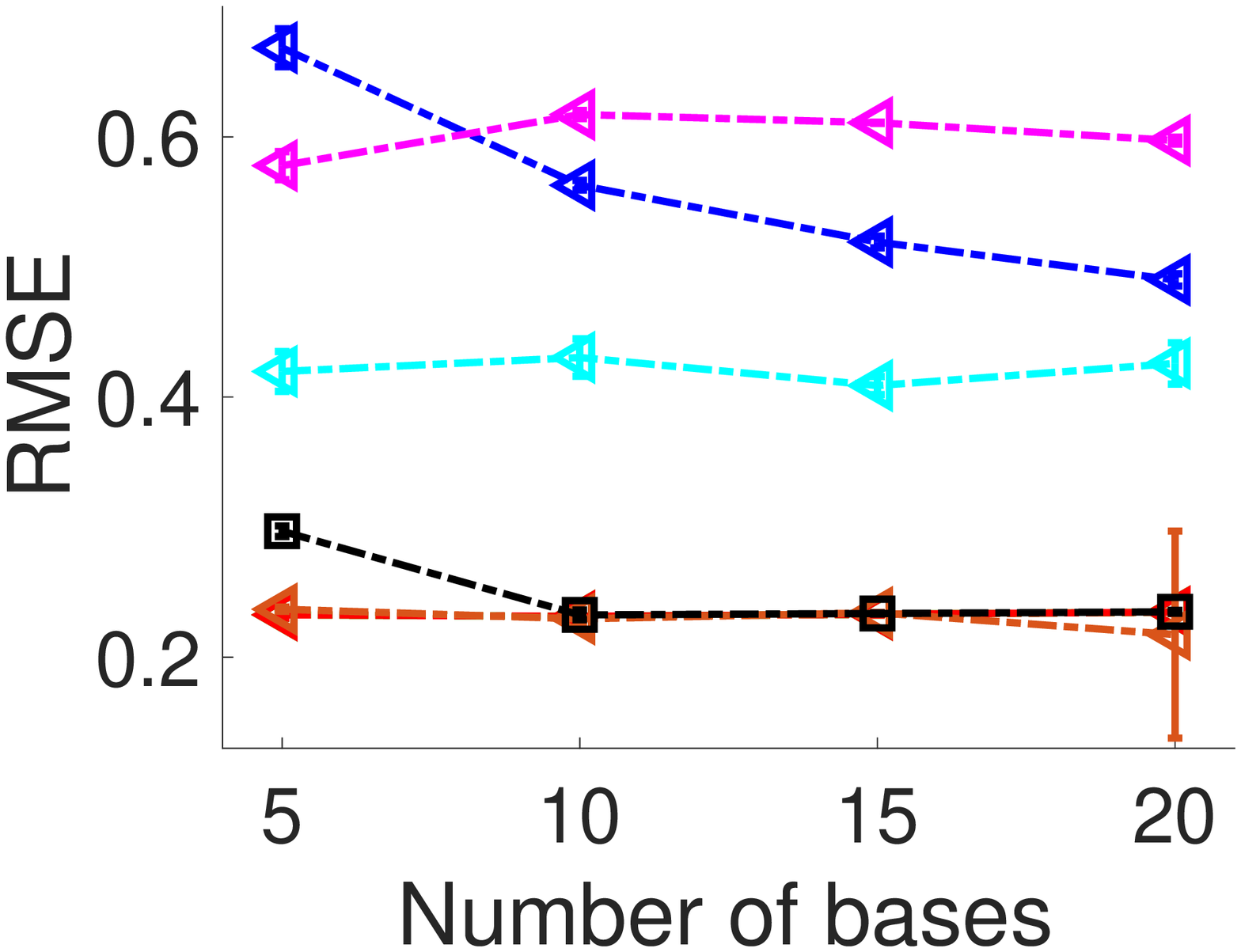}
			\caption{\textit{Poisson-I}}
		\end{subfigure}
		&
		\begin{subfigure}[t]{0.23\textwidth}
			\centering
			\includegraphics[width=\textwidth]{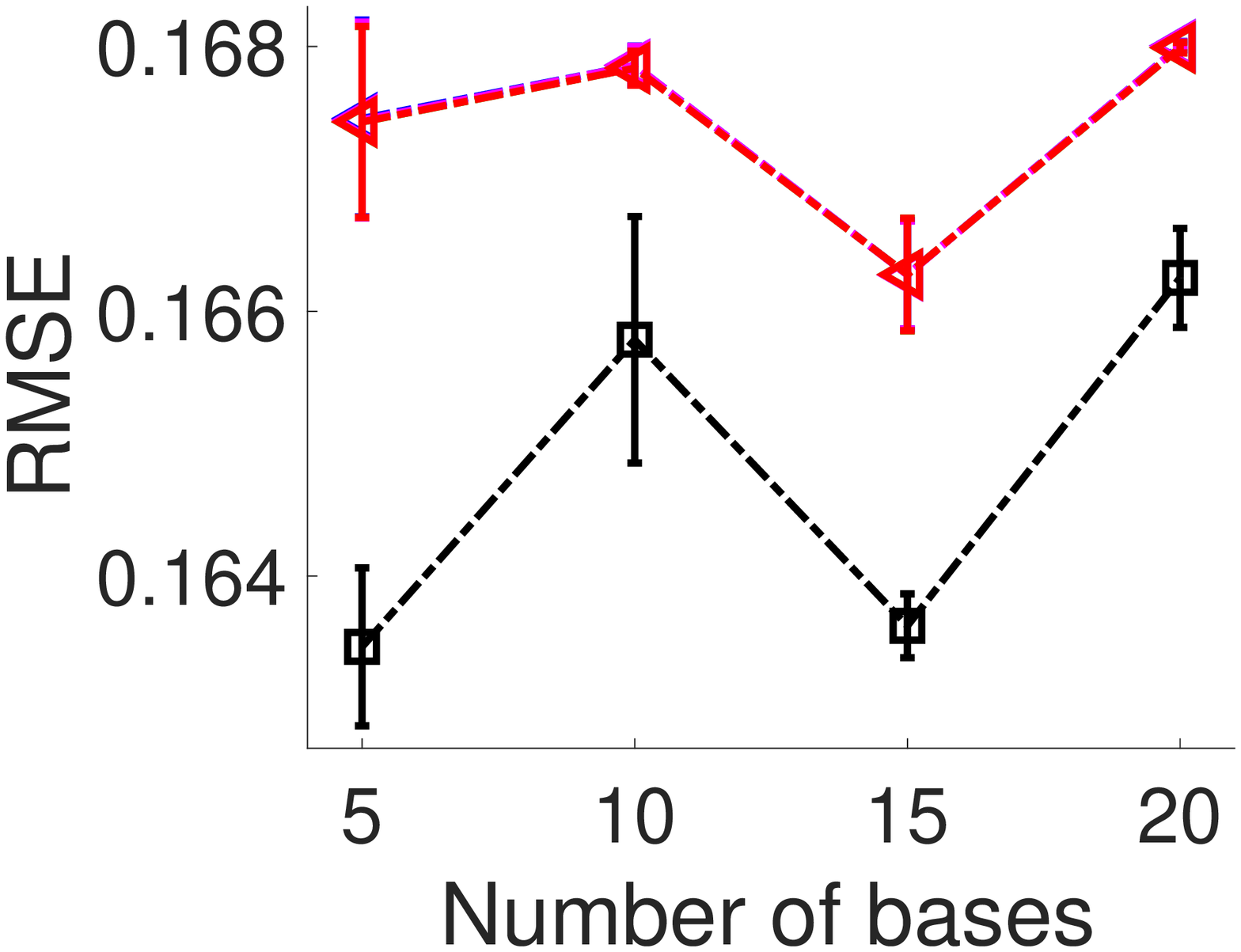}
			\caption{\textit{Heat-I}}
		\end{subfigure}\\
		\begin{subfigure}[t]{0.23\textwidth}
			\centering
			\includegraphics[width=\textwidth]{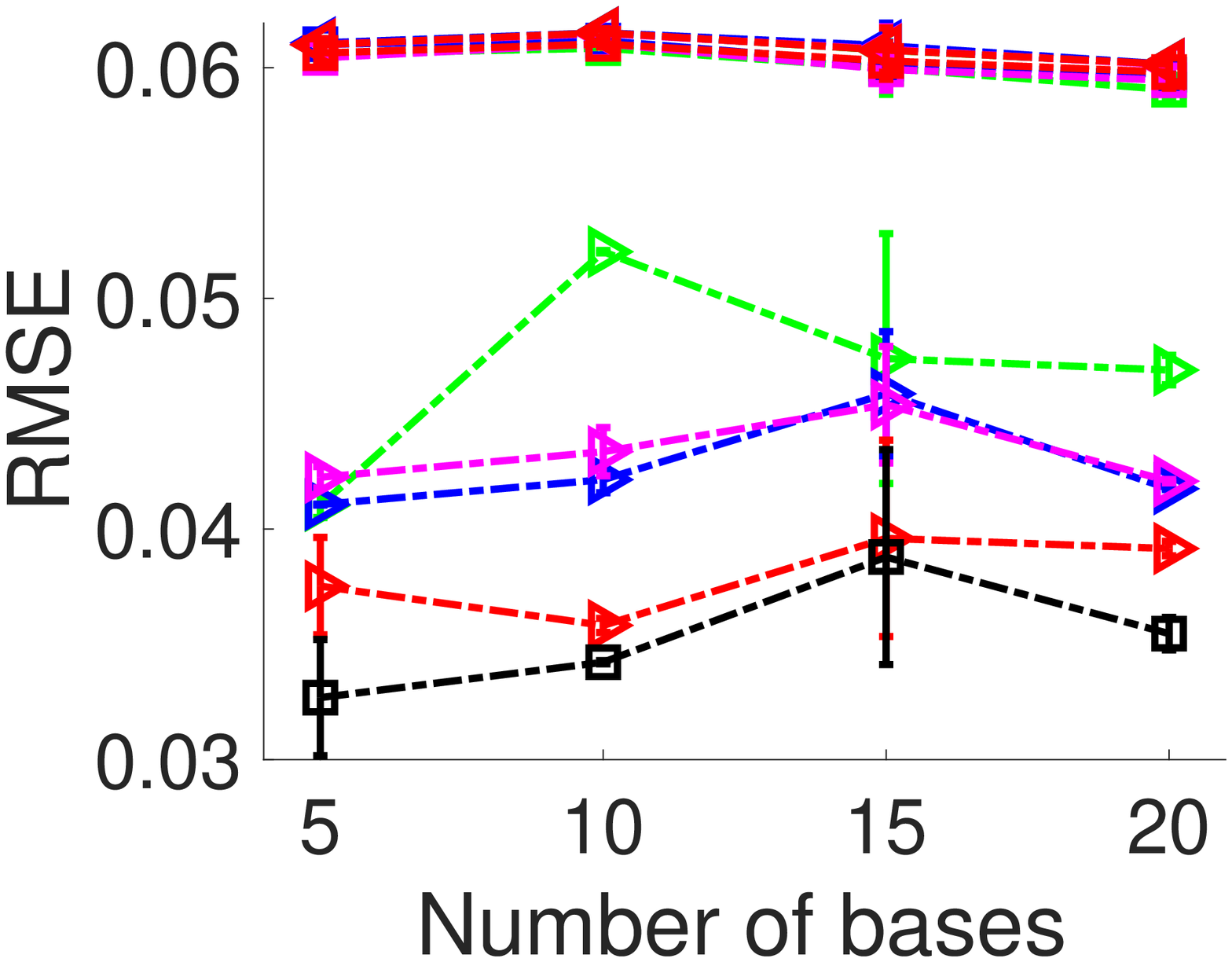}
			\caption{\textit{Burgers-II}}
		\end{subfigure} 
		&
		\begin{subfigure}[t]{0.23\textwidth}
			\centering
			\includegraphics[width=\textwidth]{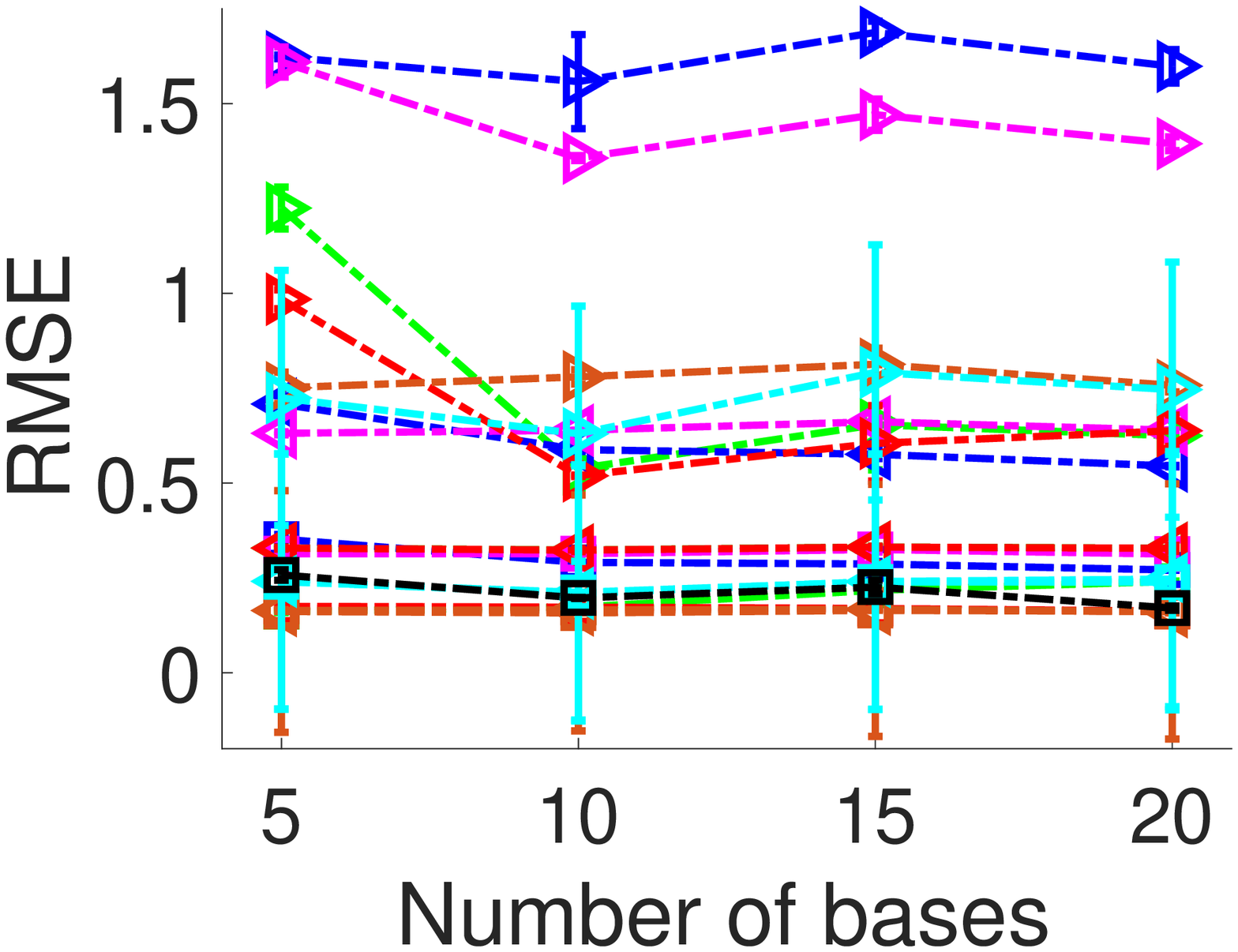}
			\caption{\textit{Poisson-II}}
		\end{subfigure}
		&
		\begin{subfigure}[t]{0.23\textwidth}
			\centering
			\includegraphics[width=\textwidth]{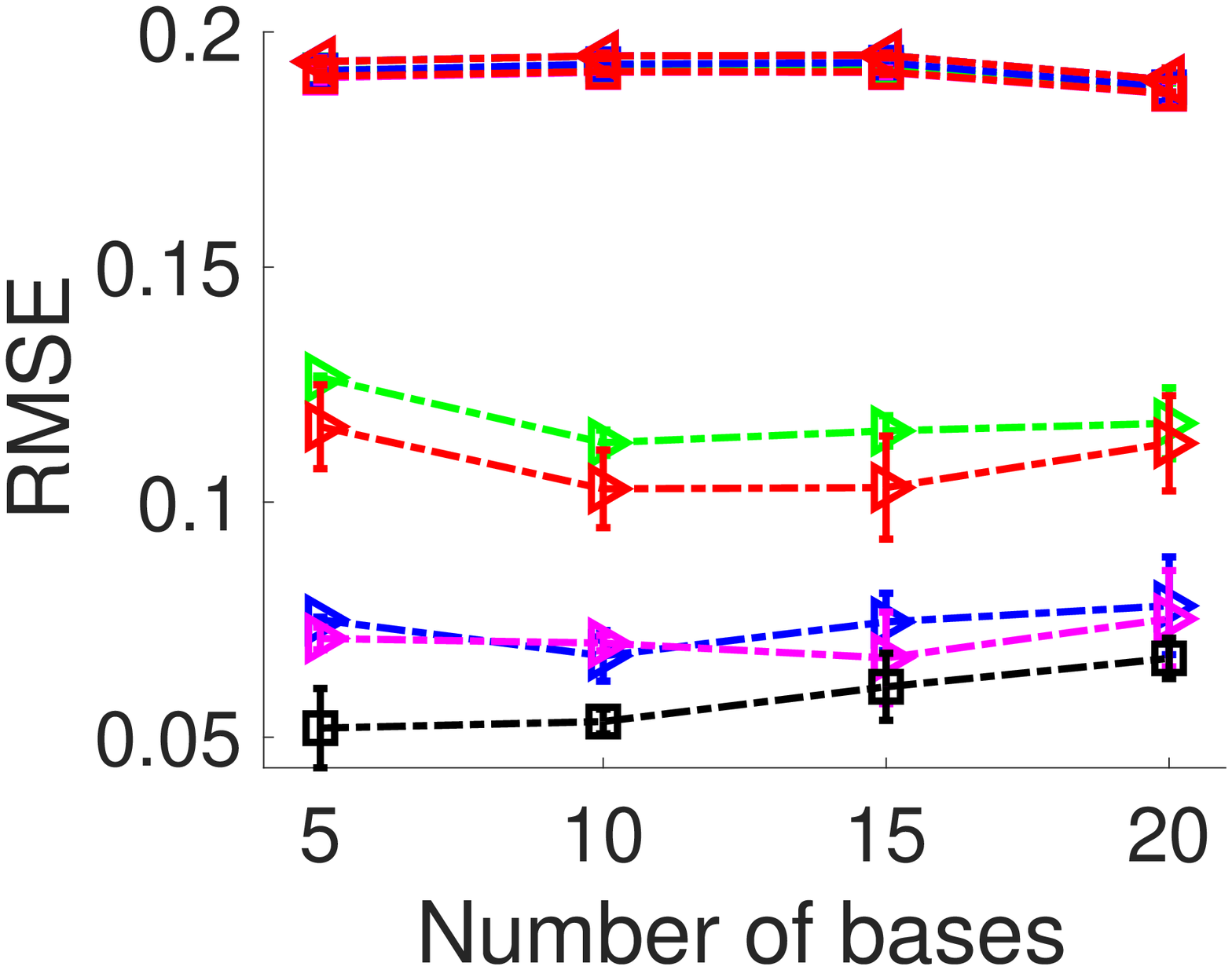}
			\caption{\textit{Heat-II}}
		\end{subfigure}
		&
		\begin{subfigure}[t]{0.23\textwidth}
			\centering
			\includegraphics[width=\textwidth]{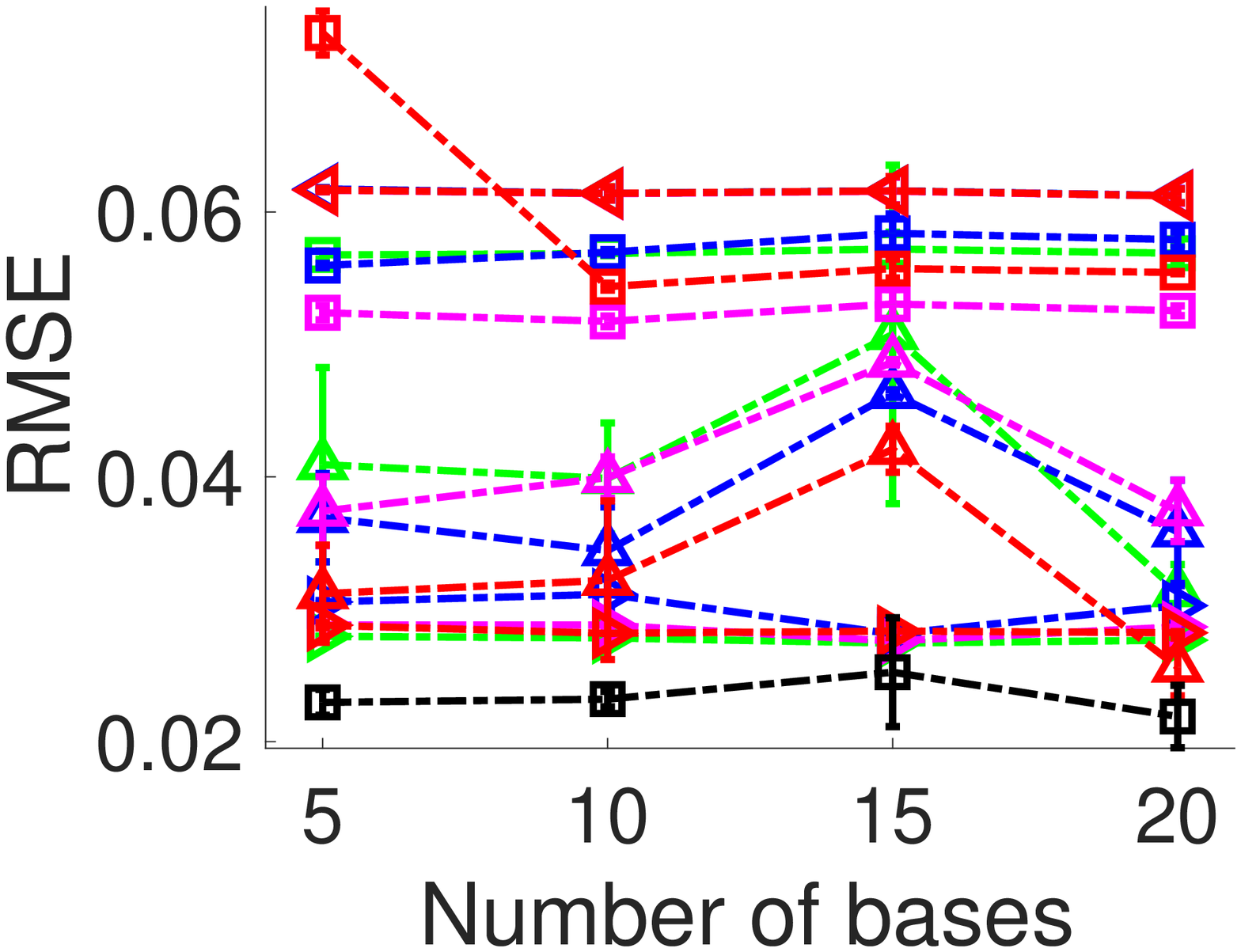}
			\caption{\textit{Burgers-III}}
		\end{subfigure}
	\end{tabular}
	\vspace{-0.1in}
	\caption{\small The root-mean-squared-error (RMSE) of all the multi-output regression methods on small datasets in seven evaluation settings. In each setting, the results are averaged from 5 runs. After the dash in each caption (\eg ``\textit{Burgers-II}'') is how many fidelities across the training data. -F\{1,2,3\} indicates the model trained with a particular fidelity's examples and -F-ALL  with all the examples. Note that quite a few methods obtained very close results and their curves overlap (\eg in ``\textit{Heat-I}'').} 	
	\label{fig:small-rmse}
	\vspace{-0.14in}
\end{figure*}

\begin{figure*}
	\centering
	\setlength\tabcolsep{0.05pt}
	\renewcommand{\arraystretch}{0.05}
	\begin{tabular}[c]{ccccccccccccccccccc}
		\multicolumn{19}{c}{\includegraphics[width=0.2\linewidth]{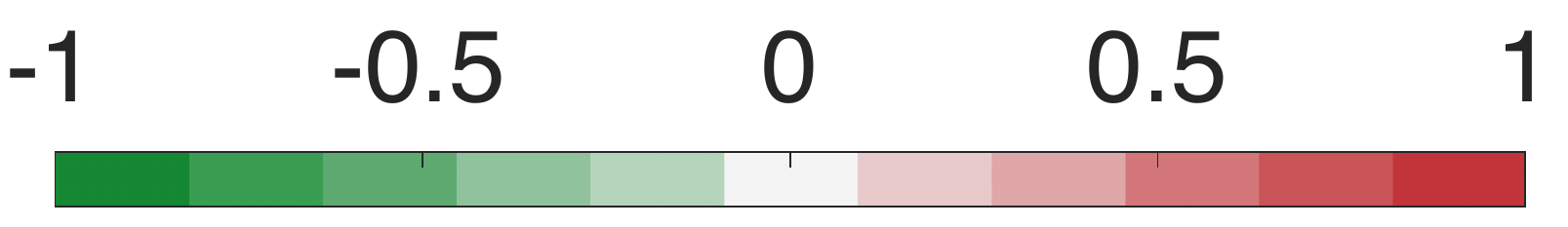}} \\
		\includegraphics[width=0.05\linewidth]{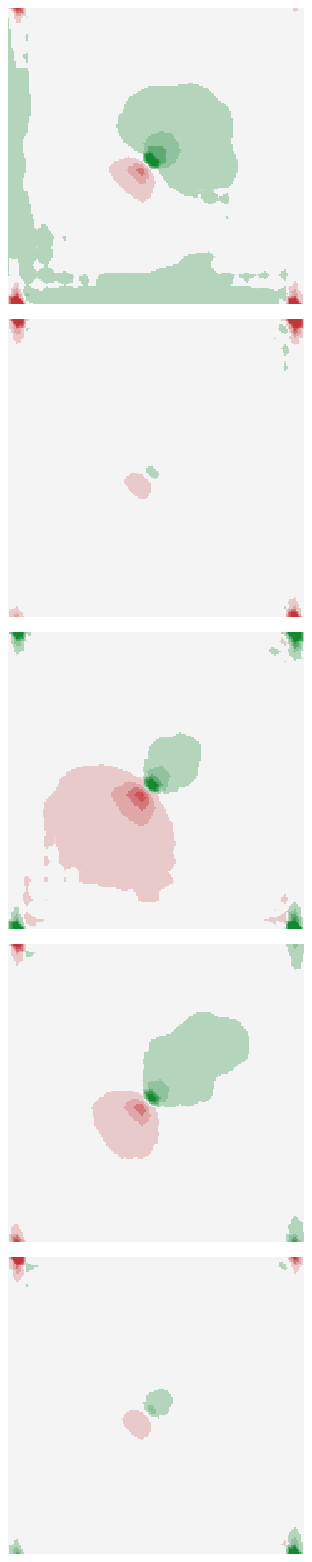}
		& \includegraphics[width=0.05\linewidth]{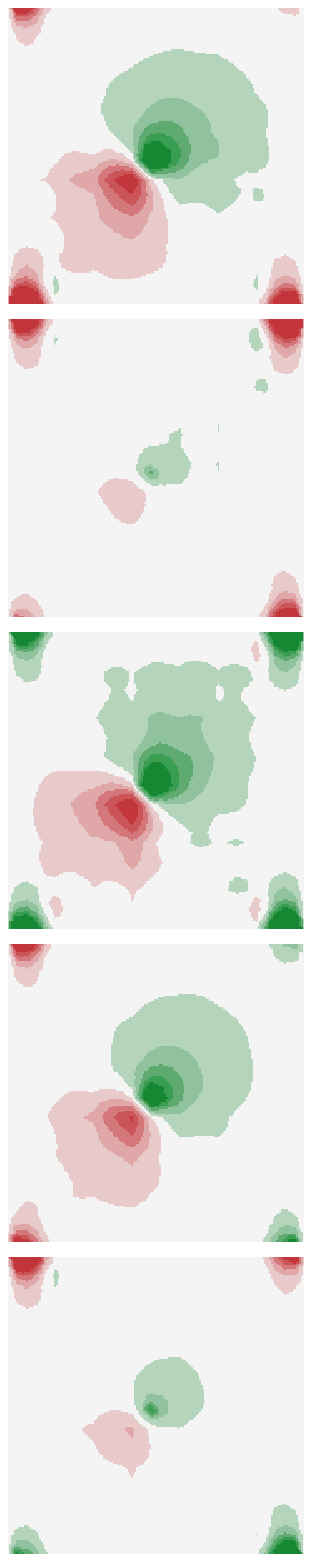} 
		& \includegraphics[width=0.05\linewidth]{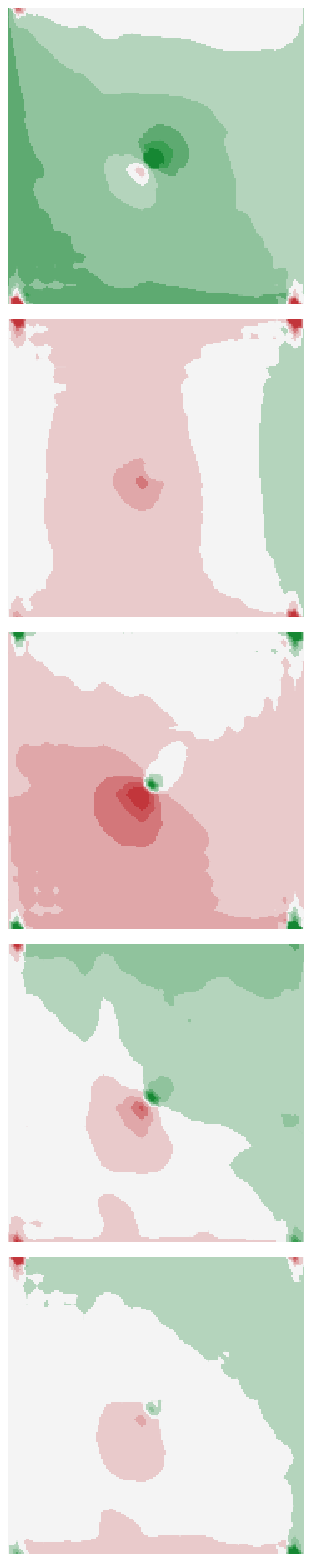} 
		& \includegraphics[width=0.05\linewidth]{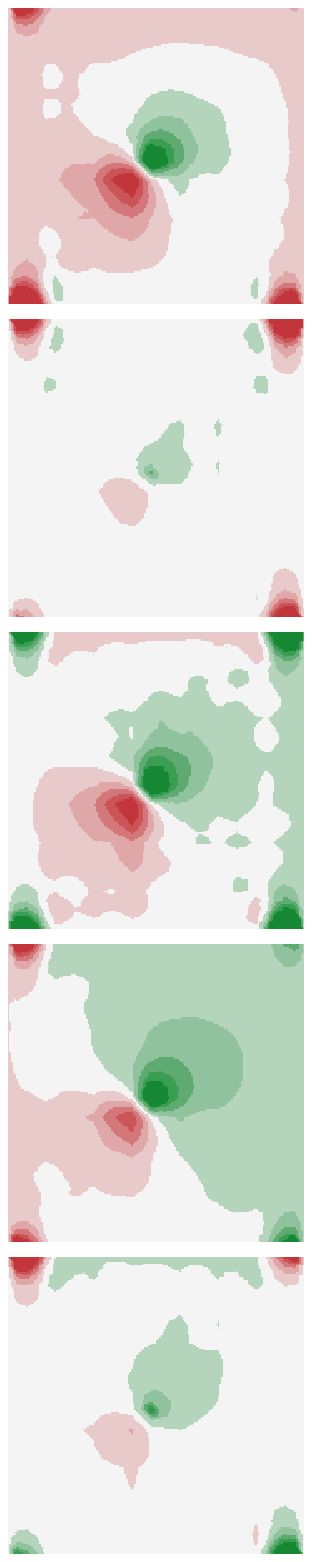} 
		& \includegraphics[width=0.05\linewidth]{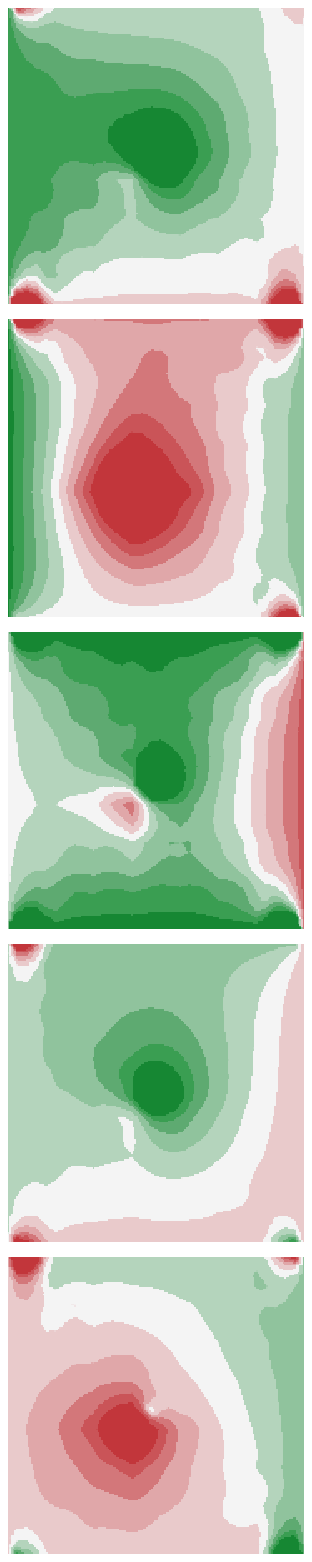}
		& \includegraphics[width=0.05\linewidth]{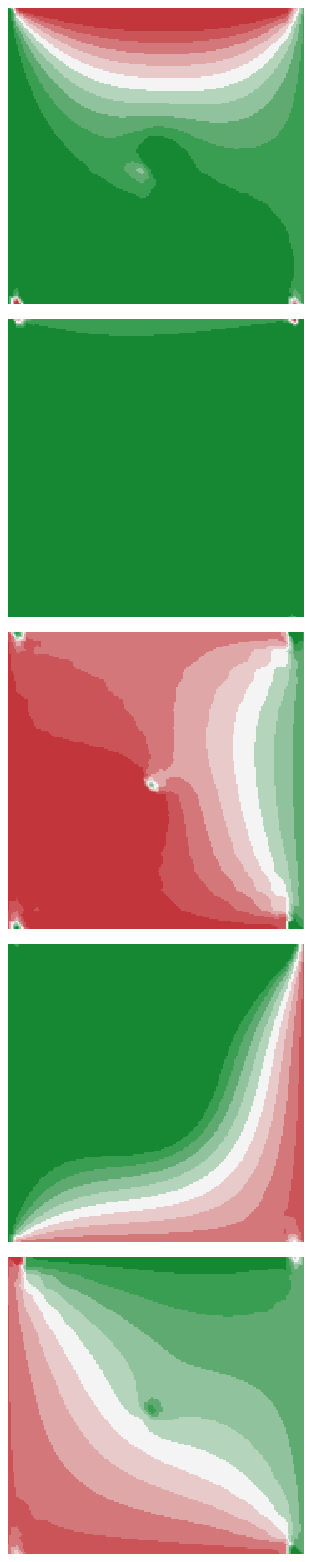}
		& \includegraphics[width=0.05\linewidth]{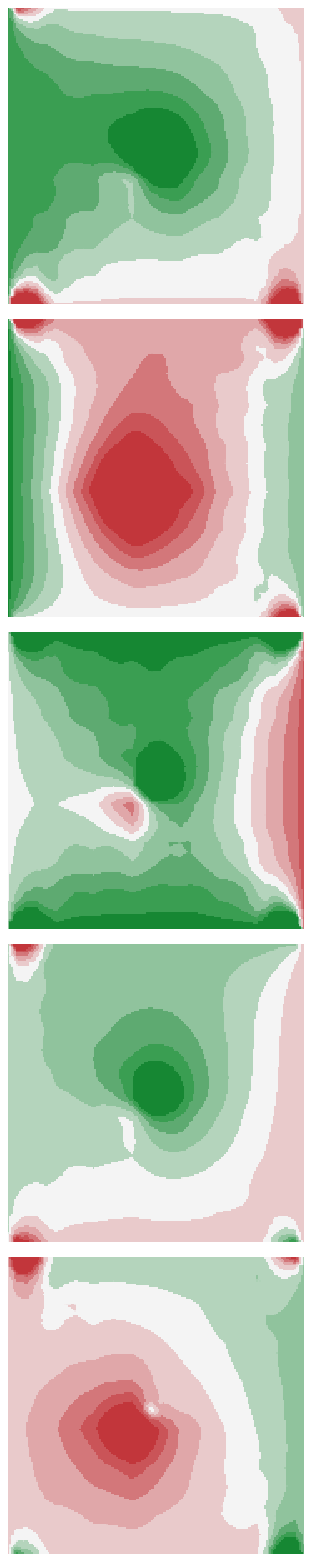}
		& \includegraphics[width=0.05\linewidth]{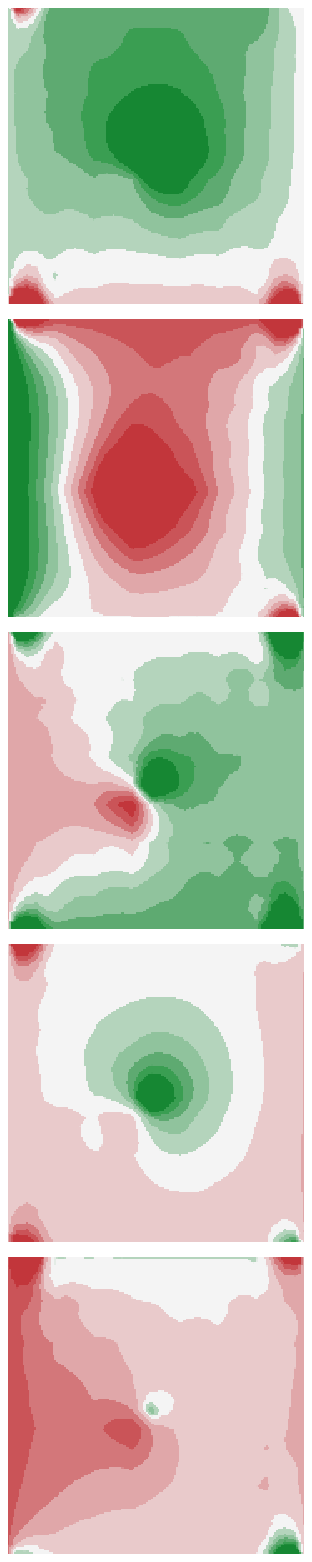}
		& \includegraphics[width=0.05\linewidth]{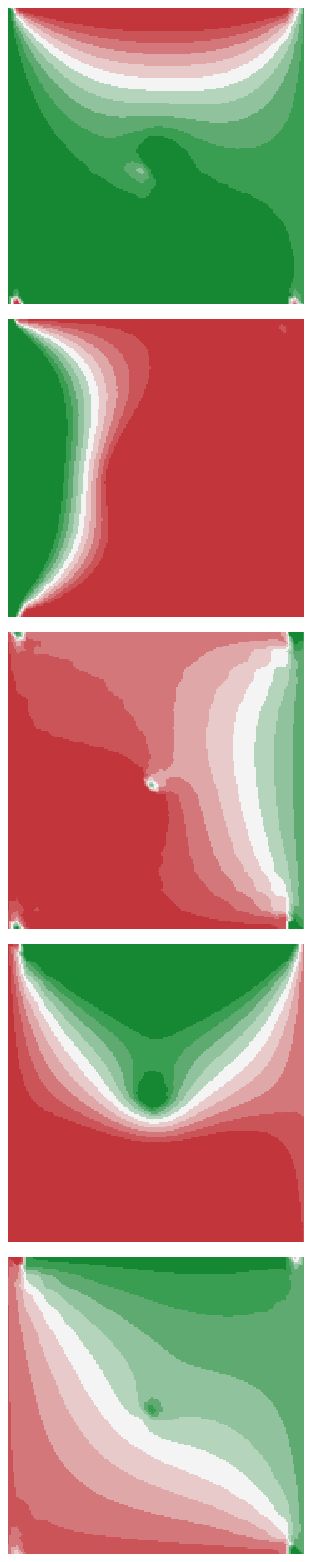}
		& \includegraphics[width=0.05\linewidth]{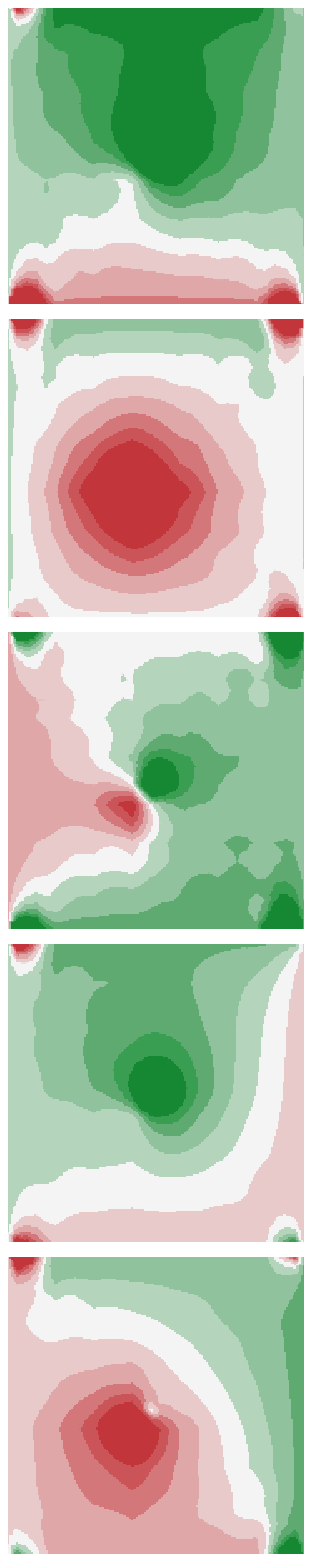}
		& \includegraphics[width=0.05\linewidth]{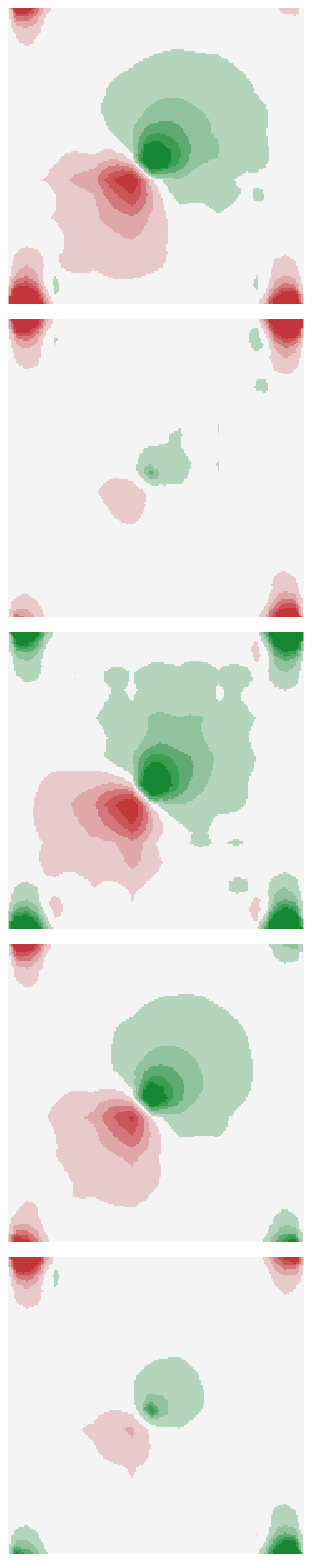}
		& \includegraphics[width=0.05\linewidth]{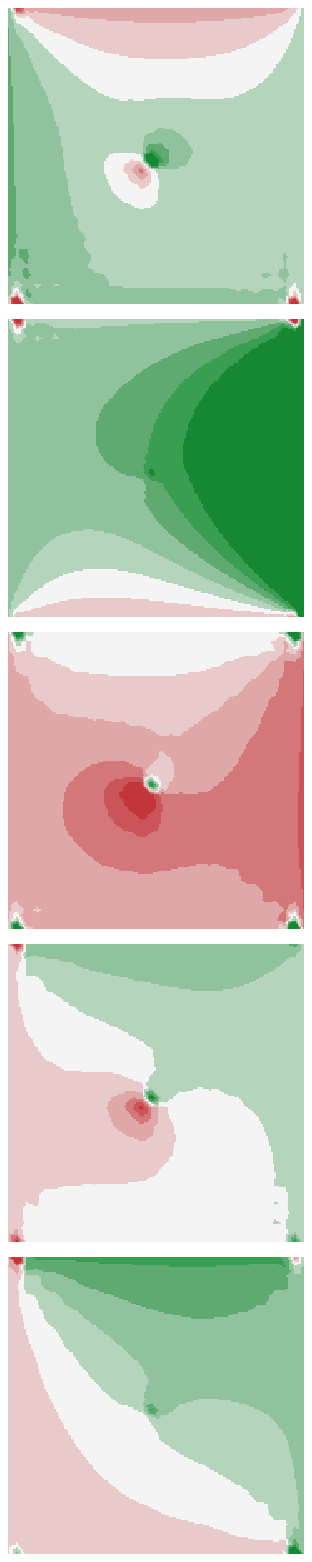}		
		& \includegraphics[width=0.05\linewidth]{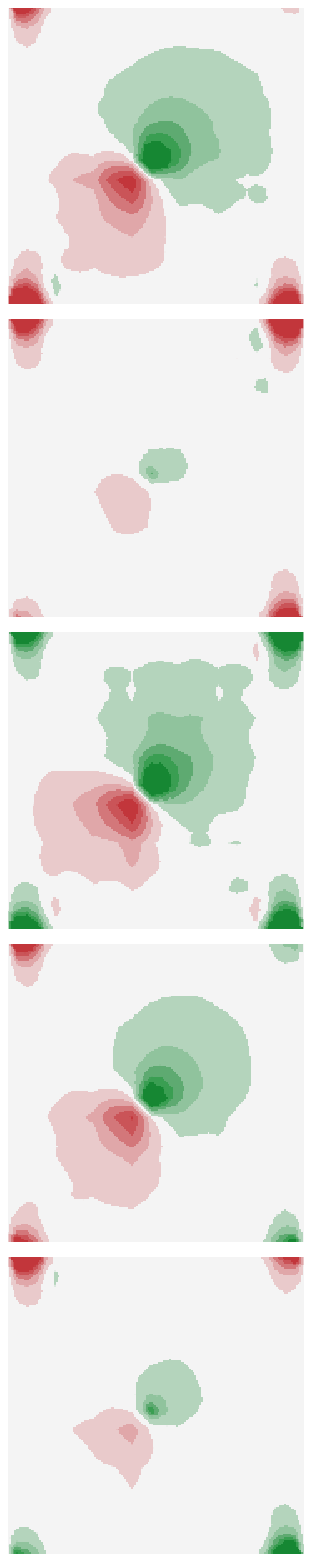}
		& \includegraphics[width=0.05\linewidth]{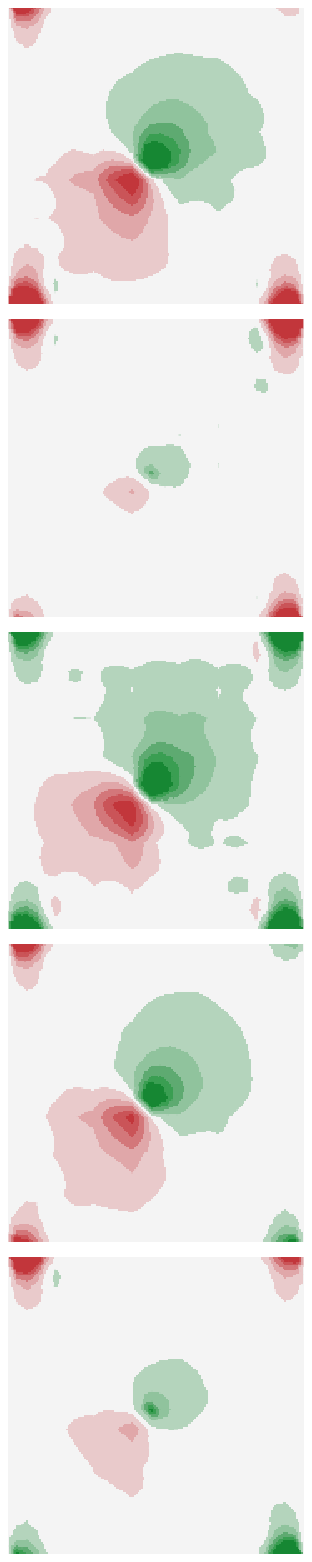}
		& \includegraphics[width=0.05\linewidth]{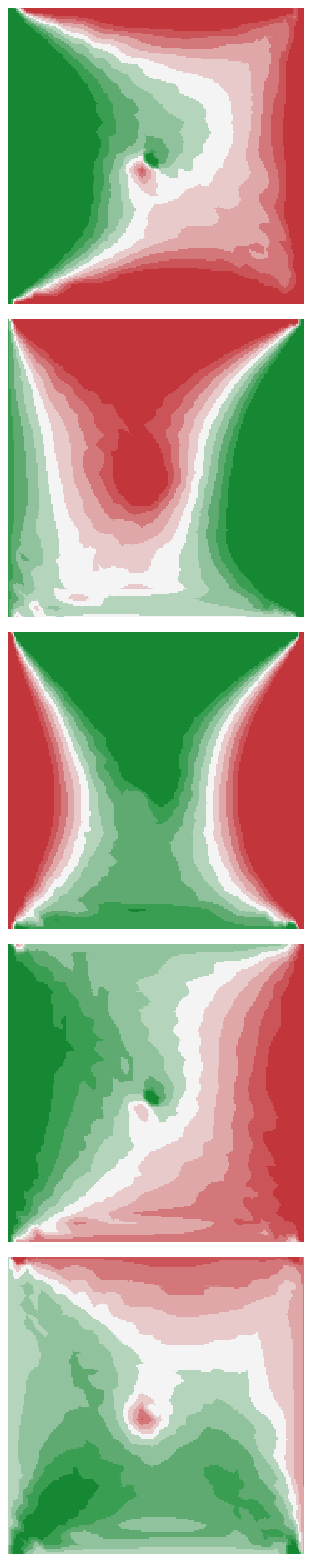}
		& \includegraphics[width=0.05\linewidth]{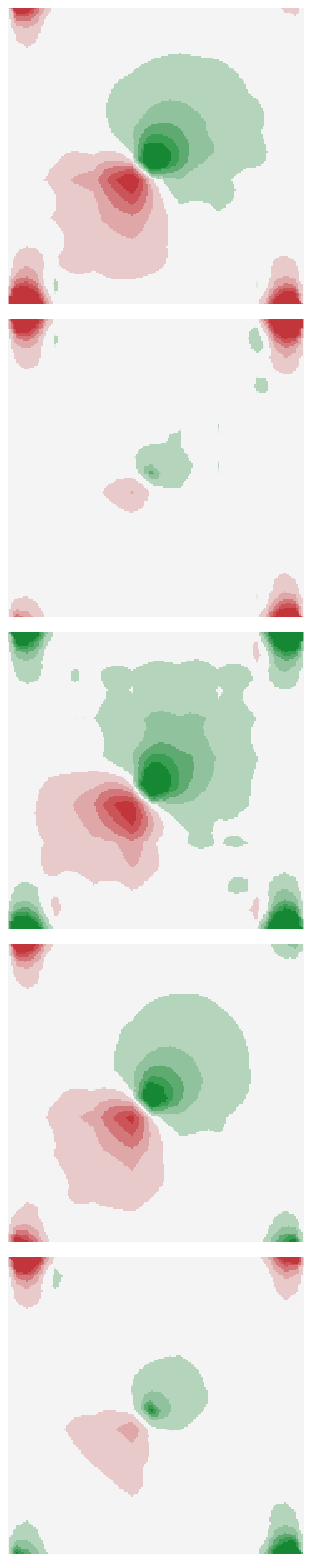}
		& \includegraphics[width=0.05\linewidth]{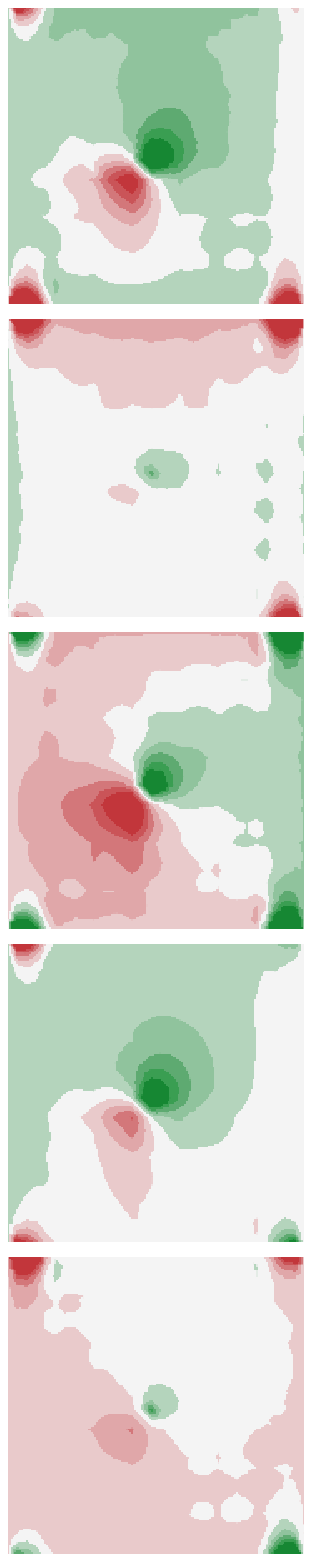}
		& \includegraphics[width=0.05\linewidth]{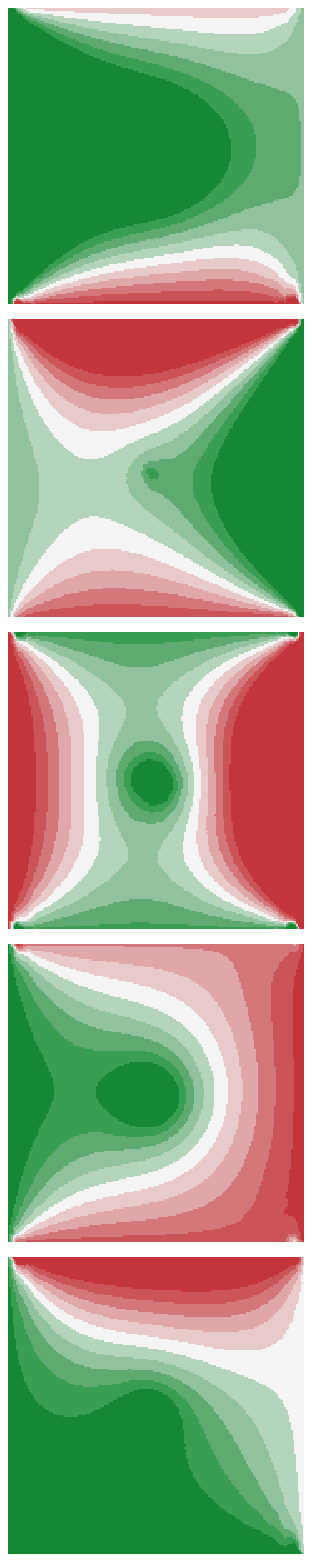}
		& \includegraphics[width=0.05\linewidth]{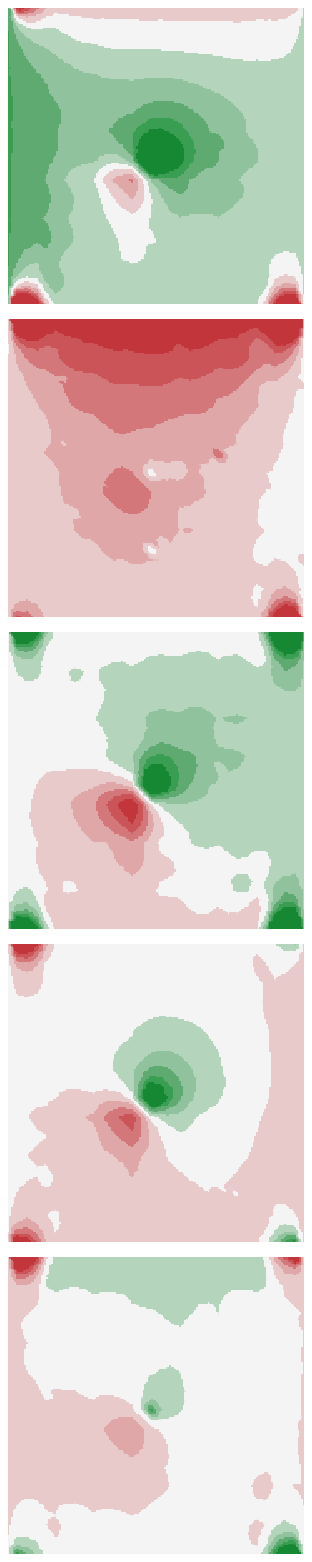}\\
		\rotatebox{265}{\scriptsize{\ours}} 
		&
		\rotatebox{265}{\scriptsize{PCA-GP-F1} }
		&
		\rotatebox{265}{\scriptsize{PCA-GP-F2}} 
		&
		\rotatebox{265}{\scriptsize{PCA-GP-ALL} }
		&
		\rotatebox{265}{\scriptsize{KPCA-GP-F1} }
		&
		\rotatebox{265}{\scriptsize{KPCA-GP-F2} }
		&
		\rotatebox{265}{\scriptsize{KPCA-GP-ALL} }
		&
		\rotatebox{265}{\scriptsize{IsoMap-GP-F1} }
		&
		\rotatebox{265}{\scriptsize{IsoMap-GP-F2} }
		&
		\rotatebox{265}{\scriptsize{IsoMap-GP-All}} 
		&
		\rotatebox{265}{\scriptsize{HOGP-F1} }
		&
		\rotatebox{265}{\scriptsize{HOGP-F2}}
		&
		\rotatebox{265}{\scriptsize{HOGP-ALL}} 
		&
		\rotatebox{265}{\scriptsize{SCGP-F1} }
		&
		\rotatebox{265}{\scriptsize{SCGP-F2} }
		&
		\rotatebox{265}{\scriptsize{SCGP-ALL}} 
		&
		\rotatebox{265}{\scriptsize{GPRN-F1} }
		&
		\rotatebox{265}{\scriptsize{GPRN-F2}}
		&
		\rotatebox{265}{\scriptsize{GPRN-ALL}}
	\end{tabular}
	\vspace{-0.15in}
	\caption{\small Visualization of local errors. Each image represents the difference between the prediction and ground-truth over individual outputs of one test example in \textit{Poisson-II}. }
	\label{fig:local}
	\vspace{-0.1in}
\end{figure*}
\begin{figure*}
	\centering
	\begin{tabular}[c]{ccc}
		\begin{subfigure}[t]{0.235\textwidth}
			\centering
			\includegraphics[width=0.99\textwidth]{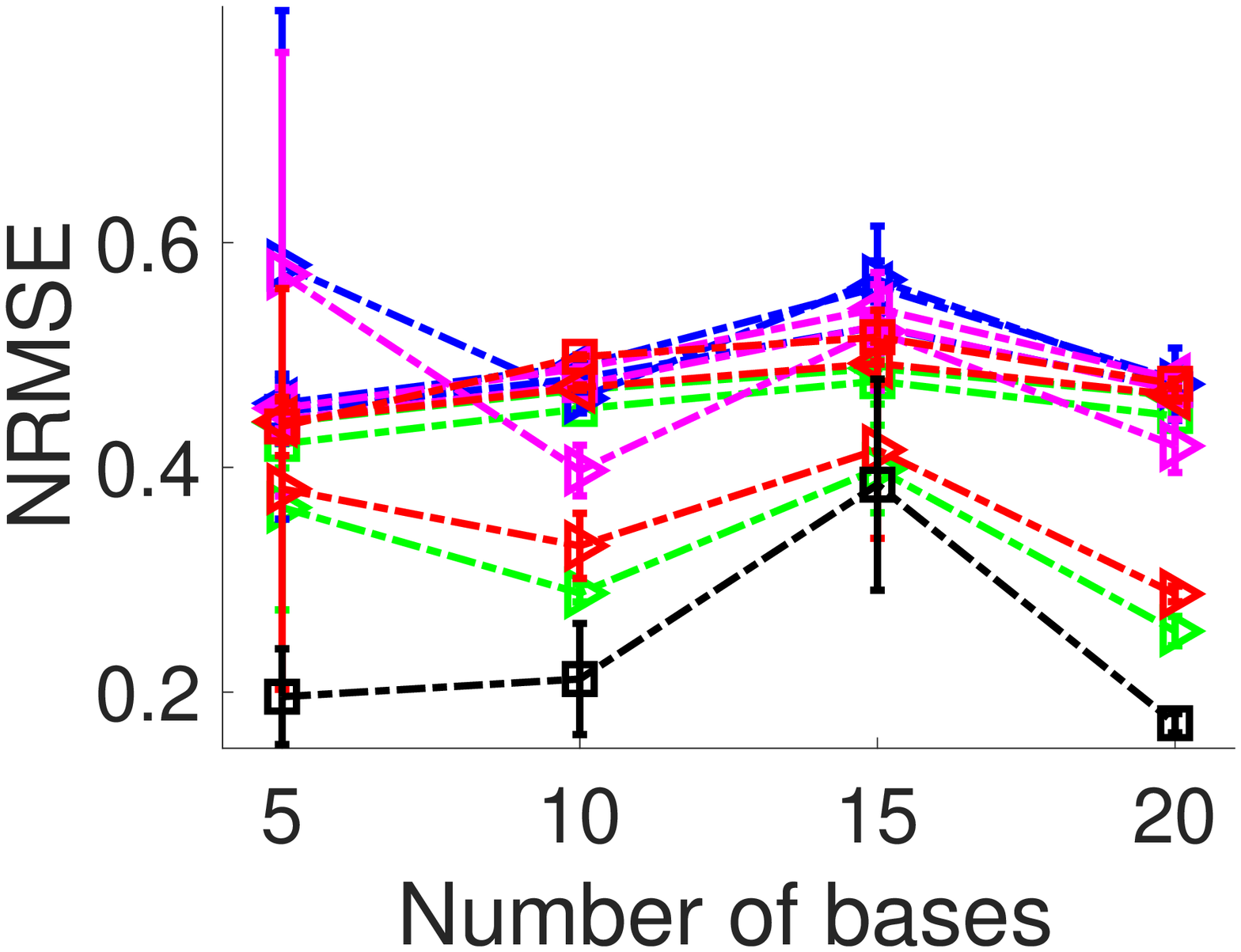}
			\caption{\textit{F1 = 120, F2 =  10}}
		\end{subfigure}
		&
		\begin{subfigure}[t]{0.235\textwidth}
			\centering
			\includegraphics[width=0.99\textwidth]{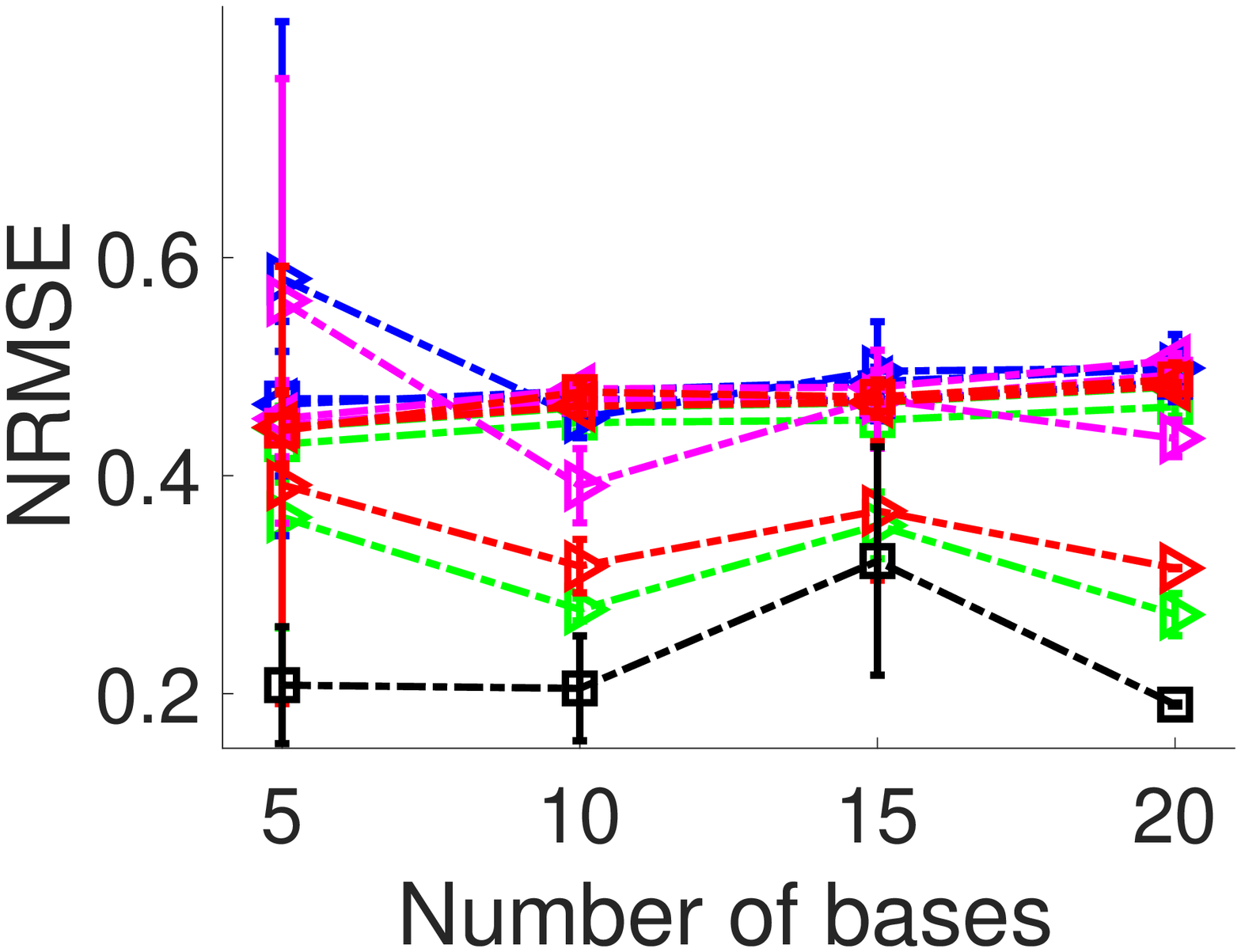}
			\caption{\textit{F1 = 160, F2 = 10}}
		\end{subfigure}
		&
		\begin{subfigure}[t]{0.235\textwidth}
			\centering
			\includegraphics[width=0.99\textwidth]{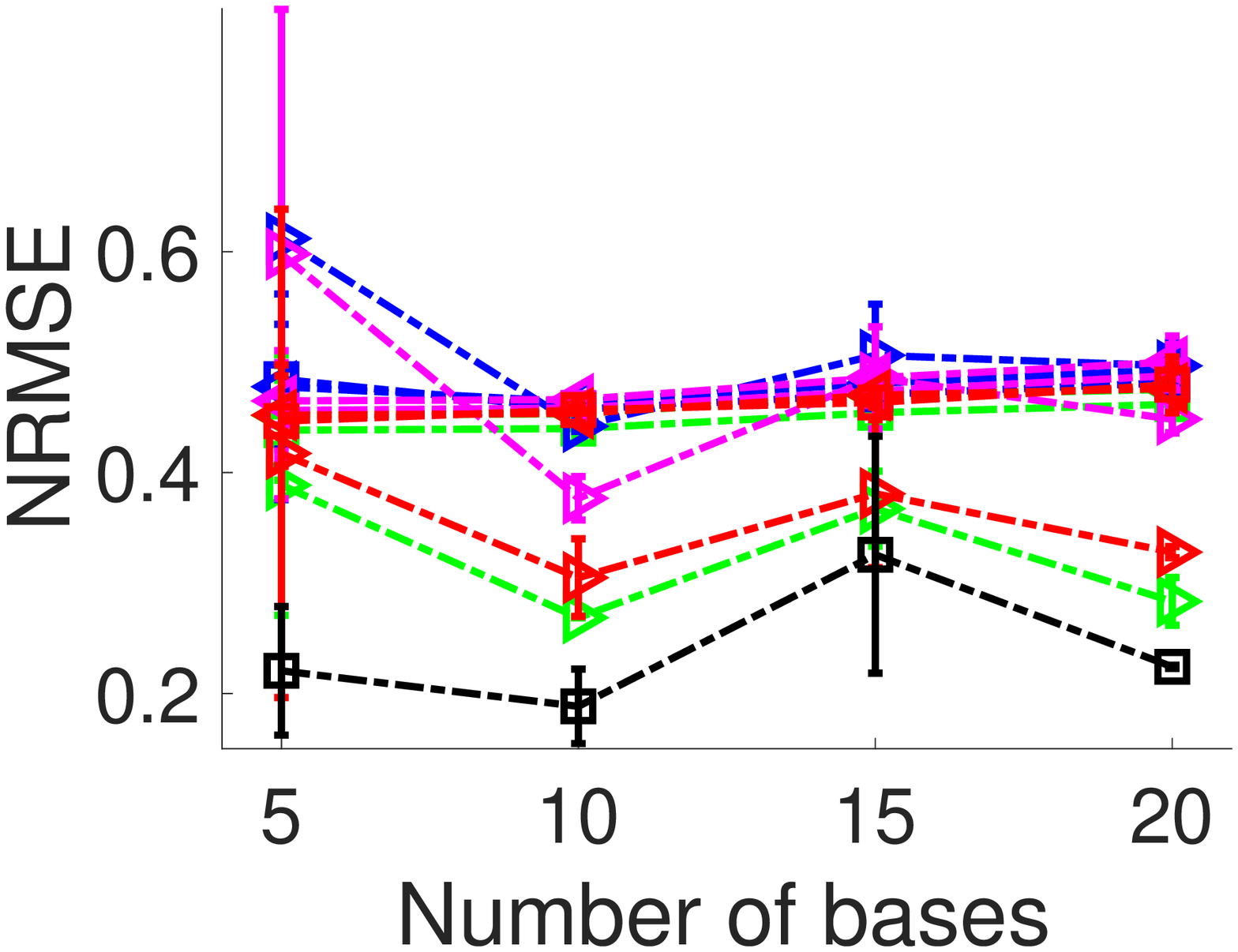}
			\caption{\textit{F1=200, F2 = 10}}
		\end{subfigure} \\
		\begin{subfigure}[t]{0.235\textwidth}
			\centering
			\includegraphics[width=0.99\textwidth]{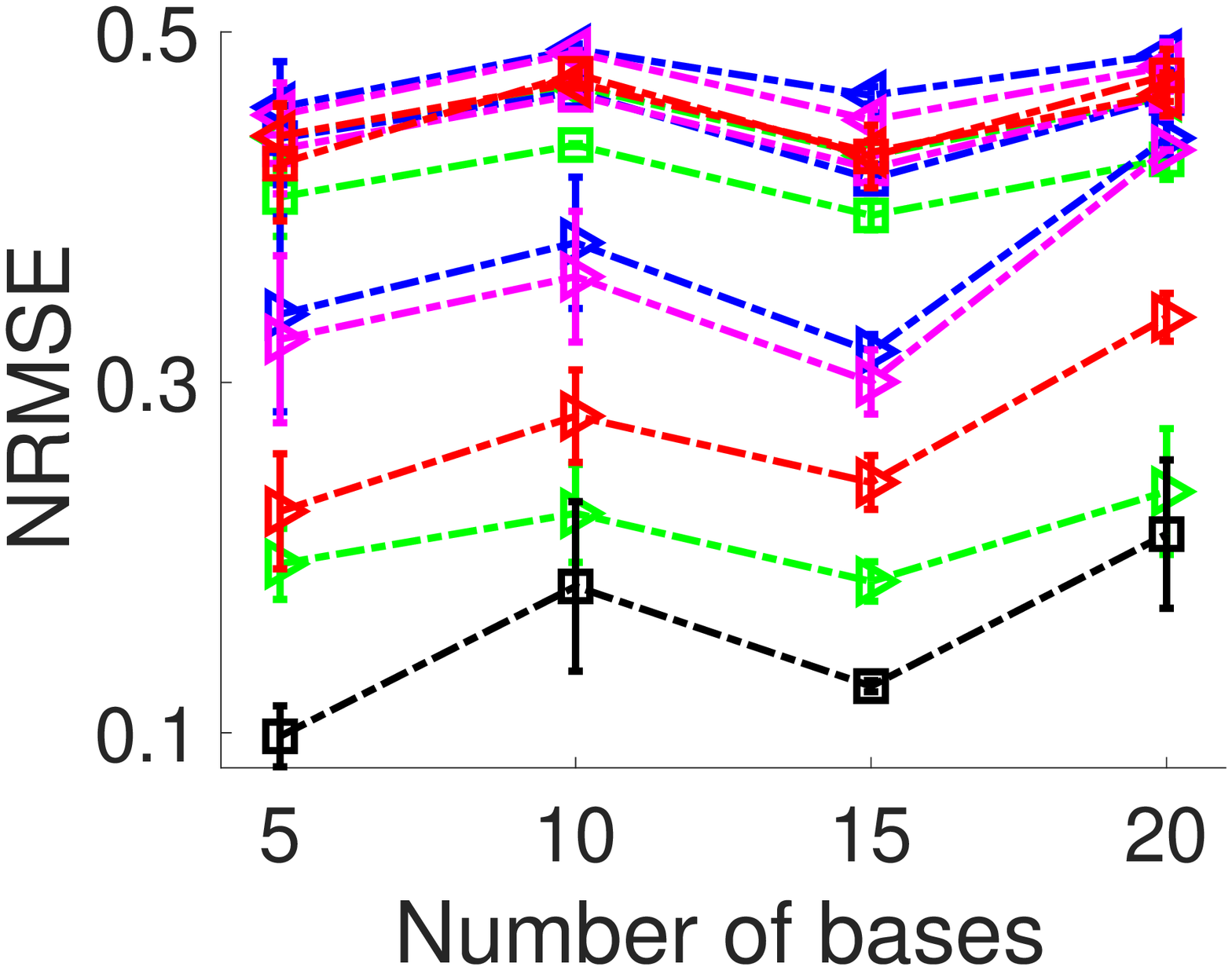}
			\caption{\textit{F1 = 120, F2 = 20}}
		\end{subfigure}
		&
		\begin{subfigure}[t]{0.235\textwidth}
			\centering
			\includegraphics[width=0.99\textwidth]{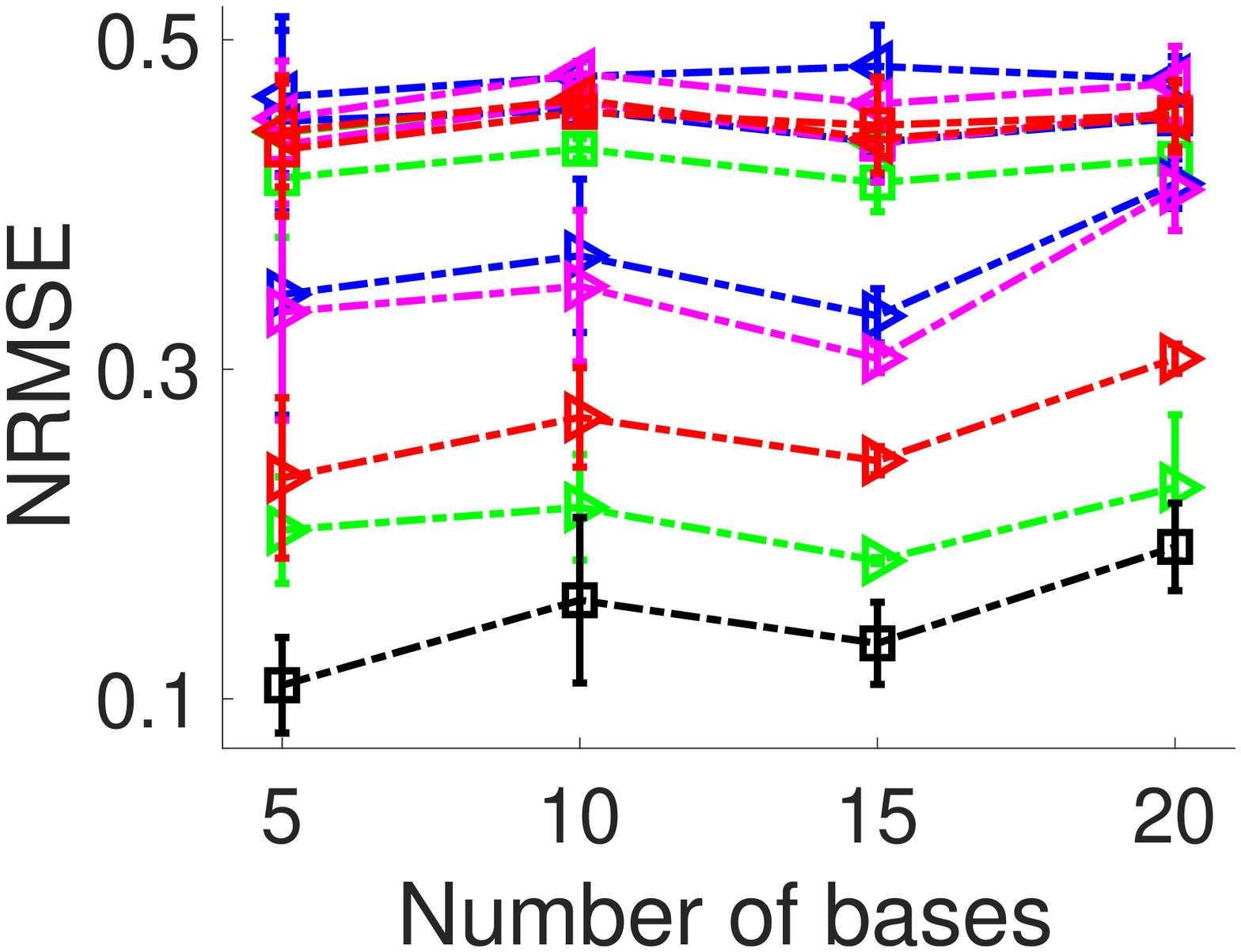}
			\caption{\textit{F1= 160, F2= 20}}
		\end{subfigure}
		&
		\begin{subfigure}[t]{0.235\textwidth}
			\centering
			\includegraphics[width=0.99\textwidth]{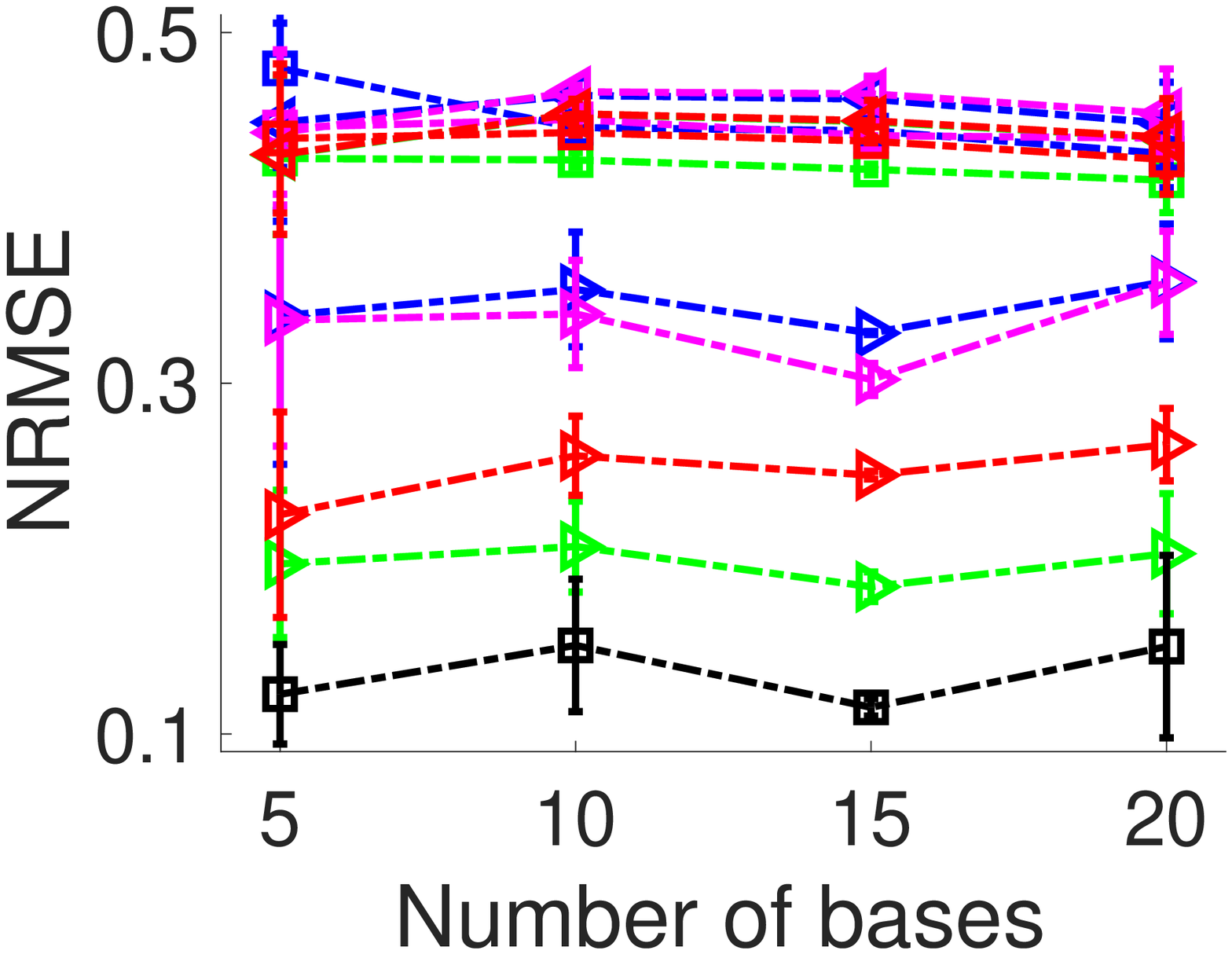}
			\caption{\textit{F1 = 200, F2 = 20}}
		\end{subfigure}
	\end{tabular}
	\vspace{-0.1in}
	\caption{\small  Normalized RMSE in predicting 1 million dimensional pressure fields of lid-driven cavity flows.  F1 and F2 are the numbers of examples with fidelity 1 and 2, respectively.  In each setting, the results are averaged from $5$ test sets.   } 
	\label{fig:large-rmse}
	\vspace{-0.3in}
\end{figure*}

We first examined \ours in predicting a relatively small number of solution outputs. These datasets were collected from solving three fundamental partial differential equations (PDEs),  Burgers', Poisson's and heat equations~\citep{olsen2011numerical} in small spatial/temporal domains. The sizes of the output fields for the three PDEs are $128 \times 128$, $32 \times 32$ and $100 \times 100$ (and so the output dimensions are $16K$, $1K$ and $10K$), respectively. In each example, the inputs are initial conditions and PDE parameters. We used numerical solvers to compute the solution field. 
The fidelity of the outputs are determined by the number of nodes/steps used in the solvers. The more the nodes/steps, the higher the fidelity. The details of the PDEs and data generation are provided in the supplementary material. For Burger's equation, we considered three training settings: (1) \textit{Burgers-I}, 400 examples of fidelity-1 (the lowest fidelity), (2) \textit{Burgers-II}, 400 fidelity-1 examples mixed with 4 fidelity-2 examples, (3) \textit{Burgers-III}, 400 fidelity-1, 40 fidelity-2 and $4$ fidelity-3 examples. 
 Similarly, we considered two training settings for Poisson's and heat equations: (4) \textit{Poisson-I} and (5) \textit{Heat-I}, 400 fidelity-1 examples, (6) \textit{Poisson-II}, 400 fidelity-1 and 10 fidelity-2 examples, and (7) \textit{Heat-II}, 400 fidelity-1 and 4 fidelity-2 examples. For each setting, we used 112 examples with one higher fidelity for testing; we randomly sampled the input parameters and  generated $5$ training and test datasets. Note that the high-fidelity samples are much less than the low-fidelity ones; the ratio ranges from $1/100$ and $1/10$. While the output dimensions are relatively small ($\sim 10^4$), the size of training data are even smaller ($\sim 10^2$).
 
 \textbf{Competing Methods}. We compared \ours with three popular LMC methods/variants for scalable multi-output regression:  (1) PCA-GP~\citep{higdon2008computer}, (2) IsoMap-GP~\citep{xing2015reduced},and (3) KPCA-GP~\citep{xing2016manifold}, which  obtain the bases or low-rank structures from Principal Component Analysis (PCA),  IsoMap~\citep{balasubramanian2002isomap} and Kernel PCA~\citep{scholkopf1998nonlinear}, respectively. In addition, we compared with (4) GPRN~\citep{wilson2012gaussian}, (5) SCGP, the sparse convolved GP~\citep{alvarez2009sparse}, and (6) HOGP, high-order Gaussian process for regression~\citep{zhe2019scalable}, a recent approach that tensorizes the outputs and can flexibly capture nonlinear output correlations and efficiently handle very high-dimensional outputs.
 
 {\textbf{Parameter settings.}} We implemented \ours with TensorFlow~\citep{abadi2016tensorflow}, and used Adam~\citep{kingma2014adam} for stochastic  optimization. In the training, we set the learning rate to $10^{-3}$ and ran Adam for $5$K epochs.  For SCGP, we used the implementation from the authors' group (\url{https://github.com/SheffieldML/multigp}). For GPRN, we tested the efficient implementation (\url{https://github.com/trungngv/gprn}) from ~\citet{nguyen2013efficient}. We used their default settings. All the other methods were implemented with Matlab and used L-BFGS for optimization. We used RBF kernel for all the methods. For each dataset,  \ours integrates the examples of all the fidelities for training. Since the competing methods are developed for single-fidelity data,  we conducted their training on the examples of each fidelity separately, and on all the examples merged together. For instance, -F1 denotes training  with the examples of fidelity-1, -F2 with fidelity-2, and -ALL with all the examples. For overlapping inputs across fidelities (See Sec. \ref{sect:model}), we preserve the higher-fidelity examples in the merged set. We varied the number of bases from $\{5, 10, 15, 20\}$, and ran all the methods on the 5 training/test datasets in each setting.
 For \ours, we decomposed the bases according to the shapes of the output fields (see Sec. \ref{sect:bases}). We computed the average root-mean-square error (RMSE) and test log likelihood, and their standard deviations of all the methods. The RMSEs are reported in Fig. \ref{fig:small-rmse}. Due to the space limit, the test log likelihoods are reported in the supplementary material.   \cmt{Note that  quite a few methods obtained very close results and so their curves are overlapping.} GPRN and SCGPR are only feasible for the smallest datasets \textit{Poisson-I} and \textit{Poisson-II} (with $\sim 1K$ outputs). For other dataset ( $\ge 10K$ outputs), they either failed with excessive memory consumption, crashed or ran forever without responses. These might be due to the  cost in estimating  a large number of GPs and complex computation in convolution kernels. 
 
 
 From Fig. \ref{fig:small-rmse}, we can see that \ours obtains the smallest prediction error in almost all the cases.  In many cases, \ours significantly outperforms the competing approaches (p-value $<$ 0.05, shown by the non-overlapping standard error bars~\citep{Minka02bar}). Note that while SCGP exhibits excellent performance on \textit{Poisson-I} and \textit{-II}, it is inefficient and cannot deal with larger numbers of outputs, \eg over 10K.  \ours exhibits superior performance in terms of the test log likelihood as well (see Fig. 1 in the supplementary material). Note that for the competing methods, simply combining all the examples of different fidelities fails to achieve an improvement. In most cases, the performance is in between only training with samples of the lowest fidelity and  higher ones  (\eg HOGP on \textit{Burgers-II} and \textit{Heat-II}, PCA-GP on \textit{Burgers-II}, \textit{Heat-II} and \textit{Burgers-III}).  Therefore, it demonstrates the effectiveness of our approach in integrating multi-fidelity examples, even the high fidelity samples take a tiny portion.  On the single-fidelity data (see Fig. \ref{fig:small-rmse}a-c), our model usually improves upon the LMC methods as well. It might because  the proposed nonlinear coregionalization more accurately captures the (nonlinear) output correlations and is less overfitting. Finally, we also examined training our model without bases decomposition: the inference is much slower and the performance is comparable or even worse. For example, in \textit{Burgers-I} setting, bases \#=15, both approaches obtain almost the same RMSE, but the bases decomposition has 3.7x speed-up.

  \vspace{-0.12in} 
 \subsection{Local Output Recovery}
 \vspace{-0.05in}
 Next, we examined how the outputs are individually recovered, \ie how the predictive performance varies locally. To this end, we randomly selected a few test samples, and visualized the difference between the prediction and ground-truth of every single output. Fig. \ref{fig:local} shows the results of $5$ test samples in \textit{Poisson-II} setting. 
As we can see, in most regions (rendered by grey), \ours achieves (almost) zero error, and only in a few small regions, it obtains small errors shown in light colors. By contrast, most competing methods result in larger errors (showed in darker colors), spreading over the vast majority of the output regions. Note that PCA-GP-F1, HOGP-F1/ALL and SCGP-F1/ALL obtained very similar local output predictions.
 In other settings, \ours exhibits better results as well. See the supplementary material for details.
 Therefore, our method not only yields a superior global accuracy (as shown in Fig. \ref{fig:small-rmse}), but locally also better recovers each individual output. 
\vspace{-0.1in}
\subsection {Large-Scale Flow Simulation}
\vspace{-0.05in}
Finally, we applied \ours in a large-scale physical simulation problem. We aimed to predict a one-million dimensional pressure field for lid-driven cavity flows~\citep{bozeman1973numerical}. When the fluid is inside a cavity and driven by a lid (or several lids) on the edge, the internal pressure can be unevenly distributed, leading to turbulent flows.  Given the boundary condition, the pressure field can be determined by solving the incompressible Navier-Stokes (NS)  equations~\citep{chorin1968numerical}, which are known to be computationally challenging. To predict the high-dimensional field, we prepared training examples of two fidelities. We varied the number of low fidelity samples from \{120, 160, 200\} and high fidelity samples from \{10, 20\}. 
For each fidelity combination, we randomly sampled the boundary conditions and simulated 5 test sets, each including $30$ examples ($3\times 10^7$ outputs). The ground-truth are computed with very dense grids in finite difference. For \ours, we decomposed each basis with three $100$ dimensional vectors. We reported the average normalized root-mean-square error (N-RMSE)  and standard deviation in Fig. \ref{fig:large-rmse}. As we can see,  our method consistently improves upon the competing methods, and in many cases significantly ($p<0.05$). Again, even combing the examples of all the fidelities, the competing methods failed to obtain improved accuracy. The results confirm the advantages of \ours in learning a function with massive outputs from very limited data with different fidelities, which is common in physical simulation. The average per-epoch/-iteration time for \ours, PCA-GP, KPCA-GP, IsoMap-GP and HOGP are $36.6$, $11.7$, $167.6$, $99.1$ and $3,417.1$ seconds, respectively (when the bases \# is $5$). 
Therefore, \ours is much faster than HOGP and has a comparable speed to the other scalable multi-output regression approaches.
\ours also exhibits smaller local errors (in recovering individual outputs). The local visualization results \cmt{and the running time of \ours} are  provided in the supplementary material.

\cmt{
\begin{figure*}
	\centering
	\setlength\tabcolsep{0.05pt}
	\begin{tabular}[c]{ccccccccc}
		\multicolumn{9}{c}{\includegraphics[width=0.3\linewidth]{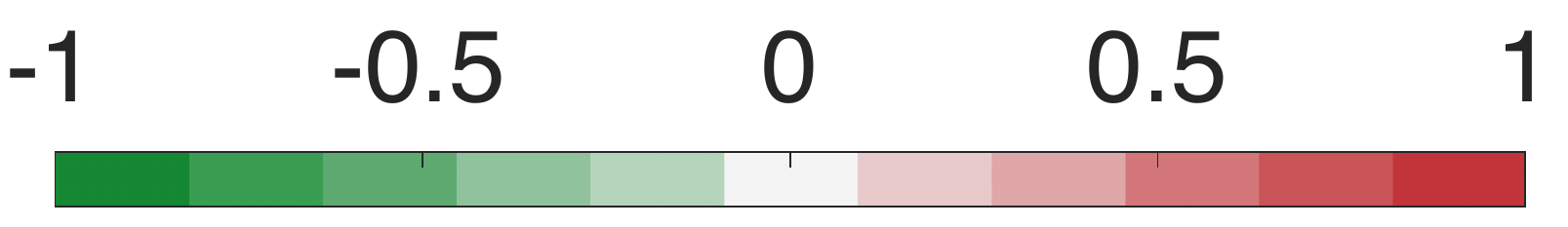}} \\
		\includegraphics[width=0.08\linewidth]{./fig/poisson_400x10/poisson_400x4_mfnc_s5.eps}
		&
		\includegraphics[width=0.08\linewidth]{./fig/poisson_400x10/poisson_400x4_pcagpf1_s5.eps} & 
		\includegraphics[width=0.08\linewidth]{./fig/poisson_400x10/poisson_400x4_pcagpf2_s5.eps} & \includegraphics[width=0.08\linewidth]{./fig/poisson_400x10/poisson_400x4_kpcaf1_s5.eps}
		&
		\includegraphics[width=0.08\linewidth]{./fig/poisson_400x10/poisson_400x4_kpcaf2_s5.eps}
		&
		\includegraphics[width=0.08\linewidth]{./fig/poisson_400x10/poisson_400x4_isomapf1_s5.eps}
		&
		\includegraphics[width=0.08\linewidth]{./fig/poisson_400x10/poisson_400x4_isomapf2_s5.eps}
		&
		\includegraphics[width=0.08\linewidth]{./fig/poisson_400x10/poisson_400x4_hogpf1_s5.eps}
		&
		\includegraphics[width=0.08\linewidth]{./fig/poisson_400x10/poisson_400x4_hogpf2_s5.eps}		
	\end{tabular}
	\caption{\small Visualization of local errors. Each image represents the difference between the prediction and ground-truth over individual outputs of a test example in \textit{Poisson}-2. From the left column to the right are the results of \ours, PCA-GP-\{1,2\}, KPCA-GP-\{1,2\}, IsoMap-GP-\{1,2\} and HOGP-\{1,2\}}. 
	\label{fig:local}
\end{figure*}
}

\vspace{-0.15in}
\section{Conclusion}
\vspace{-0.1in}
We have presented \ours, a multi-fidelity high-order GP model for physical simulation. In the future, we will explore \ours in other domains, such as multi-resolution large-scale sensor networks output prediction. We will further extend  \ours for multi-fidelity Bayesian optimization~\citep{song2019general} and active learning for complex system optimization and design problems.  

\bibliographystyle{apalike}
\bibliography{MFHOGP}

\section*{Supplementary Materials}
\section{Graphical Model Representation}
\begin{figure}[!htb]
	\centering
	\includegraphics[width=\linewidth]{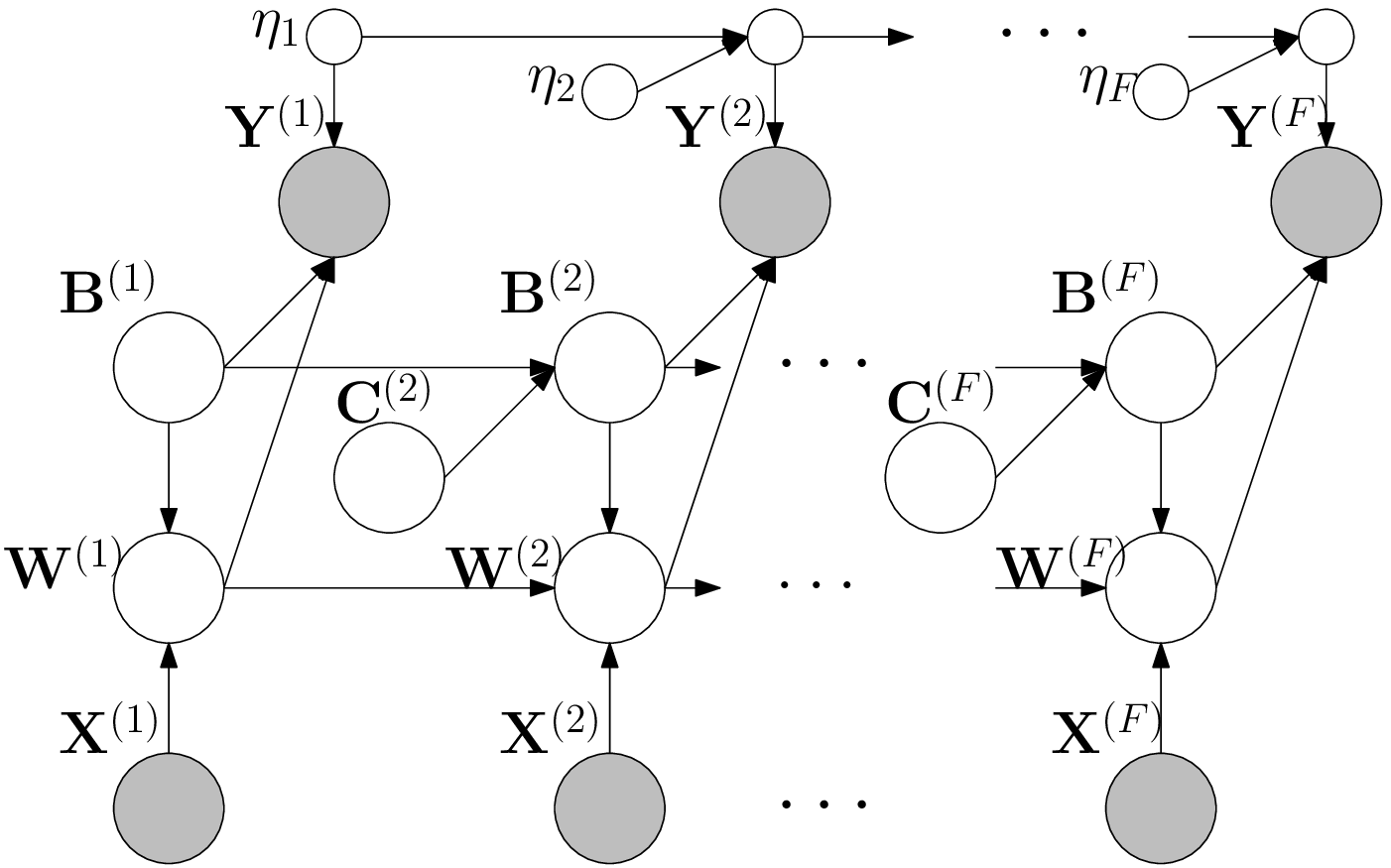}
	\caption{Graphical representation of MFHoGP. Note that $\{\W^{(j)}\}_{j=1}^F$ is sampled from a deep matrix GP prior. The joint probability is given in Eq. (4) of the main paper.} 
	\label{fig:grpahical_model}
\end{figure}

To facilitate illustration, we provide the graphical representation of our model in Fig. \ref{fig:grpahical_model}. 
\section{Proof of Lemma 3.1}
\noindent \textbf{Lemma 3.1.} In the proposed nonlinear coregionalization model, the marginal distribution of the output matrix $\Y$ is 
\[
p(\Y|\X,\B) = \N\big(\vec(\Y)|\0,  (\B^\top \K_{BB}\B)\otimes \K + \eta^{-1}\I\big).
\]
Given two arbitrary outputs $y_m(\x_i)$ and $y_t(\x_j)$, \ie the $m$-th output for input $\x_i$ and $t$-th output for input $\x_j$, \cmt{their covariance is} we have $\mathrm{cov}\big(y_m(\x_i), y_t(\x_j)\big) = k(\x_i, \x_j) \tilde{b}_m^\top  \K_{BB}  \tilde{b}_t + \eta^{-1}\cdot \mathbbm{1}(\x_i = \x_j, m=t)$, where $\tilde{b}_m$ and $\tilde{b}_t$ are the $m$-th and $t$-th column of $\B$, respectively, and $\mathbbm{1}(\cdot)$ is the indicator function. 

\begin{proof}
	First, from the likelihood $p(\Y|\W, \B) = \N\big(\vec(\Y)|\vec(\W\B), \eta^{-1}\I \big)$, we can obtain that $\vec(\Y) = \vec(\W\B) + \bepsilon$,
	where $\bepsilon \sim \N(\bepsilon |\0, \eta^{-1}\I)$. Using the property of vectorization~\citep{minka2000old}, we can derive that $\vec(\W\B) = (\B^\top \otimes \I)\vec(\W)$, and hence
	 \[
	 \mathrm{cov}\big(\vec(\W\B)\big) = (\B^\top \otimes \I) \mathrm{cov}\big(\vec(\W)\big) 
	 (\B \otimes \I).
	 \]
	  Since $p(\W|\X, \B) = \MN(\W|\0, \K, \K_{BB}) = \N\big(\vec(\W)|\0, \K_{BB}\otimes \K\big)$, we have $\mathrm{cov}(\vec(\W))= \K_{BB} \otimes \K$. Therefore
	\[
	\mathrm{cov}\big(\vec(\W\B)\big)  = (\B^\top \otimes \I) (\K_{BB} \otimes \K)  (\B \otimes \I) = (\B^\top \K_{BB} \B)\otimes \K
	\]
	and 
	\[
	\mathrm{cov}\big(\vec(\Y)\big) = (\B^\top \K_{BB} \B)\otimes \K+ \eta^{-1}\I.
	\]
	Finally, since $\vec(\Y)$ is an affine transformation of $\vec(\W)$ plus an independent Gaussian noise, it must follow a multivariate Gaussian distribution. Obviously, $\EE\big(\vec(\Y)\big) = \0$. Finally, we have 
	\[
	p(\Y|\X,\B) = \N\big(\vec(\Y)|\0,  (\B^\top \K_{BB}\B)\otimes \K + \eta^{-1}\I\big).
	\]
\end{proof}

\section{Experimental Details}
\subsection{Data Preparation for Small Solution Fields}
As mentioned in our main paper, the small datasets were collected from solving three fundamental partial differential equations (PDEs),  Burgers', Poisson's and heat equations, in small spatial/temporal domains. Each equation plays an important role in scientific and engineering applications. The details of the PDEs and data generation are listed as follows.

\noindent \textbf{Burgers' equation} is considered as a canonical nonlinear hyperbolic PDE; it is widely used to describe various physical phenomena, such as fluid dynamics~\citep{chung2010computational}, nonlinear acoustics~\citep{sugimoto1991burgers} and traffic flows~\citep{nagel1996particle}.  Because it can develop  discontinuities  (shock waves) based on a normal conservation equation, it also serves as a benchmark test case for many numerical solvers and surrogate models~\citep{kutluay1999numerical,shah2017reduced,raissi2017physics}.
The viscous version of this equation is given by $\frac{\partial u}{\partial t} + u \frac{\partial u}{\partial x} = v \frac{\partial^2 u}{\partial x^2},$ where $u$ represents the volume, $x$ indicates a spatial location, $t$ denotes the time, and $v$ represents the viscosity. We set $x\in[0,1]$\cmt{(in $\si{\meter}$)}, $t \in [0,3]$\cmt{ (in $\si{\second}$)}, and $u(x,0)=\sin(x\pi/2)$ with homogeneous Dirichlet boundary conditions.  We uniformly sampled viscosities $v \in [0.001,0.1]$\cmt{ (in $ \si[inter-unit-product = \ensuremath{{}\cdot{}}] {\milli\pascal\second}$)} as the input parameter to generate the solution field. 
The equation is solved using the finite element with hat functions in space and backward Euler in time domains. The spatial-temporal domain is discretized into a $16\times16$ regular rectangular mesh for the first (lowest) fidelity solver. The subsequent solvers of higher fidelities double the nodes in each mesh dimension, \eg $32\times32$ for the second fidelity and $64\times64$ for the third fidelity.
The result fields (\ie outputs) are computed from a $128\times128$ spatial-temporal regular mesh.

\noindent \textbf{Poisson's equation} is an elliptic PDE commonly used in mechanical engineering and physics to describe potential fields, \eg gravitational and electrostatic fields~\citep{chapra2010numerical}. It is a generalization of Laplace's equation~\citep{persides1973laplace} and written as $\Delta u = 0$, where $\Delta $ is the Laplace operator and $u$ indicates the volume.
Despite its simplicity, Poisson's equation is frequently seen in physics and often serves as a basic test case for surrogate models~\citep{lagaris1998artificial,tuo2014surrogate}.
In our experiment, we set a 2D spatial domain $\textbf{x} \in [0,1] \times [0,1]$ with Dirichlet boundary conditions. The constant values of the four boundaries and the centre of the rectangle domain are used as the input parameters, each of which ranges from $0.1$ to $0.9$. 
We uniformly sampled the input parameters to generate the corresponding potential fields as the outputs. The PDE is solved using the finite difference method with the first order centre differencing scheme and regular rectangle meshes. We used an $8\times8$ mesh for the coarsest level solver. The subsequent refined solver uses a finer mesh that doubles the node in each dimension. The result potential fields are computed with  a $32\times32$ spatial-temporal regular grid.

\noindent \textbf{Heat equation} is a basic PDE that describes how heat flows evolve over time. Although originally introduced in 1822 to explain heat flows only, the heat equation is ubiquitous in many scientific fields, such as probability theory~\citep{spitzer1964electrostatic,burdzy2004heat} and financial mathematics~\citep{black1973pricing}. 
Hence, it is also widely used as a surrogate model  ~\citep{efe2003proper,raissi2017machine}.
The heat equation is given by $\frac{\partial u}{\partial t} + \alpha \Delta u =0 $, where $u$ represents the heat, $\alpha$ the thermal conductivity, and $\Delta$ the Laplace operator.
We set a 2D spatial-temporal domain $x\in[0,1]$, $t \in [0,5]$ with the Neumann boundary condition at $x=0$ and $x=1$, and $u(x,0)=H(x-0.25)-H(x-0.75)$, where $H(\cdot)$ is the Heaviside step function. 
The input parameters include the flux rate of the left boundary at $x=0$ (ranging from 0 to 1), the flux rate  of the right boundary at $x=1$ (ranging from  -1 to 0), and the thermal conductivity (ranging from 0.01 to 0.1).
The equation is solved using finite difference in space and backward Euler in time domains. The spatial-temporal domain is discretized into a $16\times16$ regular rectangular mesh for the first (lowest) fidelity solver. A refined solver uses a $32\times32$ mesh for the second fidelity. The result fields are computed on a $100\times100$ spatial-temporal grid.

\subsection{Data Preparation for Large-Scale Simulations of Lid-Driven Cavity Flows}
We also examined \ours in lid-driven cavity flows~\citep{bozeman1973numerical}, a classic computational fluid dynamics problem. The problem describes how liquid inside a cavity is driven by the lids on the walls, making the pressures vary locally and eventually leading to laminar and turbulent flows inside the cavity.
The simulation of lid driven cavity flows involves solving the incompressible Navier-Stokes (NS) equation~\citep{chorin1968numerical},  $\rho (\textbf{u} \cdot \nabla) \textbf{u} = -\nabla p + \mu \nabla^2 \textbf{u}$, where $\rho$ is the density, $p$ the pressure, $\textbf{u}$ the velocity, and $\mu$ the dynamic viscosity. 
The PDE is well known to be challenging to solve due to their complicated behaviours under large Reynolds numbers.
It is thus commonly used as a benchmark test case for numerical solvers~\citep{bozeman1973numerical,strang1973analysis} and surrogate models~\citep{terragni2011local,xing2016manifold}.
In our experiments, we considered a square cavity $\textbf{x} \in [0,1] \times [0,1]$ filled with liquid and the time $t\in[0,1]$. The top lid is given a tangential velocity to drive the fluid to flow while the other lids on the remaining walls stay steady. No-slip conditions are applied to all the lids. Given the Reynold number $Re \in [10,500]$ and the top boundary velocity ranging from 0 to 1, the spatial-temporal pressure field is computed on a $100\times100$ regular mesh at $100$ evenly spaced time points. Hence, we have one million outputs for each input setting. 
We used the SIMPLE algorithm~\citep{caretto1973two} with a stagger grid~\citep{versteeg2007introduction}, the up-wind scheme~\citep{versteeg2007introduction} for the spatial difference, and the implicit time scheme with fixed time steps to solve the PDE. 
For the lowest fidelity solver, we used a $16\times16$ spatial mesh and $10,000$ time step to ensure the numerical stability.
A subsequent finer solver (of the second fidelity) uses a $32 \times 32$ spatial mesh and the same number of time steps.

\subsection{Test Log Likelihood}
\begin{figure*}
	\centering
	\setlength\tabcolsep{0.01pt}
	\begin{tabular}[c]{cccc}
		\raisebox{0.04in}
		{
			\begin{subfigure}[t]{0.23\textwidth}
				\centering
				\includegraphics[width=\textwidth]{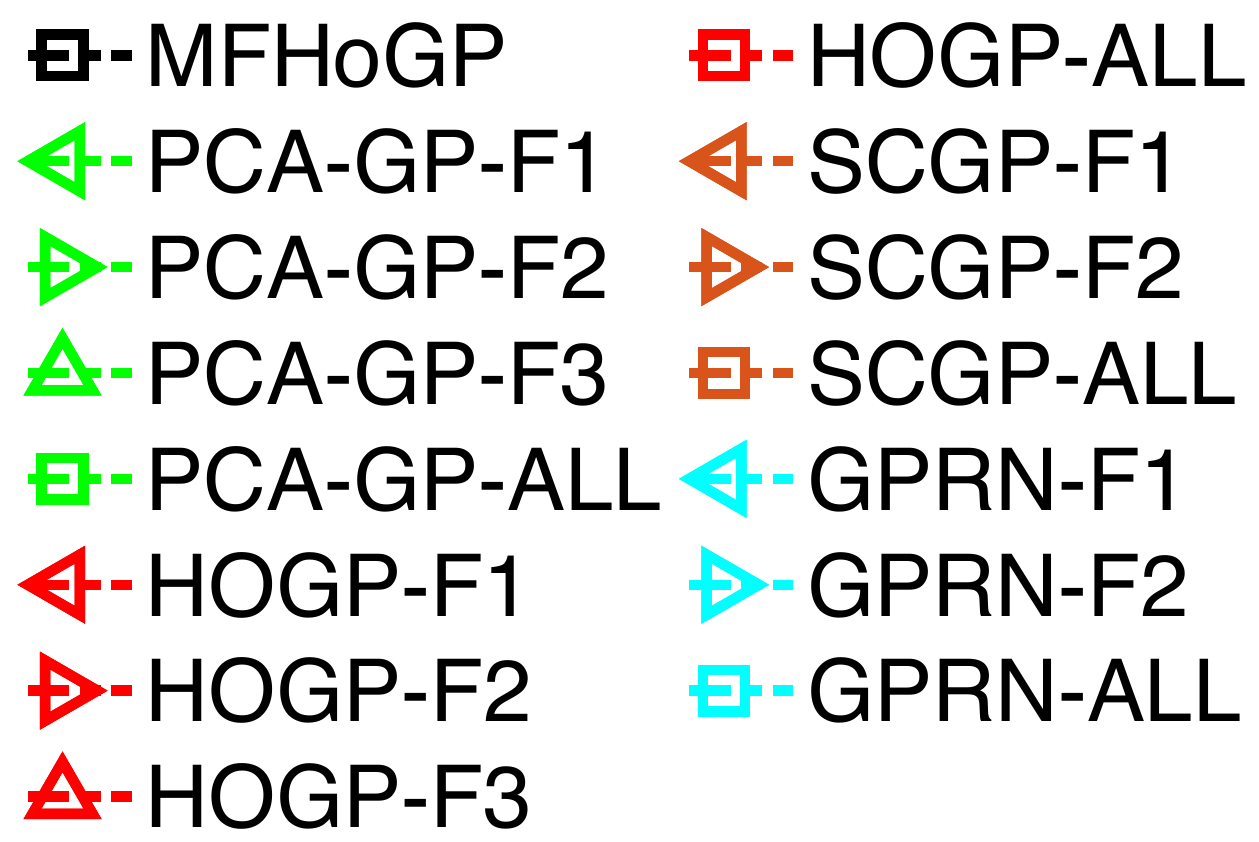}
			\end{subfigure}
		}
		&
		\begin{subfigure}[t]{0.23\textwidth}
			\centering
			\includegraphics[width=\textwidth]{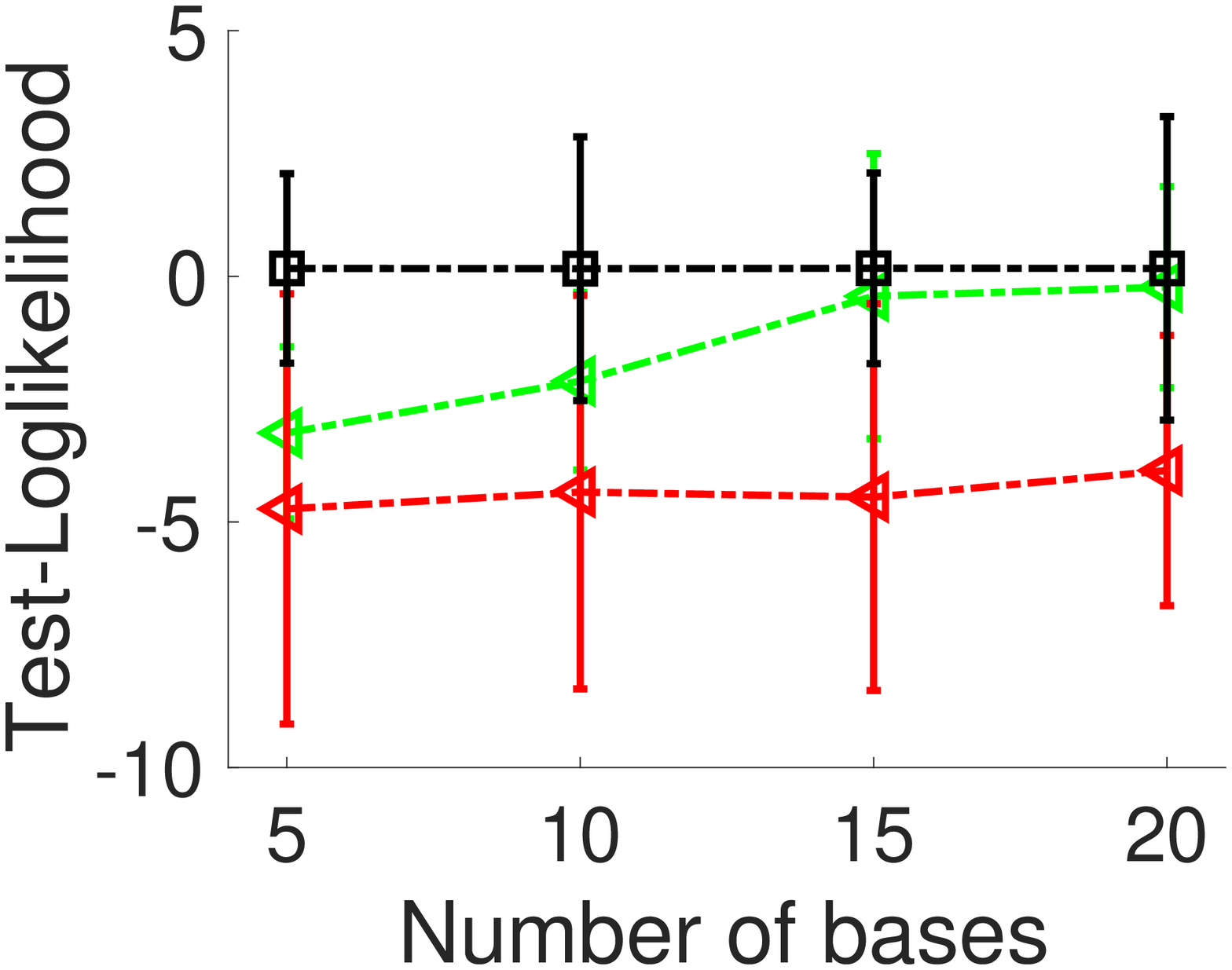}
			\caption{\textit{Burgers-I}}
		\end{subfigure} 
		&
		\begin{subfigure}[t]{0.23\textwidth}
			\centering
			\includegraphics[width=\textwidth]{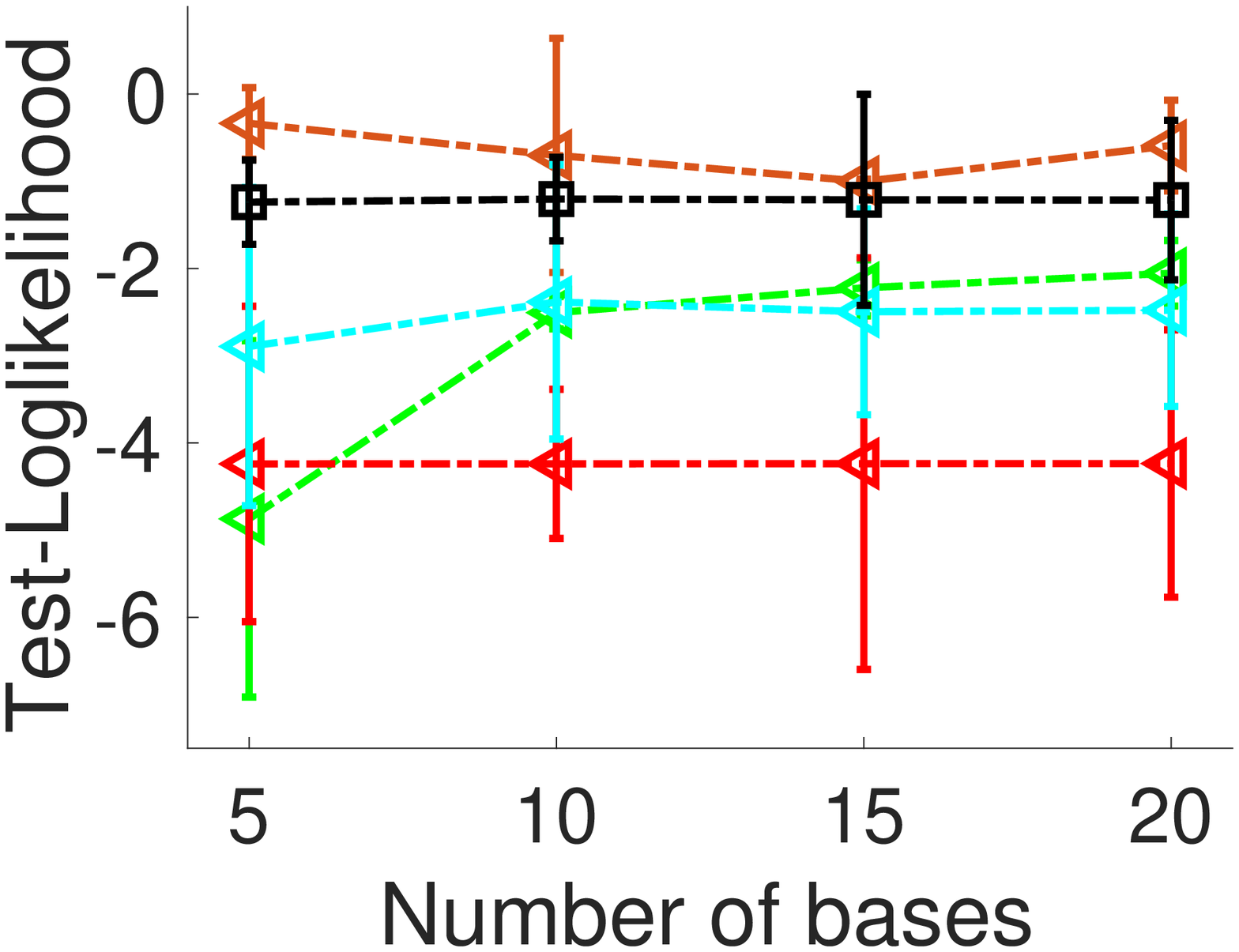}
			\caption{\textit{Poisson-I}}
		\end{subfigure}
		&
		\begin{subfigure}[t]{0.23\textwidth}
			\centering
			\includegraphics[width=\textwidth]{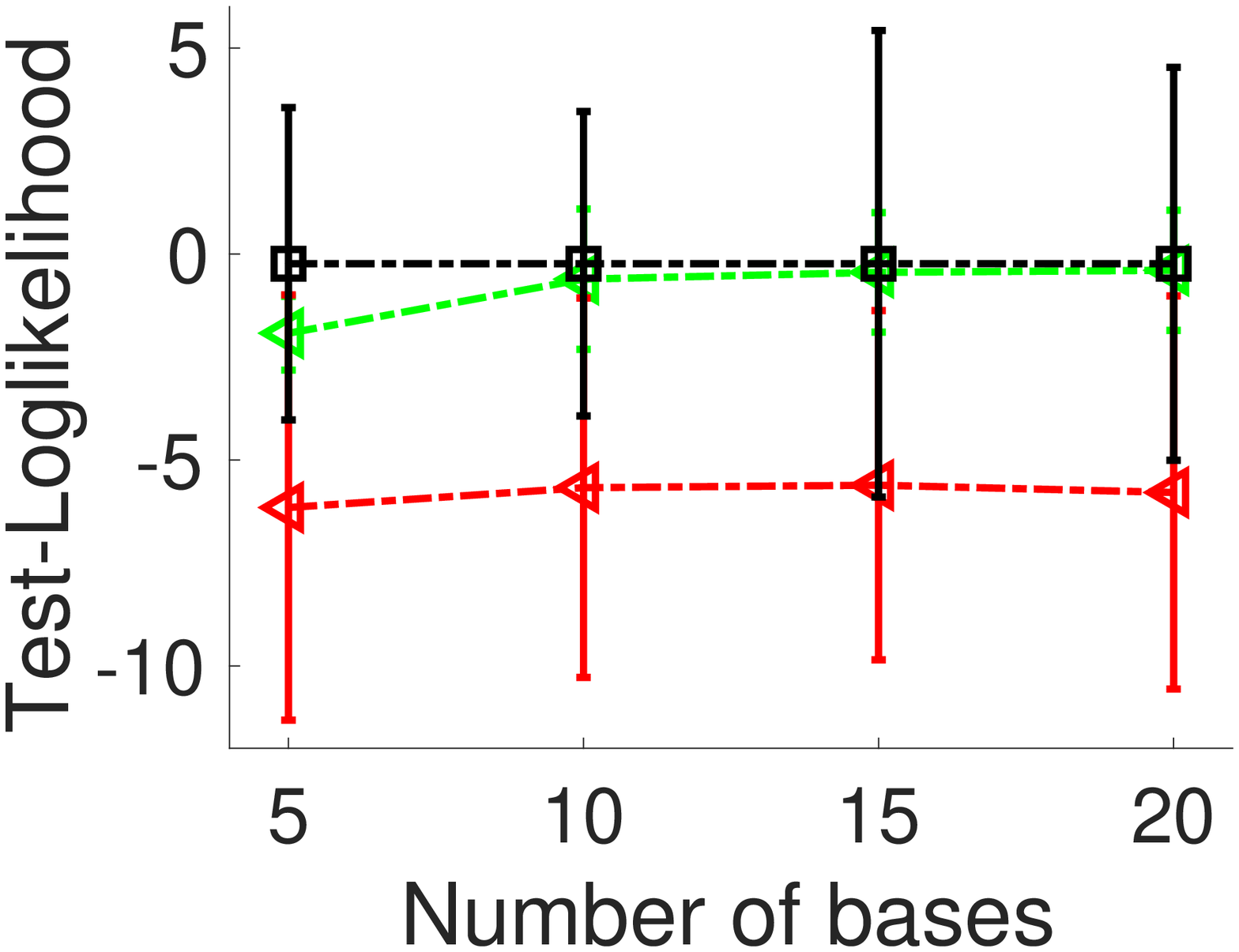}
			\caption{\textit{Heat-I}}
		\end{subfigure}\\
		\begin{subfigure}[t]{0.23\textwidth}
			\centering
			\includegraphics[width=\textwidth]{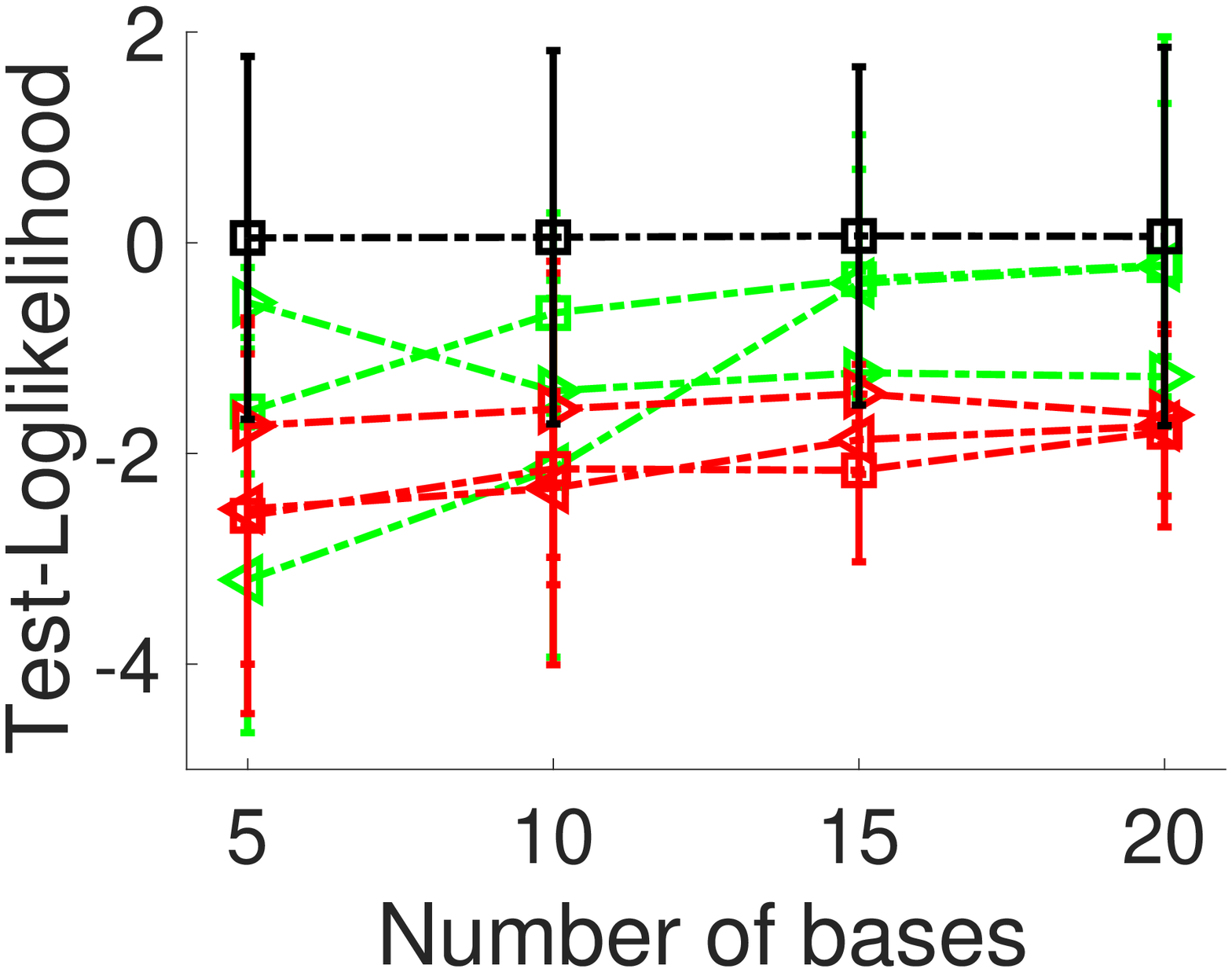}
			\caption{\textit{Burgers-II}}
		\end{subfigure} 
		&
		\begin{subfigure}[t]{0.23\textwidth}
			\centering
			\includegraphics[width=\textwidth]{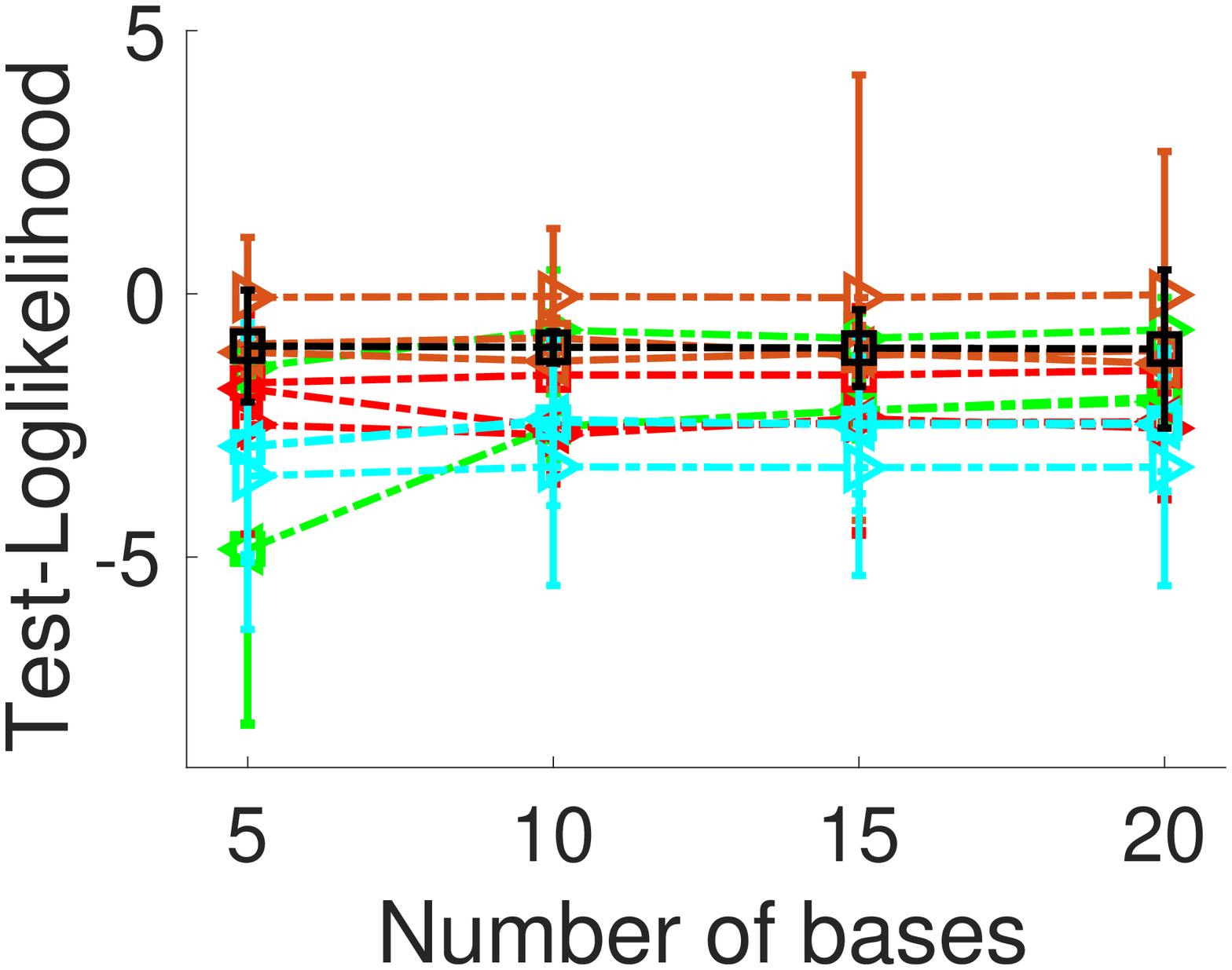}
			\caption{\textit{Poisson-II}}
		\end{subfigure}
		&
		\begin{subfigure}[t]{0.23\textwidth}
			\centering
			\includegraphics[width=\textwidth]{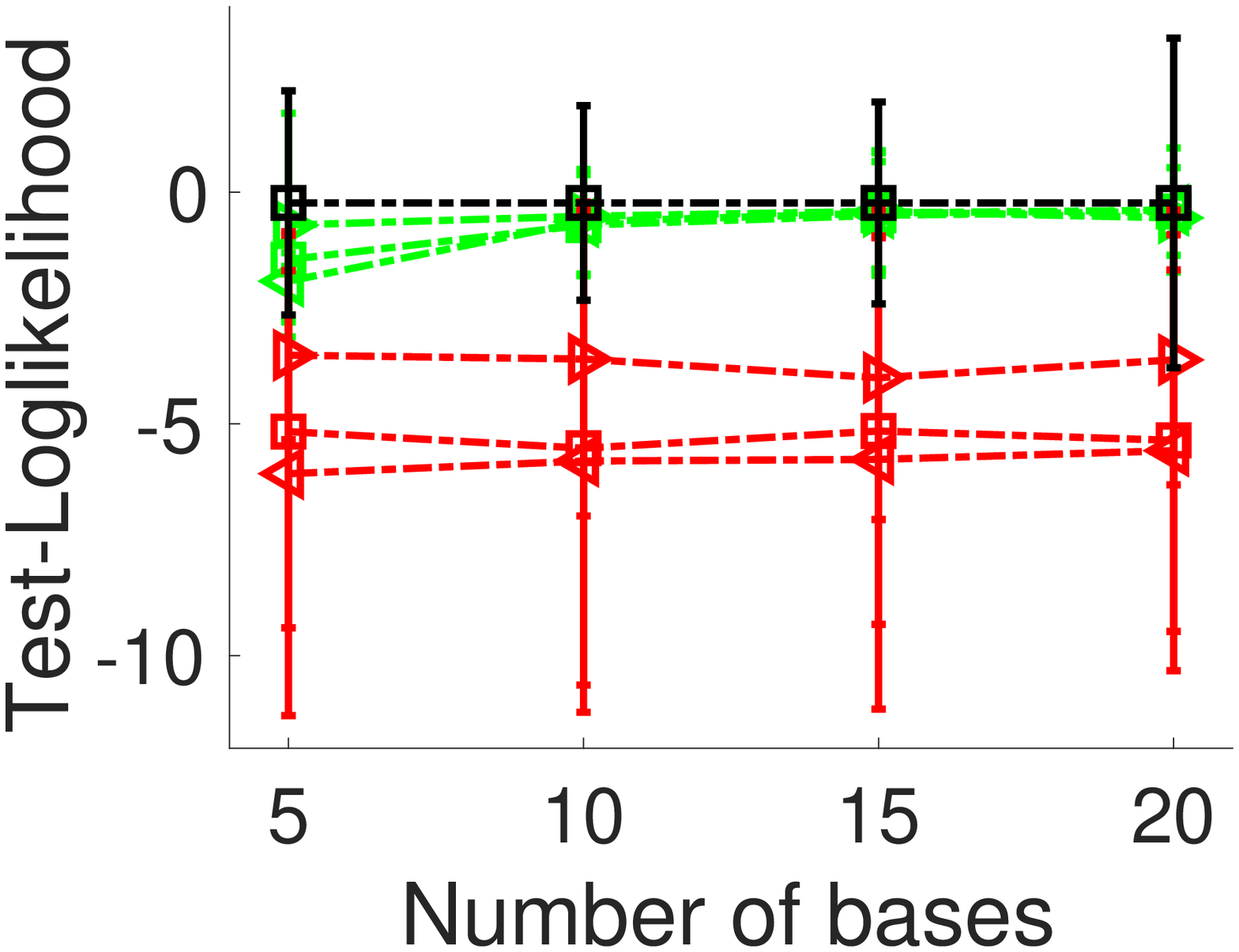}
			\caption{\textit{Heat-II}}
		\end{subfigure}
		&
		\begin{subfigure}[t]{0.23\textwidth}
			\centering
			\includegraphics[width=\textwidth]{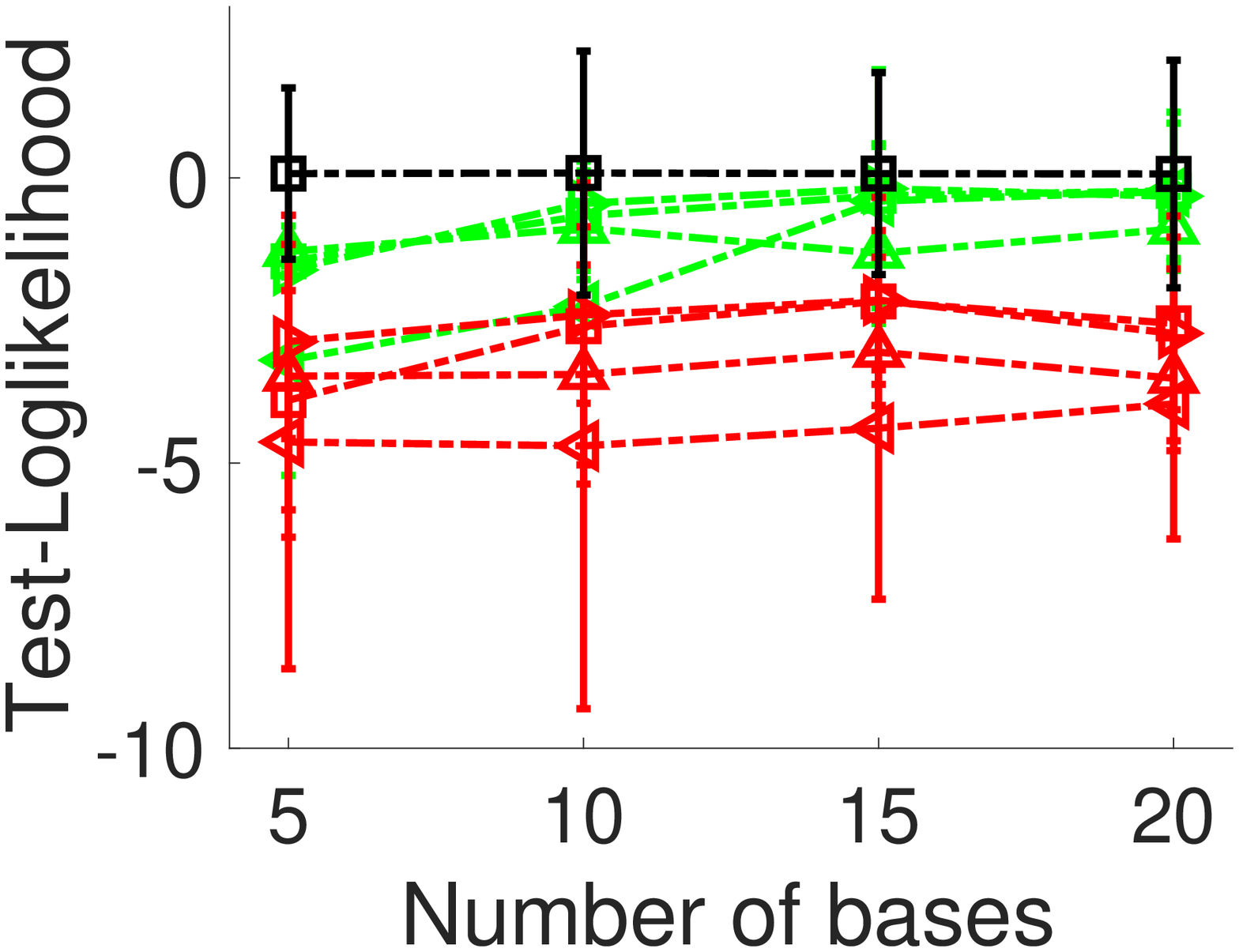}
			\caption{\textit{Burgers-III}}
		\end{subfigure}
	\end{tabular}
	\vspace{-0.1in}
	\caption{\small The test-loglikelihood (in log scale) on small datasets. The results are averaged from 5 runs. After the dash in each caption (\eg ``\textit{Burgers-II}'') is how many fidelities across the training data. -F\{1,2,3\} indicates the model trained with a particular fidelity's examples and -F-ALL  with all the examples.} 	
	\label{fig:small-ll}
	\vspace{-0.14in}
\end{figure*}

We report the test loglikelihood of  PCA-GP, HOGP,  SCGP, GPRN and \ours on small datasets in Fig.\ref{fig:small-ll}. Since KPCA-GP and IsoMap-GP are not standard probabilistic models (they do not have likelihood terms), their test loglikelihoods are unavailable.  As we can see, similar to the RMSE results (in the main paper), \ours consistently outperforms all the competing methods, except on  \textit{Poisson-I} and  \textit{Poisson-II}, SCGP obtains slightly higher likelihood. However, SCGP is inefficient and cannot handle a large number of outputs, \eg over ten thousands. 

\subsection{Local Output Recovery}
\begin{figure*}
	\centering
	\setlength\tabcolsep{0.05pt}
	\begin{tabular}[c]{ccccccccc}
		\multicolumn{9}{c}{\includegraphics[width=0.3\linewidth]{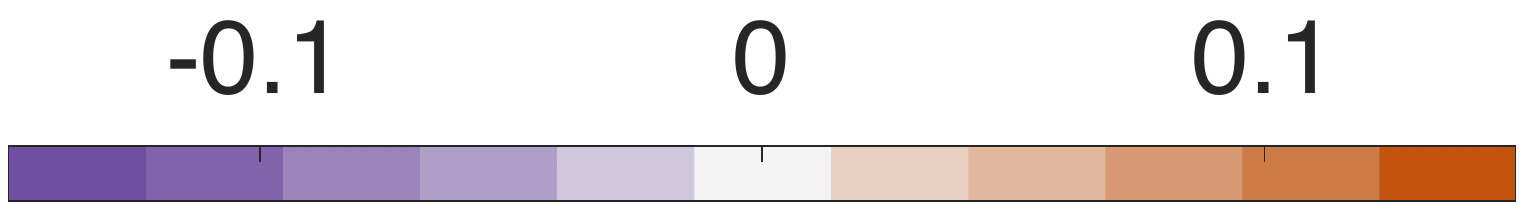}} \\
		\includegraphics[width=0.08\linewidth]{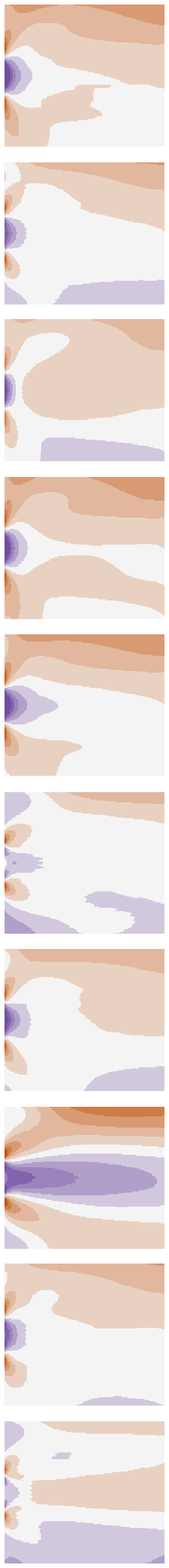}
		&
		\includegraphics[width=0.08\linewidth]{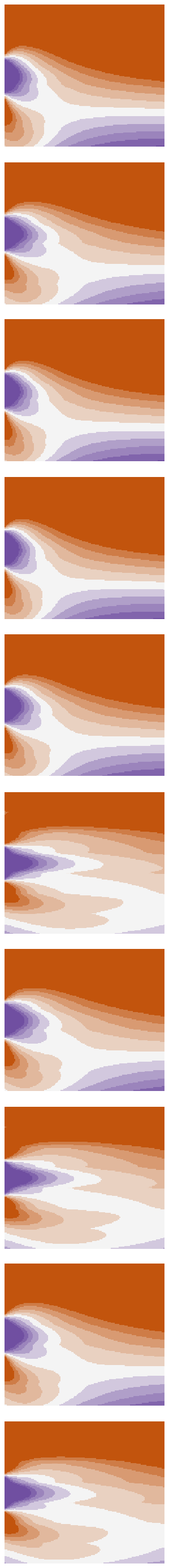} & 
		\includegraphics[width=0.08\linewidth]{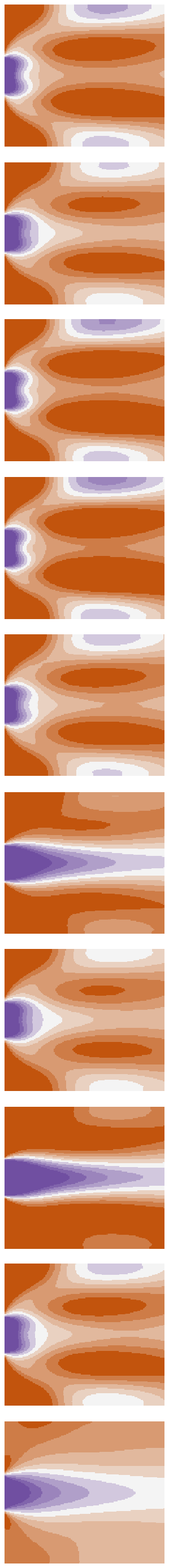} 
		&
		 \includegraphics[width=0.08\linewidth]{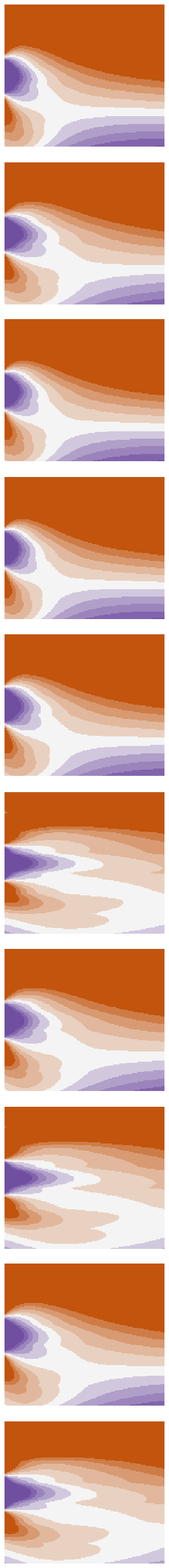}
		&
		\includegraphics[width=0.08\linewidth]{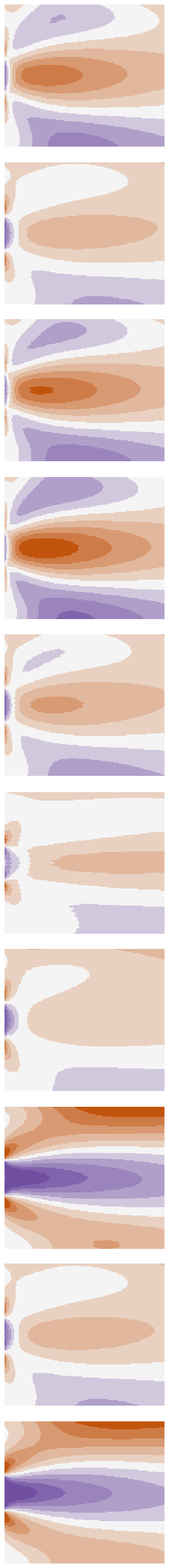}
		&
		\includegraphics[width=0.08\linewidth]{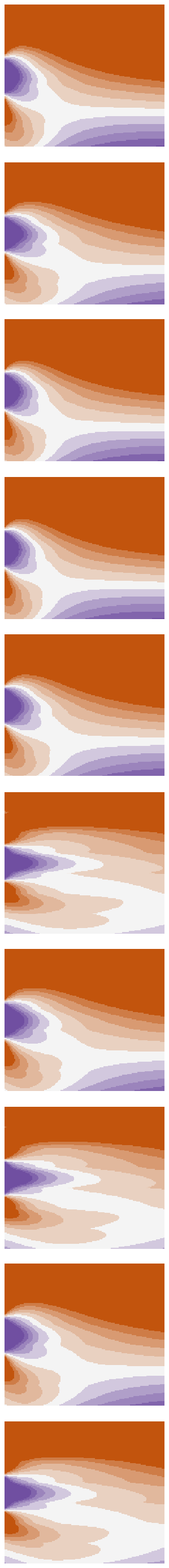}
		&
		\includegraphics[width=0.08\linewidth]{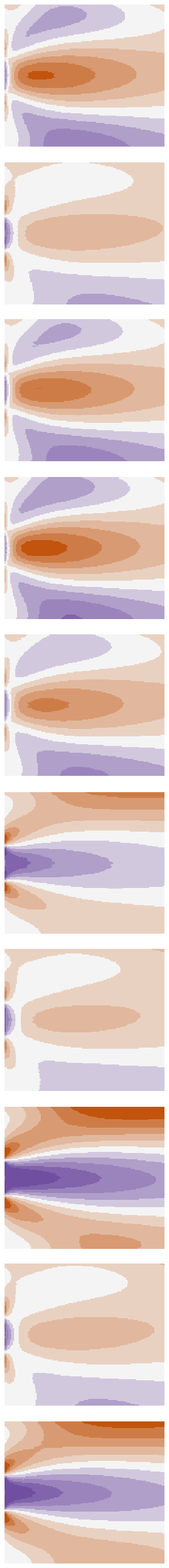}
		&
		\includegraphics[width=0.08\linewidth]{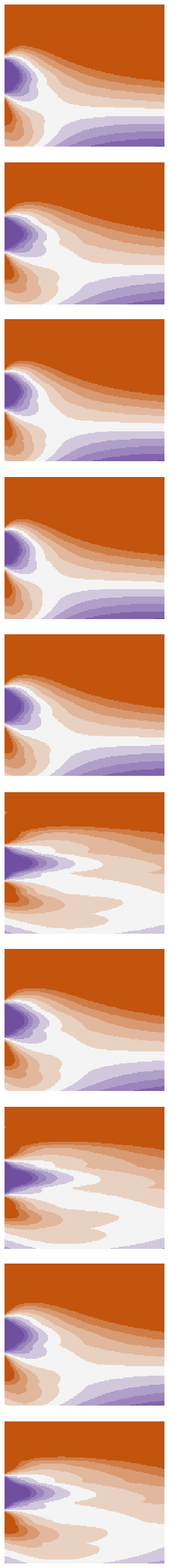}
		&
		\includegraphics[width=0.08\linewidth]{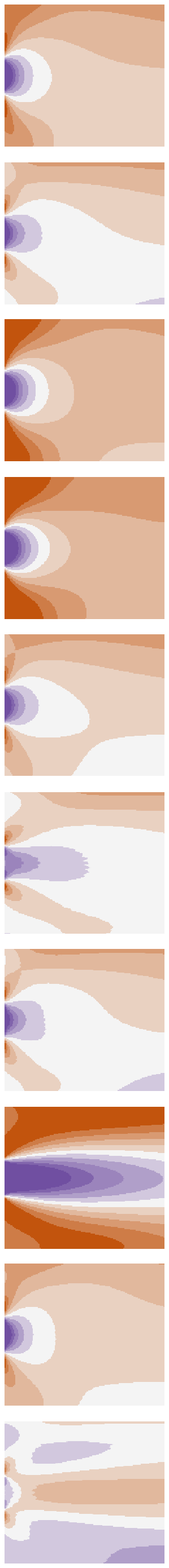}		
	\end{tabular}
	\caption{\small Visualization of local errors in \textit{Heat}-2. Each image represents the difference between the prediction and ground-truth over individual outputs of a test example. From the left column to the right are the results of \ours, PCA-GP-\{1,2\}, KPCA-GP-\{1,2\}, IsoMap-GP-\{1,2\} and HOGP-\{1,2\}.} 
	\label{fig:heat}
\end{figure*}

\begin{figure*}
	\centering
	\setlength\tabcolsep{0.05pt}
	\begin{tabular}[c]{ccccccccc}
		\multicolumn{9}{c}{\includegraphics[width=0.3\linewidth]{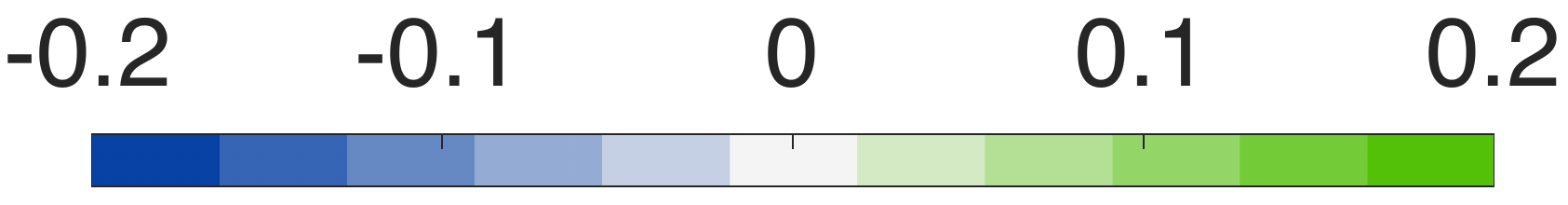}} \\
		\includegraphics[width=0.08\linewidth]{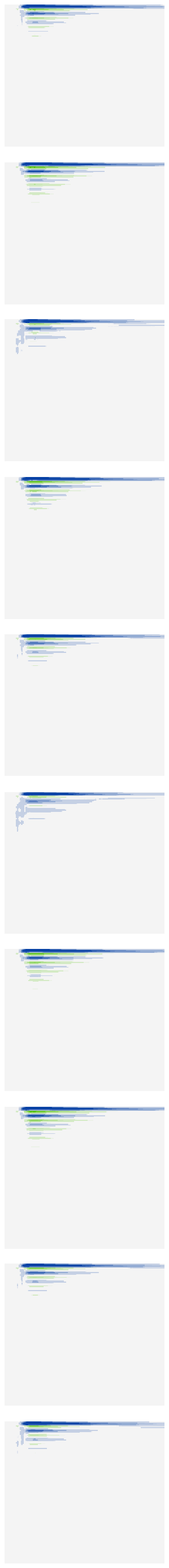}
		&
		\includegraphics[width=0.08\linewidth]{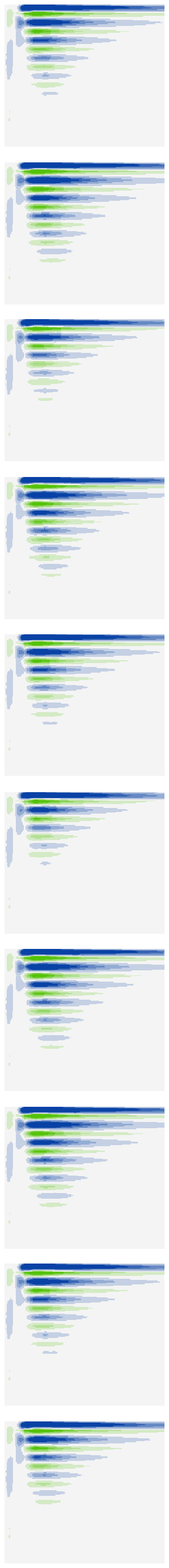} & 
		\includegraphics[width=0.08\linewidth]{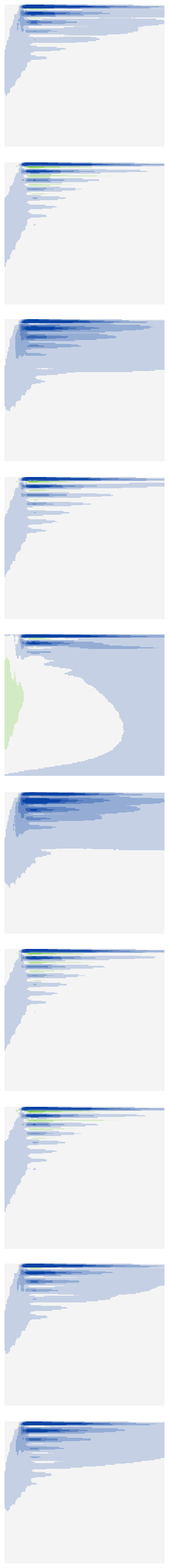} 
		&
		 \includegraphics[width=0.08\linewidth]{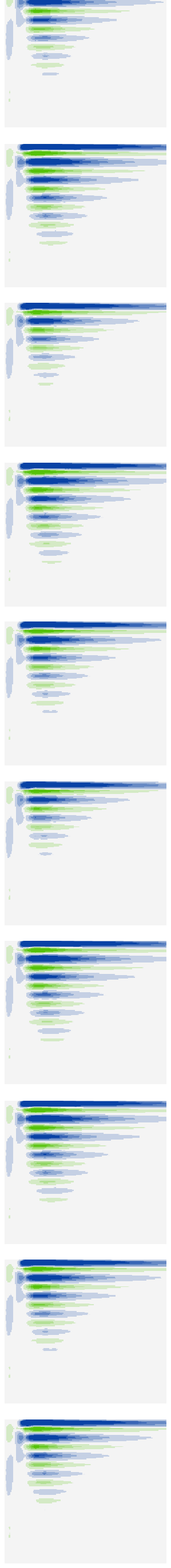}
		&
		\includegraphics[width=0.08\linewidth]{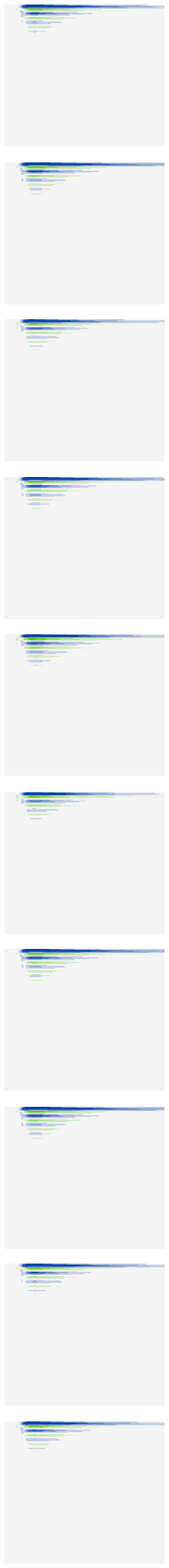}
		&
		\includegraphics[width=0.08\linewidth]{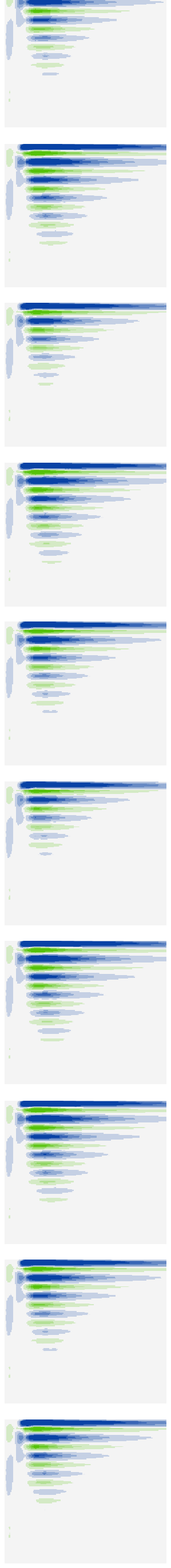}
		&
		\includegraphics[width=0.08\linewidth]{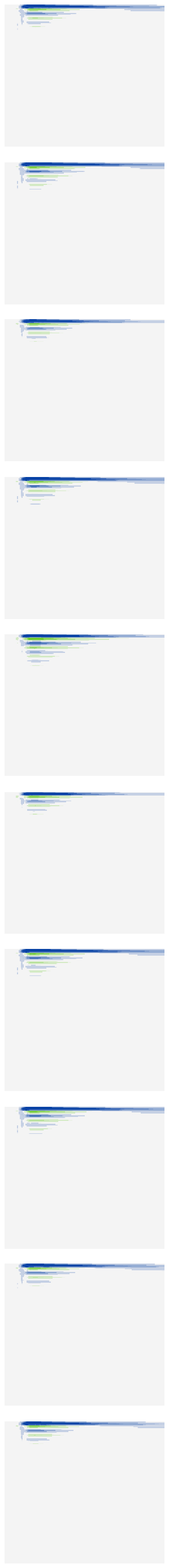}
		&
		\includegraphics[width=0.08\linewidth]{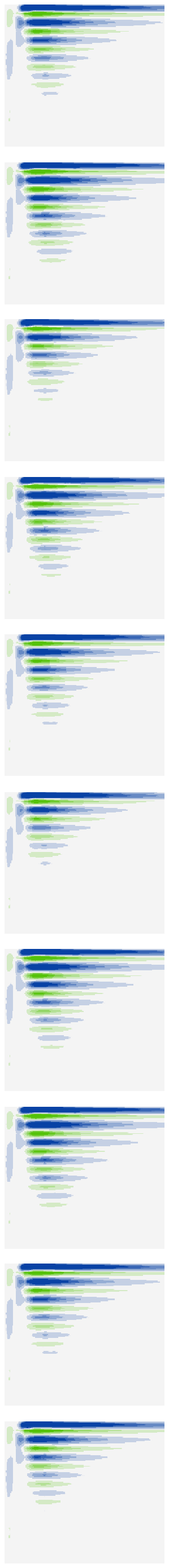}
		&
		\includegraphics[width=0.08\linewidth]{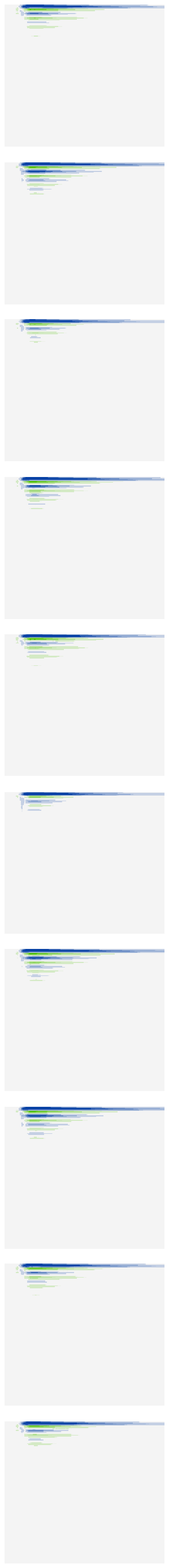}		
	\end{tabular}
	\caption{\small Visualization of local errors in \textit{Burgers}-2. From the left column to the right are the results of \ours, PCA-GP-\{1,2\}, KPCA-GP-\{1,2\}, IsoMap-GP-\{1,2\} and HOGP-\{1,2\}. } 
	\label{fig:burger2}
\end{figure*}

\begin{figure*}
	\centering
	\setlength\tabcolsep{0.05pt}
	\begin{tabular}[c]{ccccccccccccc}
		\multicolumn{13}{c}{\includegraphics[width=0.3\linewidth]{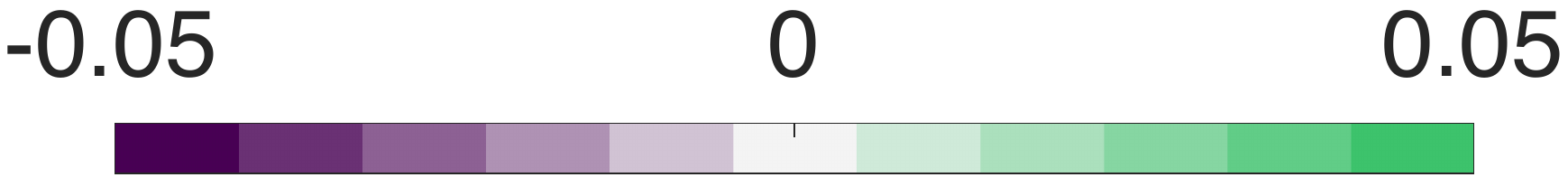}} \\
		\includegraphics[width=0.06\linewidth]{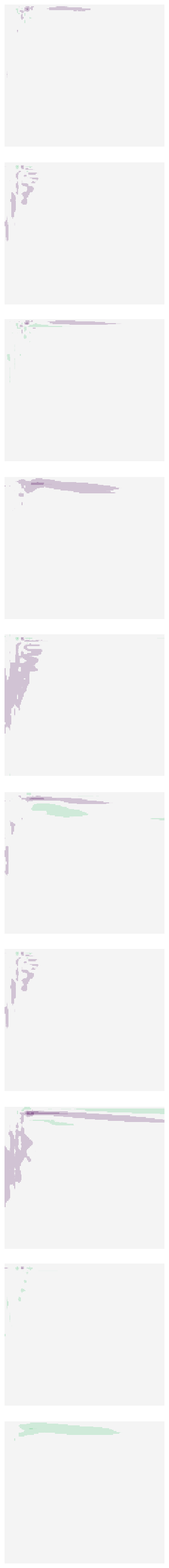}
		&
		\includegraphics[width=0.06\linewidth]{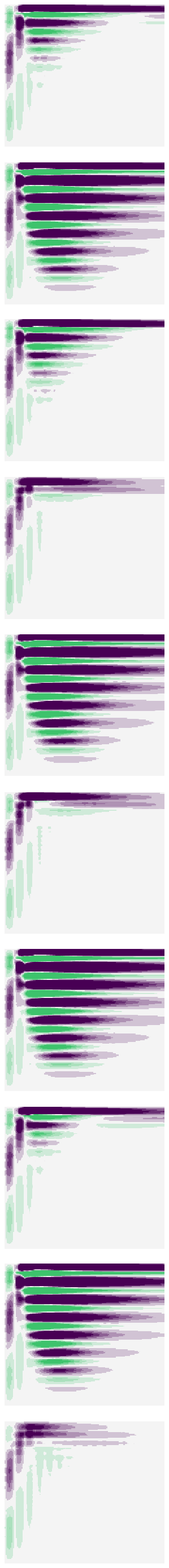} & 
		\includegraphics[width=0.06\linewidth]{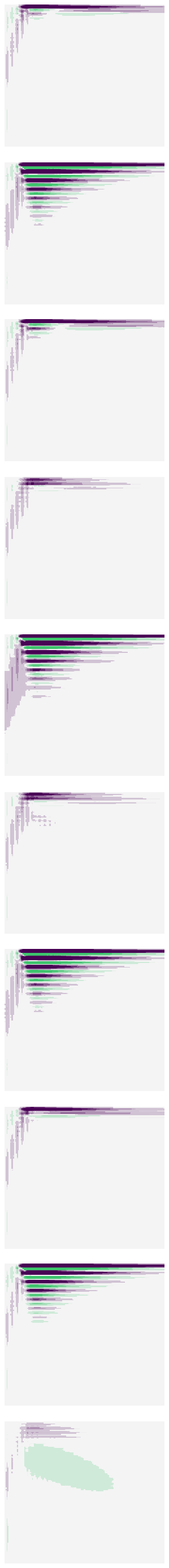} 
		&
		\includegraphics[width=0.06\linewidth]{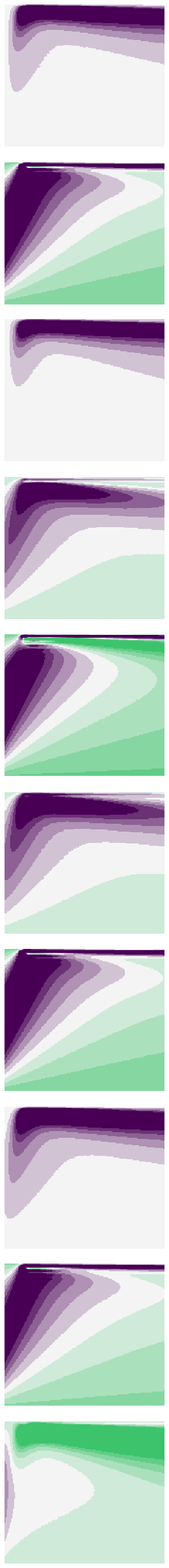} 
		&
		 \includegraphics[width=0.06\linewidth]{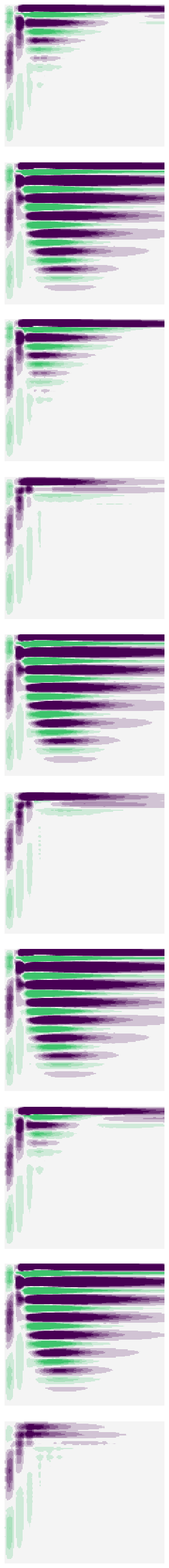}
		&
		\includegraphics[width=0.06\linewidth]{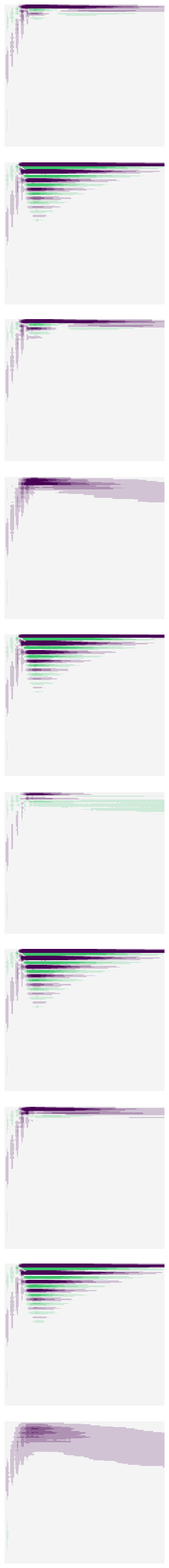}
		&
		\includegraphics[width=0.06\linewidth]{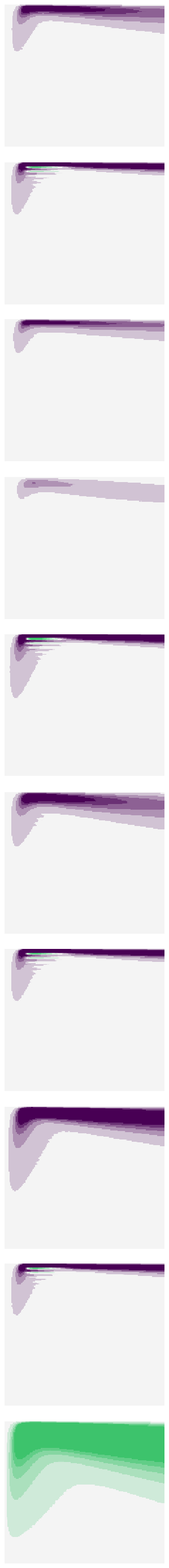}
		&
		\includegraphics[width=0.06\linewidth]{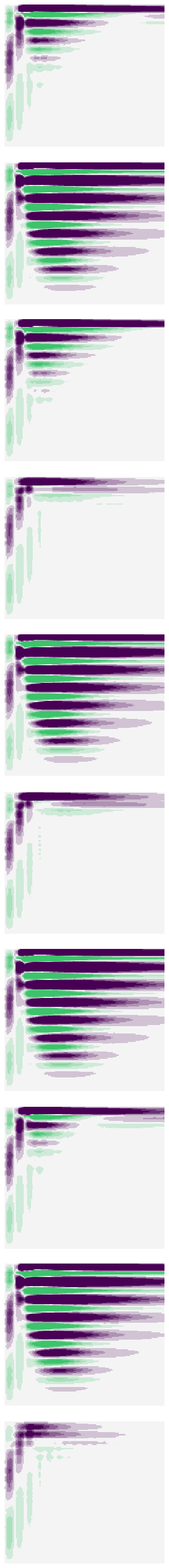}
		&
		\includegraphics[width=0.06\linewidth]{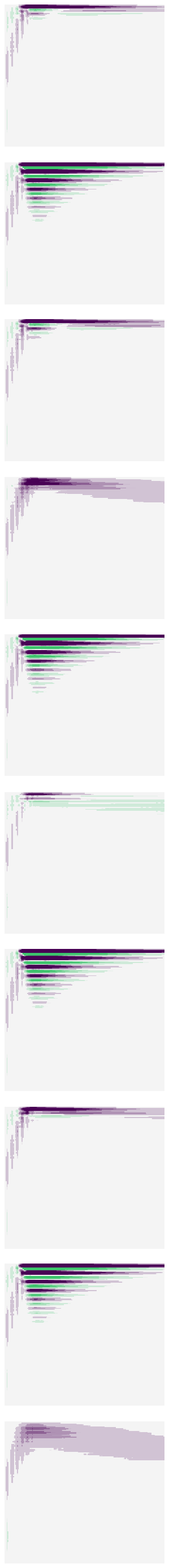}
		&
		\includegraphics[width=0.06\linewidth]{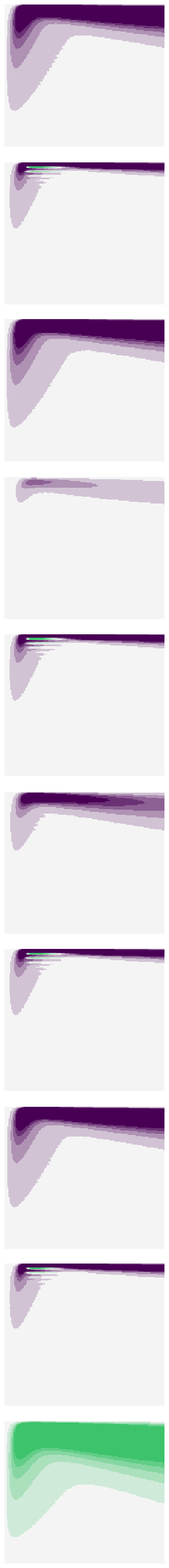}
		&
		\includegraphics[width=0.06\linewidth]{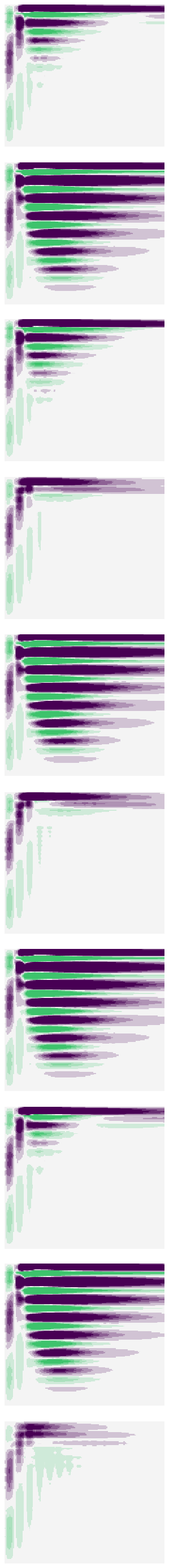}
		&
		\includegraphics[width=0.06\linewidth]{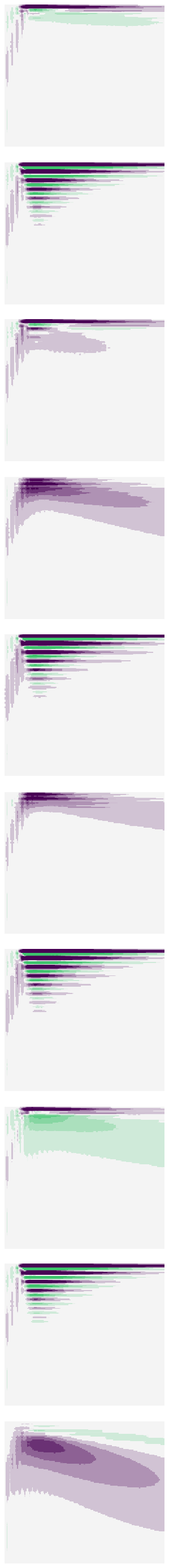}		
		&
		\includegraphics[width=0.06\linewidth]{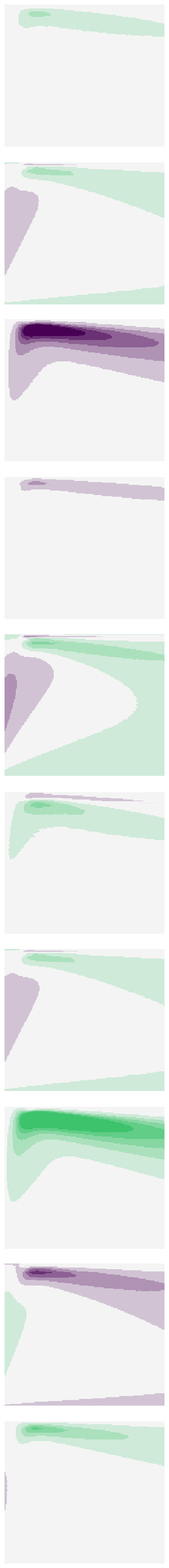}
	\end{tabular}
	\caption{\small Visualization of local errors in \textit{Burgers}-3. From the left column to the right are the results of \ours, PCA-GP-\{1,2,3\}, KPCA-GP-\{1,2,3\}, IsoMap-GP-\{1,2,3\} and HOGP-\{1,2,3\}.}  
	\label{fig:burger3}
\end{figure*}

\begin{figure*}
	\centering
	\setlength\tabcolsep{0.05pt}
	\begin{tabular}[c]{ccccccccc}
		\multicolumn{9}{c}{\includegraphics[width=0.3\linewidth]{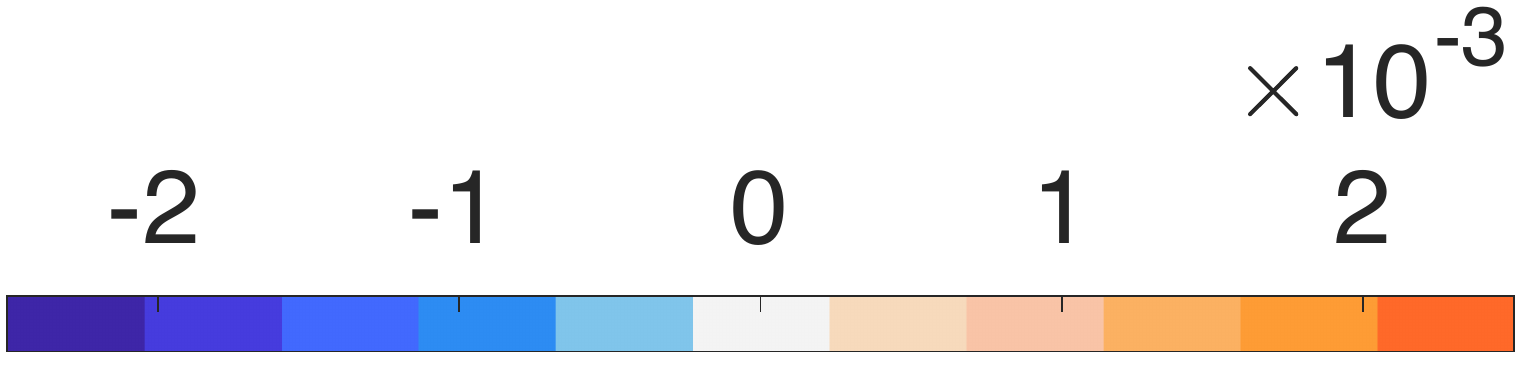}} \\
		\includegraphics[width=0.08\linewidth]{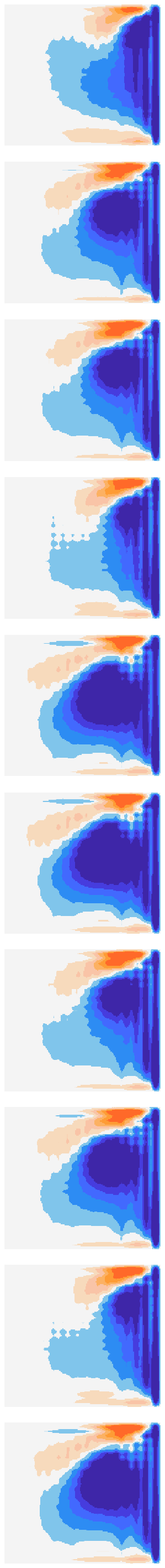}
		&
		\includegraphics[width=0.08\linewidth]{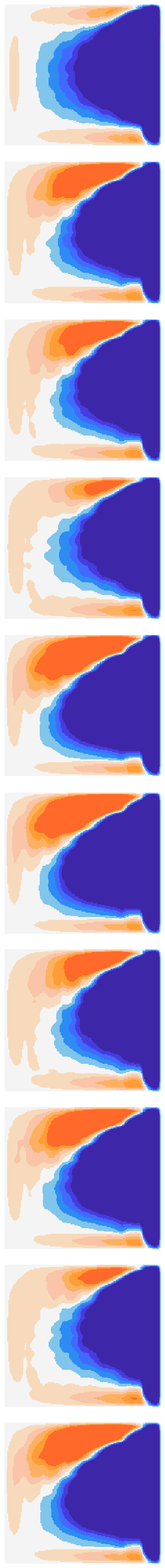} & 
		\includegraphics[width=0.08\linewidth]{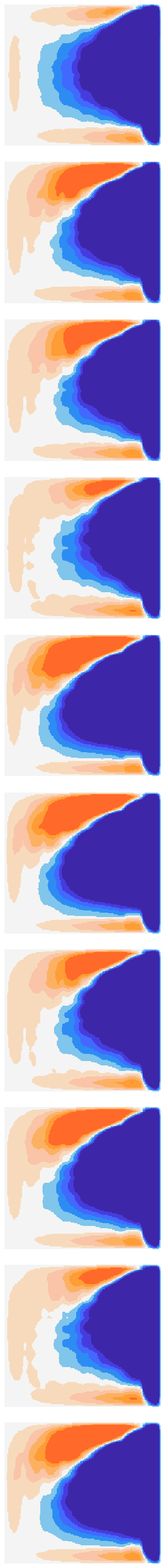} 
		&
		 \includegraphics[width=0.08\linewidth]{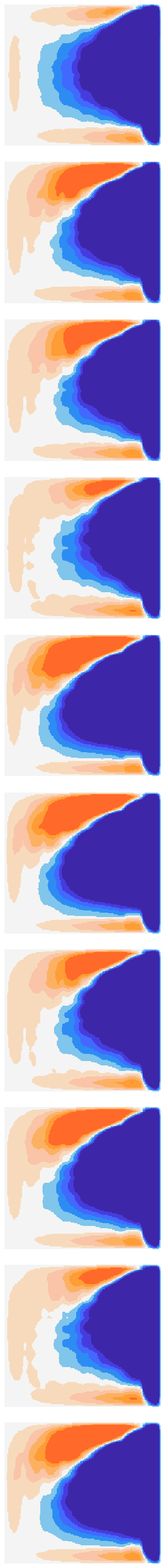}
		&
		\includegraphics[width=0.08\linewidth]{./fig/NS_200x20/NS_200x20_kpcaf2_s10.eps}
		&
		\includegraphics[width=0.08\linewidth]{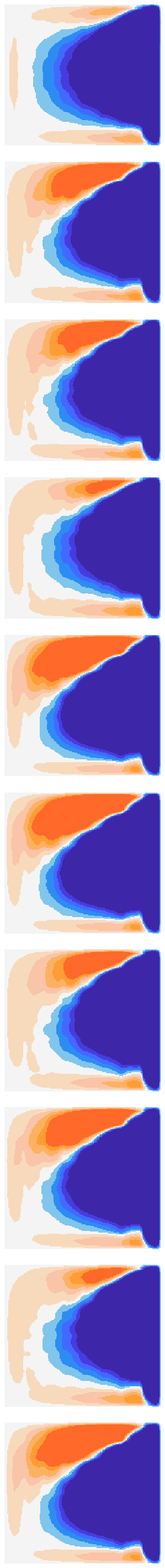}
		&
		\includegraphics[width=0.08\linewidth]{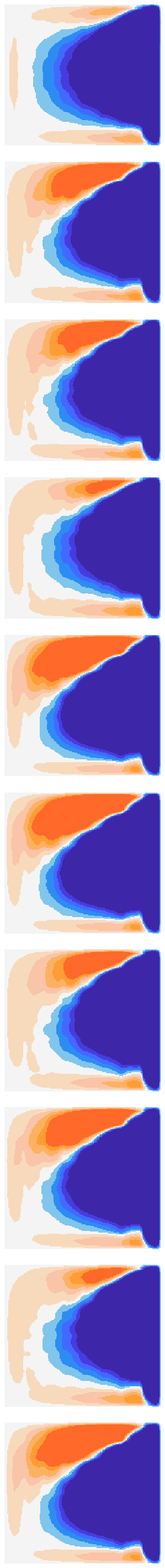}
		&
		\includegraphics[width=0.08\linewidth]{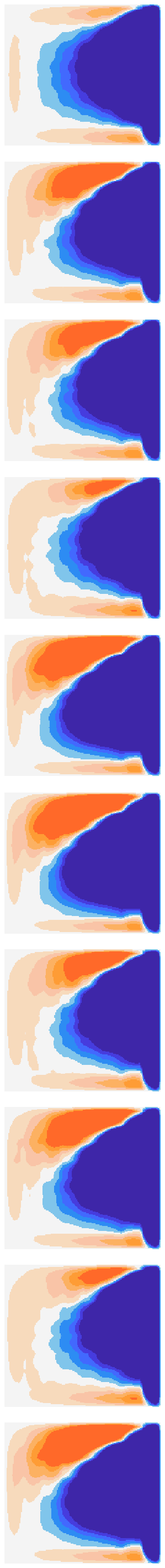}
		&
		\includegraphics[width=0.08\linewidth]{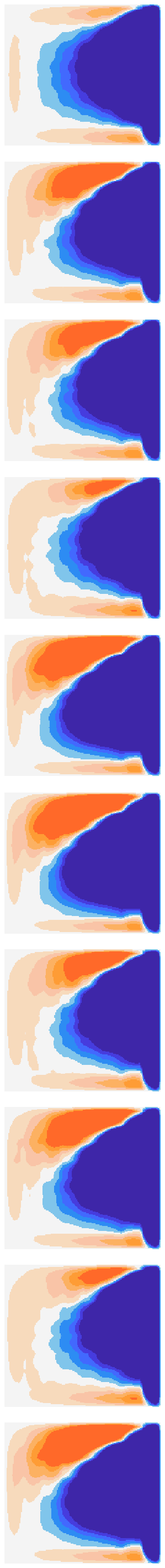}		
	\end{tabular}
	\caption{\small Visualization of local errors in the one-million pressure field prediction for lid-driven cavity flows. Each image represents the difference between the prediction and ground-truth over individual outputs of a test spatial field ($100 \times 100$) at a randomly chosen time point, in the setting of \textit{F1=200, F2=20}.  From the left column to the right are the results of \ours, PCA-GP-\{1,2\}, KPCA-GP-\{1,2\}, IsoMap-GP-\{1,2\} and HOGP-\{1,2\}.}
	\label{fig:ns1}
\end{figure*}

We supplement the local prediction results in the settings of 
\textit{Heat}-2, \textit{Burgers}-2 and \textit{Burgers}-3, which are shown in Fig. \ref{fig:heat}, \ref{fig:burger2} and \ref{fig:burger3}, respectively. We also show the local prediction results of simulating lid-driven cavity flows in Fig. \ref{fig:ns1}  \textit{(F1=200, F2=20)}. 
Each column are the results of one method.  The leftmost is \ours.  For each setting, we show the difference between the prediction and  ground-truth in 10 randomly selected test fields. 
As we can see, \ours always obtains better predictions for individual outputs. This is implied by the fact that most output regions of \ours are rendered by lighter colours. Hence, it confirms that our method not only yields a superior global prediction accuracy but also better recovers individual outputs locally. 


\end{document}